\documentclass[Afour,sageh, times]{sagej}
\usepackage{url}
\usepackage{graphicx} 
\usepackage{amsmath} 
\usepackage{amssymb}  
\usepackage{amsthm}
\usepackage{multirow}
\usepackage{algorithm}
\usepackage{algpseudocode}
\usepackage{xcolor}
\usepackage[colorlinks,bookmarksopen,bookmarksnumbered,citecolor=red,urlcolor=red]{hyperref}
\usepackage{subcaption}
\usepackage{wrapfig}
\usepackage{cancel}
\usepackage[verbose, hyperpageref]{backref}

\usepackage[title]{appendix}
\usepackage[font=normalsize,labelfont=bf]{caption}

\usepackage{nicefrac}
\usepackage{amsmath,amssymb,amsfonts}
\usepackage{graphicx}
\usepackage{xcolor}
\usepackage{caption}
\usepackage{subcaption}
\usepackage{algorithm}
\usepackage{algpseudocode}
\usepackage{amsthm}
\usepackage{tabularx}
\usepackage{multirow}
\usepackage{multicol}
\usepackage{booktabs}
\usepackage{lettrine}

\newcommand{\bydef}{\triangleq}
\newcommand{\probd}{\mathbb{P}}	 
\newcommand{\lucb}{\underline{\mathrm{UCB}}}
\newcommand{\uucb}{\overline{\mathrm{UCB}}}
\newcommand{\maxim}[2]{\ensuremath{\underset{#2}{\mathrm{#1}}}}
\DeclareMathOperator*{\argmax}{arg\,max}

\theoremstyle{plain}
\newtheorem{thm}{\protect\theoremname}
\theoremstyle{remark}
\theoremstyle{plain}
\newtheorem{lem}{\protect\lemmaname}
\theoremstyle{definition}
\newtheorem{defn}{\protect\definitionname}

\providecommand{\definitionname}{Definition}
\providecommand{\lemmaname}{Lemma}
\providecommand{\theoremname}{Theorem}
\newtheorem{assumption}{Assumption}

\newtheorem{case}{Case}
\newcount\savedcase
\newenvironment{subcases}{%
	\savedcase=\value{case}%
	\edef\prevthecase{\thecase}%
	\setcounter{case}{0}%
	\renewcommand\thecase{\prevthecase.\arabic{case}}%
}
{%
	\setcounter{case}{\savedcase}%
}
\renewcommand{\qed}{$\blacksquare$}
\algrenewcommand{\algorithmiccomment}[1]{\hfill// #1}
\title{No Compromise in Solution Quality: Speeding Up Belief-dependent Continuous POMDPs via Adaptive Multilevel Simplification}
	
\author{Andrey Zhitnikov\affilnum{1}, Ori Sztyglic\affilnum{2}, and Vadim Indelman\affilnum{3}}

\affiliation{\affilnum{1} Technion Autonomous Systems Program (TASP),  Technion - Israel Institute of Technology \\
\affilnum{2} Department of Computer Science,  Technion - Israel Institute of Technology \\
\affilnum{3} Department of Aerospace Engineering, Technion - Israel Institute of Technology}
\corrauth{ Andrey Zhitnikov,  Technion - Israel Institute of Technology, Haifa 3200003, Israel.}
	
\email{\href{mailto:andreyz@campus.technion.ac.il}{andreyz@campus.technion.ac.il}}

\begin{document}
\runninghead{Zhitnikov, Sztyglic, and Indelman}

\thispagestyle{empty}

\begin{abstract}	
Continuous POMDPs with general belief-dependent rewards are notoriously difficult to solve online. In this paper, we present a complete provable theory of adaptive multilevel simplification for the setting of a given externally constructed belief tree and MCTS that constructs the belief tree on the fly using an exploration technique. Our theory allows to accelerate POMDP planning with belief-dependent rewards without any sacrifice in the quality of the obtained solution. We rigorously prove each theoretical claim in the proposed unified theory. Using the general theoretical results, we present three algorithms to accelerate continuous POMDP online planning with belief-dependent rewards. Our two algorithms, SITH-BSP and LAZY-SITH-BSP, can be utilized on top of any method that constructs a  belief tree externally. The third algorithm, SITH-PFT, is an anytime MCTS method that permits to plug-in any exploration technique. All our methods are guaranteed to return exactly the same optimal action as their unsimplified equivalents. We replace the costly computation of information-theoretic rewards with novel adaptive upper and lower bounds which we derive in this paper, and are of independent interest. We show that they are easy to calculate and can be tightened by the demand of our algorithms. Our approach is general; namely, any bounds that monotonically converge to the reward can be utilized to achieve significant speedup without any loss in performance. Our theory and algorithms support the challenging setting of continuous states, actions, and observations. The beliefs can be parametric or general and represented by weighted particles. We demonstrate in simulation a significant speedup in planning compared to baseline approaches with guaranteed identical performance.
\end{abstract}
\maketitle

\section{Introduction} 
\lettrine[findent=2pt]{\fbox{\textbf{E}}}{fficiently} solving Partially Observable Markov Decision Processes (POMDPs)  implies enabling autonomous agents and robots to plan under uncertainty \citep{Smith04uai, Kurniawati08rss, Silver10nips, Ye17jair, Sunberg18icaps, Garg19rss}. Typical sources of uncertainty are the imprecise actions, sensor type, sensor noise, imprecise models, and unknown agent surroundings. However,  solving a POMDP is notoriously hard. Specifically, it was proven to be PSPACE-complete \citep{Papadimitriou87math}.  

\begin{figure}[t]
	\centering
	\includegraphics[width=\columnwidth]{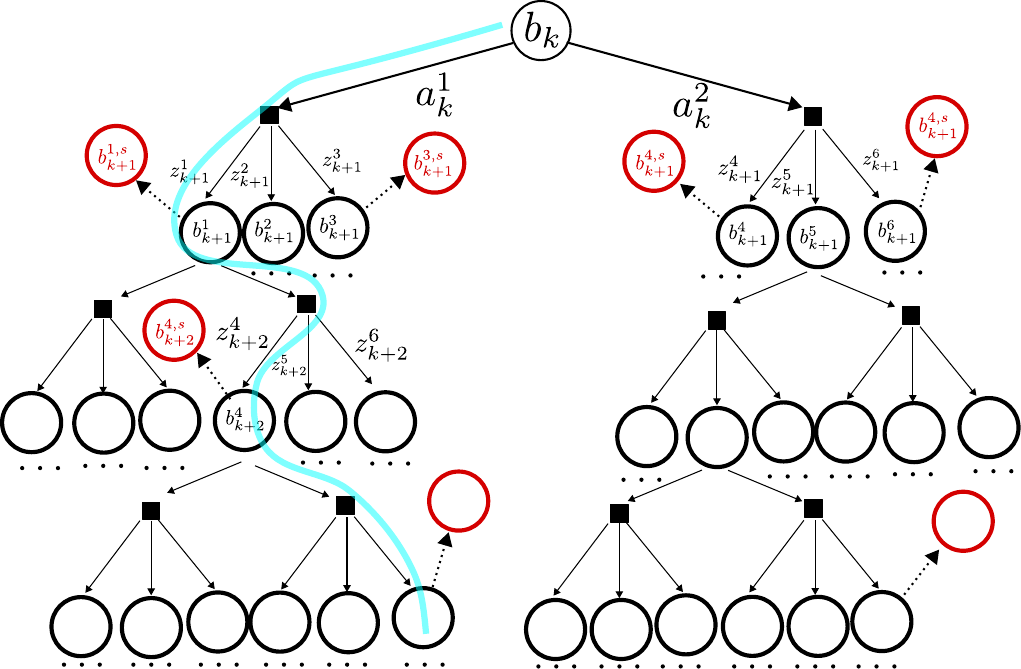}
	\caption{Schematic visualization of the belief tree and the inplace simplification. The superscript in this visualization denotes the index in the belief tree. By $b^s$ we denote the simplified version of the belief $b$.}  
	\label{fig:BeliefTreeInplaceSimp}
\end{figure}

The actual POMDP state is hidden. Instead, at each time step, the robot shall decide which action to take based on the distribution over the state, given the corresponding history of performed actions and observations received so far. Such a distribution received the name ``belief''.  In a planning session, the robot has to take into account all possible future actions interleaved with possible observations.  Each such future history of the length of predefined horizon defines a lace of the future beliefs (blue lace in Fig.~\ref{fig:BeliefTreeInplaceSimp}) and corresponding cumulative rewards named return. Solving POMDP in the most common sense means finding a mapping from belief to action called policy, which maximizes the expected return.

Earlier \emph{offline} solvers such as \citep{Smith04uai, Kurniawati08rss} are applicable to small or moderately sized discrete POMDP. These methods require passage over all possible states and observations \citep{Kochenderfer22book} since they are built on value iteration of $\alpha$-vectors, so called full-width methods \citep{Silver10nips}.  
More recent  \emph{online} solvers are suitable for POMDPs with large but discrete action, state, and observation spaces \citep{Ye17jair, Silver10nips}. Still, continuous state, action, and observation spaces remain to be an open problem \citep{Sunberg18icaps}. Another challenging aspect of solving POMDP and the subject of interest in this paper is general belief distributions represented by weighted particles.  
Further in the manuscript we will regard the combination of both, nonparametric beliefs and a fully continuous POMDP as a {\bf nonparametric fully continuous} setting. 

In a fully continuous setting with parametric or general beliefs one shall resort to  sampling of future possible actions and observations. 
In a sampled form, this abundance of possible realizations of action-observation pairs constitutes a \emph{belief tree}.
Building the full belief tree is intractable since each node in the tree repeatedly branches with all possible actions and all possible observations as illustrated in Fig.~\ref{fig:BeliefTreeInplaceSimp}. The number of nodes grows exponentially with the horizon. This problem is known as the \emph{curse of history}. 

The reward function in a classical POMDP is assumed to have a specific structure, namely, to be the expectation with respect to the belief of the state-dependent reward function. While alleviating the solution, this formulation does not support more general, belief-dependent reward functions, such as information-theoretic rewards. 

However, POMDP planning with belief-dependent rewards is essential for various problems in robotics and Artificial Intelligence (AI), such as informative planning \citep{Hollinger14ijrr}, active localization \citep{Burgard97ijcai}, active Simultaneous Localization and Mapping (SLAM)  \citep{Stachniss05rss}, Belief Space Planning (BSP) \citep{Indelman15ijrr, VanDenBerg12ijrr, Platt10rss}. The authors of \citep{Araya10nips} provide an extensive motivation for general belief-dependent rewards. One of the widely used such rewards  is Information Gain, which  involves the difference between differential entropies of two consecutive in time beliefs. Such a reward is crucial in exploration tasks because, in these tasks, the robot's goal is to decrease uncertainty over the belief. For instance, uncertainty measures such as differential entropy and determinant of the covariance matrix of the belief cannot be represented as  expectation over a state-dependent reward with respect to the belief.  Another example of a  belief-dependent reward is entropy over discrete variables that correspond to data association hypotheses \citep{Pathak18ijrr}.   
Computationally-efficient  information-theoretic BSP approaches have been investigated in recent years, considering Gaussian  distributions  \citep{Kopitkov17ijrr, Kopitkov19ijrr, Elimelech22ijrr, Kitanov24ijrr}.

Yet, POMDP planning with general belief-dependent rewards in particular, when the beliefs are represented by particles exacerbate the computational challenge of the solution even more. For example information theoretic rewards such as  differential entropy, are computationally expensive.

Let us focus for the moment on differential entropy. Even if the belief is parametric but not Gaussian, calculating the exact value of differential entropy involves intractable integrals. This fact also motivates to use a weighted particles representation for the belief. In this case differential entropy can be estimated, for instance by Kernel Density Estimation (KDE) \citep{Fischer20icml}  or a model-based estimator \citep{Boers10fusion}. However, these estimators have quadratic cost in the number of samples and are usually the bottleneck of planning algorithms. The reason is that  this increased computational burden is incurred for all nodes in the belief tree.  Importantly, the estimation errors of these estimators with respect to differential entropy over theoretical belief are out of the reach due to the unavailability of both, the theoretical belief and the entropy on top of it. Yet, due to the convergence  of the belief represented by particles to the theoretical belief (almost sure convergence \citep{Crisan02tsp}), the mentioned above estimators converge to the exact differential entropy. 
This prompts us to use {\bf as many belief particles as possible} to get closer to the theoretical belief. Nevertheless, increasing the number of belief particles severely impacts planning time.

In this paper we accelerate online decision making in the setting of nonparametric fully continuous POMDPs with general belief dependent rewards. Crucially, planning performance of our accelerated approach is the same as that of the baseline approaches without our acceleration. Before stating our contributions, we review the most relevant works in this context.

\subsection{Related Work} \label{sec:RelatedWork}
Allowing general belief-dependent rewards in POMDP while solving such a problem efficiently is a long standing effort. Some previous seminal works such as  $\rho$-POMDP \citep{Araya10nips, Fehr18nips} as well as \citep{Dressel17icaps} have focused on discrete domains, small sized spaces and have tackled the offline solvers. Furthermore, these approaches are limited to piecewise linear and convex or Lipschitz-continuous rewards.  Another work named POMDP-IR \citep{Spaan2015aamas} suggest an interesting framework for specific form of information rewards involving manipulations on the action space.  Still, in \citep{Araya10nips, Fehr18nips, Dressel17icaps}  the state, action and observation spaces are discrete and small sized.  Another line of works is Belief Space Planning (BSP) \citep{Platt10rss, VanDenBerg12ijrr, Indelman15ijrr}. 
These approaches are designed for fully continuous POMDPs, but limited to Gaussian beliefs. 
In striking contrast, our approach is centered in the more challenging fully continuous domain and nonparametric general beliefs represented by particles while at the same time our framework is general and supports also exact parametric beliefs. 

One way to tackle a nonparametric fully continuous setting with belief dependent rewards is to reformulate POMDP as a Belief-MDP (BMDP). On top of this reformulation one can utilize MDP sampling based methods such as Sparse Sampling (SS) proposed by \cite{Kearns02jml}. However, this algorithm still suffers from the curse of history and such that increasing the horizon is still problematic.  

Monte Carlo Tree Search (MCTS) made a significant breakthrough in overcoming the course of history by building the belief tree incrementally and exploring only the ``promising''  parts of the tree using the exploration strategy. An inherent part of MCTS based algorithms is the exploration strategy designed to balance exploration and exploitation while building the belief tree.  Most widely used exploration technique is Upper Confidence Bound (UCB) \citep{Kocsis06ecml}.

MCTS algorithms assume that calculating the reward over the belief node does not pose any computational difficulty. Information-theoretic rewards violate this assumption.
When the reward is a general function of the belief, the origin of the computational burden is shifted towards the reward calculation. Moreover, in a non-parametric setting, belief-dependent rewards require a complete set of belief particles at each node in the belief tree. Therefore, algorithms such as POMCP  \citep{Silver10nips}, and its numerous predecessors are inapplicable since they simulate each time a single particle down the tree when expanding it. DESPOT based algorithms behave similarly \citep{Ye17jair}, with the DESPOT-$\alpha$ as an exception \citep{Garg19rss}. 
DESPOT-$\alpha$ simulates a complete set of particles. However, the DESPOT-$\alpha$ tree is built using  $\alpha$-vectors, such that they are an indispensable part of the algorithm. The standard $\alpha$-vectors technique requires that the reward is state dependent, and the reward over the belief is merely expectation over the state reward. In other words DESPOT-$\alpha$ does not support belief-dependent rewards since it contradicts the application of the $\alpha$-vectors.

The only approach posing no restrictions on the structure of belief-dependent reward and not suffering from limiting assumptions 
is Particle Filter Tree (PFT). The idea behind  PFT is to apply MCTS over Belief-MDP (BMDP). The authors of \citep{Sunberg18icaps} augmented PFT  with Double Progressive Widening (DPW) to support continuous spaces in terms of actions, states and observations, and coined the name PFT-DPW. PFT-DPW utilizes the UCB strategy and maintains a complete belief particle set at each belief tree node.  
Recently, \cite{Fischer20icml} presented Information Particle Filter Tree (IPFT), a method to incorporate  information-theoretic rewards into PFT. 
The IPFT simulates small subsets of particles sampled from the root of the belief tree and averages entropies calculated over these subsets, enjoying a fast runtime. However,  differential entropy estimated from a small-sized particle set can be significantly biased. This bias is unpredictable and unbounded, therefore, severely impairs the performance of the algorithm. In other words, celerity comes at the expense of quality. 
Oftentimes, the policy defined by this algorithm is very far from optimal given a time budget.   
\cite{Fischer20icml} provides guarantees solely for the asymptotic case, i.e, the number of subsampled from the root belief state samples (particles) tends to infinity. Asymptotically their algorithm behaves precisely as the PFT-DPW in terms of running speed and performance. Yet, in practice the performance of IPFT in terms of optimality can degrade severely compared to PFT-DPW. Moreover, \cite{Fischer20icml} does not provide any  study of  comparison of IPFT against PFT-DPW with an information-theoretic reward. Another undesired characteristic of IPFT is that the averaging of the differential entropies is done implicitly and the number of averaged entropies per belief is the visitation count of the corresponding belief. Therefore, to properly compare IPFT with PFT-DPW one shall increase the number of simulations inside IPFT algorithm. We explain this aspect more thoroughly in Section~\ref{sec:SafeLoc}. Prompted by these insights, we chose the  PFT-DPW as our \emph{baseline} approach, which we aim to accelerate. In contrast to IPFT designed specifically for differential entropy, our approach is suitable for any belief dependent reward  and 
explicitly guarantees an \emph{identical} solution to PFT-DPW with an information-theoretic reward, for \emph{any} size of particle set representing the belief and serving as input to PFT-DPW.

The computational burden incurred by the complexity of POMDP planning inspired many research works to focus on approximations of the problem on top of existing solvers, e.g., multilevel successive approximation of a motion model \citep{Hoerger19isrr}, lazy belief extraction on top of a particle based representation \citep{Hoerger21icra},  linearity based solvers \citep{Hoerger20sp}, and averaging differential entropy estimated from tiny subsets of particles \citep{Fischer20icml}. Typically, these works provide only asymptotical guarantees \citep{Hoerger19isrr, Fischer20icml}, or no guarantees at all. In addition many of these approximations leverage the assumption that the belief-dependent reward is an averaged state-dependent reward, e.g, \citep{Hoerger19isrr, Hoerger21icra}, and therefore cannot accommodate belief dependent-rewards with general structure (e.g.~do not support information-theoretic rewards such as differential entropy).

Recently, the novel paradigm of \emph{simplification} has appeared in literature  \citep{Zhitnikov22ai, Barenboim22ijcai, Barenboim23nips, Zhitnikov24tro,  Sztyglic22iros, Elimelech22ijrr, Shienman22isrr, Kitanov24ijrr, LevYehudi24aaai}. The simplification is concerned with carefully replacing the nonessential elements of the decision making problem and quantifying the impact of this relaxation. Specifically, simplification methods are accompanied by stringent guarantees. A prominent aspect of a simplification paradigm is the usage of the bounds over the reward or the objective function. As opposed to approximations, the simplification framework always keeps some sort of connection to the original unsimplified problem and by that provides deterministic guarantees relative to the given solver. Despite that various objective function bounds have been practiced in  \citep{Ye17jair, Smith04uai, Walsh10aaai, Kochenderfer22book}, these techniques are not applicable in the realm of belief-dependent rewards and a fully continuous setting. In addition commonly these approaches assume that the state dependent reward is trivially bounded from below and above by some constant.     

\subsection{Contributions}
This work is about accelerating online decision making while obtaining exactly the same solution as without acceleration. Specifically, we contribute an adaptive multi-level simplification framework that accounts for belief-dependent rewards, possibly nonparametric beliefs, and continuous state, observation and action spaces. 

	
Our framework accepts as input adaptive monotonical and computationally inexpensive bounds over the exact or estimated reward. Given such reward bounds, it accelerates online decision-making.   Specifically, given such adaptive monotonical reward bounds, 
it is possible to adaptively bound the value function for any given policy and expedite decision-making. If the value function bounds for different candidate policies do not overlap, we do not pay in terms of quality, namely, we obtain the same solution as the equivalent unsimplified method.  In the case these bounds do overlap, then we can progressively tighten them by invoking a process that we shall call simplification adaptation or \emph{resimplification} until they no longer overlap.

Our techniques return exactly the same solution as the unsimplified equivalent. 
Such an unsimplified baseline can correspond to 
decision-making problems where the reward can be exactly calculated (analytically), or where the reward is estimated. In either case, if the bounds over the corresponding reward are provided and satisfy the assumptions stated in Section~\ref{sec:Adaptive}, 
one can apply our framework to speedup the decision making process while obtaining the same best action as with the original rewards instead of the bounds. Such a capability is therefore particularly appealing in light of the information-theoretic rewards that are essential in numerous problems in robotics, but are often the computational bottleneck.
	
Further, there are two settings that we separately and explicitly discuss in this paper.   We  start from a given belief tree, that can be constructed by a POMDP solver that is not coupled with the solution, e.g., SS.  In this setting we can prune branches of the belief tree whenever the mentioned objective bounds for different candidate policies or actions do not overlap.  

We then discuss an anytime setting of MCTS, where the belief tree construction is coupled with the solution due to an exploration strategy (e.g.~UCB). The exploration strategy builds upon an exploratory objective. Since the exploratory objective typically requires access to the objective estimates to select an action at each arrival to a belief node, we cannot prune suboptimal candidate actions. Instead, we can only dismiss them until the next arrival to this belief node. The simplification and  reward bounds are used here to bound the exploratory objective and the value function at the root of the belief tree.

Finally, we focus on a specific simplification of nonparametric beliefs represented by particles and a differential entropy estimator as the reward function. Our simplification is subsampling of the original belief to a smaller sample size. 
We contribute novel computationally cheaper bounds over the differential entropy estimator on top of such a simplified belief and incorporate these bounds into our framework. By that we produce  a specific embodiment of the general framework presented earlier.   

To summarize, we list down the contributions of this work, in the order they are presented in the manuscript.
\begin{enumerate}
	\item  Building on {\bf any} adaptive monotonically convergent bounds over belief-dependent reward, we present in this paper a {\bf provable} general theory of adaptive multilevel simplification  with deterministic performance guarantees. 
	\item For the case of a given belief tree as in Sparse Sampling, we develop two algorithms, Simplified Information Theoretic Belief Space Planning (SITH-BSP) and a faster variant, LAZY-SITH-BSP. Both  are complementary to any POMDP solver that does not couple belief tree construction  with an objective estimation while exhibiting a significant speedup in planning with a guaranteed same planning performance. 
	\item In the context of MCTS, we embed the theory of simplification into the  PFT-DPW algorithm and introduce SITH-PFT. We  provide stringent guarantees that  exactly the same belief tree is constructed by SITH-PFT and PFT-DPW. We focus on a UCB exportation technique, but with minor adjustments, an MCTS with any exploration method will be suitable for acceleration.  
	\item We derive novel lightweight adaptive bounds on the differential entropy estimator of \citep{Boers10fusion} and prove the bounds presented are monotonic and convergent. Moreover, these bounds can be incrementally tightened. We believe these bounds are of interest on their own. The bounds are calculated using the simplified belief (See Fig.~\ref{fig:BeliefTreeInplaceSimp}). 
	We emphasize that any other bounds fulfilling assumptions declared in Section~\ref{sec:Adaptive} can be 
	utilized within our framework. 
	\item We present extensive simulations that exhibit a significant improvement in planning time without any sacrifice in planning performance.
\end{enumerate}
This paper is an extension of the work presented in \citep{Sztyglic22iros}, which 
proposed novel adaptive bounds on the differential entropy estimator of \citep{Boers10fusion} and introduced  the simplification paradigm in the context of a given belief tree.
To be precise we explicitly clarify how this work differs from the conference version of this paper \citep{Sztyglic22iros}. In this version, we extend the simplification framework to the rewards depending on a pair of consecutive-in-time beliefs, e.g., Information Gain as opposed to the conference version where such an extension was only mentioned. In this version, we provide alternative proof of these bounds and prove that these reward bounds are monotonic. In the setting of a given belief tree we present an additional algorithm, that we call LAZY-BSP. This algorithm is faster than SITH-BSP suggested in \citep{Sztyglic22iros}. Importantly, we extend our simplification framework to support also anytime MCTS planners.  Additionally, we provide extensive performance evaluation of our methods in simulations.

\subsection{Paper Organization}
The remainder of this paper is structured  as follows.  Section \ref{sec:POMDPbg} provides background in terms of POMDPs, theoretical objective and commonly used objective estimators. We devote Section \ref{sec:GeneralApproach} to our general adaptive multi-level simplification framework. In Section \ref{sec:GivenTree} we consider a given belief tree setting in which the belief tree construction is not coupled with the solution. In Section \ref{sec:MCTS} we delve into the MCTS approach in the context of our multilevel simplification. In Section \ref{sec:SpecificSimplification} we consider a specific simplification and develop novel bounds on an information-theoretic reward function. Section \ref{sec:AdaptationOverhead} assesses the general adaptation overhead of our methodology.  Finally, Section \ref{sec:SimsRes} presents simulations and results corroborating our ideas. 
In order not to disrupt the flow of the presentation, proofs  are presented in appropriate Appendices. 
\section{Background}\label{sec:POMDPbg}
In this section we present the background. To elaborate, we present a POMDP with belief dependent rewards followed by theoretical and estimated objectives that correspond to different online POMDP solvers. Our techniques work with estimated objectives.    
\subsection{POMDPs with Belief-dependent Rewards} 
A POMDP is a tuple 
\begin{align}
\langle \mathcal{X}, \mathcal{A}, \mathcal{Z}, T, \mathcal{O}, \rho, \gamma, b_{0}\rangle
\label{eq:pomdp}
\end{align}  where $\mathcal{X}, \mathcal{A}, \mathcal{Z}$ are state, action, and observation spaces, respectively.  In this paper we consider continuous state, observation and action spaces. $T(x,a,x')= \probd_T(x' | x, a) $ is the stochastic transition model from the past state $x$ to the subsequent  $x'$ through action $a$, $\mathcal{O}(z,x)=\probd_Z(z|x)$ is the stochastic observation model, $\gamma \in (0,1]$ is the discount factor, $b_0$ is the belief over the initial state (prior), and $\rho$ is the reward function. Let $h_k = \{b_0, a_0, z_1,  \dots, a_{k-1}, z_k\}$ denote \emph{history} of actions and observations obtained by the agent up to time instance $k$ and the prior belief. The posterior belief at time instant $k$ is given by $b_k(x_k)= \probd(x_k | h_k)$.

In our generalized formulation, the reward is a function of two subsequent in time beliefs, an action and an observation:  
\begin{align}
\!\rho(b_{k}, a_k, z_{k+1},b_{k+1})\! =  (1-\lambda)& r^{x}(b_k, a_k, b_{k+1})+  \label{eq:StateReward} \!\!\\
+\lambda &r^I(b_{k}, a_k, z_{k+1},b_{k+1}),\!\! \label{eq:InformationReward}
\end{align} 
where $\lambda \geq 0$. The first reward component  $r^{x}(b_k, a_k, b_{k+1})$ is the expectation over the  state and action dependent  reward $r(x_k,a_k)$ or $r(a_k, x_{k+1})$. Correspondingly, these two possibilities yield 
\begin{align}
&  r^{x}(b_k, a_k) = \underset{x_k \sim b_k}{\mathbb{E}}[r(x_k, a_k)] \approx \frac{1}{n_x}\sum_{\xi=1}^{n_x} r(x^{\xi}_k, a_k), \label{eq:previous}
\end{align}
or 
\begin{align}
&  \!\!\!\!r^{x}(a_k, b_{k+1}) \!=\!\!\!\!\!\!\!\!\! \underset{x_{k+1} \sim b_{k+1}}{\mathbb{E}}\!\!\!\![r(a_k, x_{k+1})] \! \approx \! \frac{1}{n_x}\sum_{\xi=1}^{n_x} r(a_k, x^{\xi}_{k+1}).\!\!\!\!
\end{align}
which is commonly approximated by sample mean using $n_x$ samples of the belief.  The second reward component $r^I(b_k, a_k, z_{k+1}, b_{k+1})$ is an information-theoretic reward weighted by $\lambda$, which in general can be dependent on consecutive beliefs and the elements relating them, e.g.~information gain or specific estimators as \citep{Boers10fusion} for nonparametric beliefs represented by particles. For instance, in Section \ref{sec:Bounds} we consider the entropy estimator introduced by \cite{Boers10fusion}. As will be seen in the sequel, 
although the theoretical entropy is only a function of a single belief $b_{k+1}$, the mentioned estimator utilizes $b_k$, $a_k$, $z_{k+1}$ and  $b_{k+1}$; hence the second reward component, $r^I(b_k, a_k, z_{k+1}, b_{k+1})$, depends on these quantities.

The \emph{policy} is a mapping from belief to action spaces $a_k=\pi_k(b_k)$. Let $\pi_{\ell+}$ be a  shorthand for policy for $\ell-k+L$ consecutive steps ahead  starting at index $\ell$, namely $\pi_{\ell:k+L-1}$ for $\ell \geq k$. 

\subsection{Theoretical Objective}
The decision making goal is to find an optimal policy $\pi_{k+}$ maximizing the value function as such:
\begin{equation}
	\begin{gathered}
		V(b_k, \pi_{k+}) \
		\text{s.t.} \ \ b_{\ell+1} = \psi(b_{\ell}, \pi_{\ell}(b_{\ell}), z_{{\ell}+1}), \label{eq:Obj}
	\end{gathered}
\end{equation}
where $V(b_k, \pi_k)$ is defined by 
\begin{equation}
\begin{gathered}
\underset{z_{k+1:k+L}}{\mathbb{E}}\!\Big[\sum_{\ell=k}^{k+L-1} \!\!\!\!\! \gamma^{\ell-k} \!\rho(b_{\ell}, \!\pi_{\ell}(b_{\ell}),\! z_{\ell+1},\! b_{\ell+1}) | b_k, \! \pi_{k+} \! \Big]\!\!\!\!   \label{eq:Value}
\end{gathered}
\end{equation}
and  $\psi$ is the Bayesian belief update method.  Utilizing the Bellman formulation \eqref{eq:Value} takes the form of 
\begin{equation}
\label{eq:Vbellman}	
\begin{gathered}
\!\!V(b_k, \pi_{k+})   = \!\!\underset{z_{k+1}}{\mathbb{E}} \big[\rho(b_{k}, \pi_{k}(b_k), z_{k+1}, b_{k+1}) | b_k, \pi_k\big]+ \!\!\!  \\
+\gamma \underset{z_{k+1}}{\mathbb{E}} \big[ V(\psi(b_k, a_k, z_{k+1}), \pi_{(k+1)+} )| b_k, \pi_k \big]. 	\!\!\!
\end{gathered}
\end{equation}
The action-value function under arbitrary policy is given by  
\begin{equation}
	\begin{gathered}
		\!\! Q(b_k,\! \{a_k,\! \pi_{(k+1)+}\})   {=}\!\! \underset{z_{k+1}}{\mathbb{E}}\! \big[\rho(b_{k}, a_k,\! z_{k+1},\! b_{k+1}) | b_k, \! a_k \big]\! {+} \!\! \\
		+\gamma\underset{z_{k+1}}{\mathbb{E}} \big[ V(\psi(b_k, a_k, z_{k+1}), \pi_{(k+1)+}) | b_k, a_k \big]. \!\! \label{eq:Qbellman}	
	\end{gathered}
\end{equation}
The relation between  \eqref{eq:Vbellman} and \eqref{eq:Qbellman} is $V(b_k, \pi_{k+})= Q(b_k, \{\pi_k(b_k), \pi_{(k+1)+}\})$. If $\pi$ is the optimal policy we denote it by $\pi^*$.  For clarity, let us designate for action-value function under optimal future policy $Q(b_k, \{a_k, \pi^*_{(k+1)+}\})$ a short notation $Q(b_k, a_k)$.
If $Q(b_k, a_k)$ can be calculated, the  online POMDP solution for the current belief $b_k$ will be 
\begin{align}
\pi^{\ast}_k(b_k)  \in  \maxim{\arg\max}{a_k} \ Q(b_k, a_k). \label{eq:OnlinePolicy}
\end{align}
Linearity of the expectation and the structure displayed by equations \eqref{eq:StateReward} and \eqref{eq:InformationReward} lead to a similar decomposition of action-value function \eqref{eq:Qbellman} as such
\begin{align}
	Q(\cdot) = (1-\lambda)Q^{x}(\cdot ) + \lambda Q^{I}(\cdot), \label{eq:dissectQ}
\end{align}
where $Q^{x}$ is induced by state dependent rewards and $Q^{I}$ by the information-theoretic rewards. 
 
From here on, for the sake of clarity, we will use the notation of history $h_k$ and the belief $b_k$ interchangeably for any time $k$. In a similar manner, we shall use the notations $b_k,a_k$ and $h_ka_k$ interchangeably.
\subsection{Estimated Objective} \label{sec:EstObj}
The continuous observation space makes the theoretical expectations in \eqref{eq:Value} and \eqref{eq:Qbellman} attainable in very limited cases. Generally we shall resort to estimators. 
Similar to theoretical counterparts, the relation between the estimated optimal value and action-value function reads   
\begin{align}
	\hat{V}(b_k, \pi^*_{k+}) = \maxim{\max}{a_k} \ \hat{Q}(b_k, a_k). \label{eq:OptValueSample}
\end{align}
Also in Eq.~\eqref{eq:OnlinePolicy}, the theoretical $Q(b_k, a_k)$ is substituted by the estimator $\hat{Q}(b_k, a_k)$.  Naturally, we expect from the estimator to admit the decomposition  
\begin{align}
	\hat{Q}(b_k, a_k) = (1-\lambda)\hat{Q}^{x}(b_k, a_k) + \lambda \hat{Q}^{I}(b_k, a_k). \label{eq:dissectQSample}
\end{align}
Typically the $\hat{Q}^{x}$ element is easy to calculate, thus it is out of our focus, whereas $\hat{Q}^{I}$ is computationally expensive to compute.  

Below we present two common sample based estimators that will be used in this paper.   
\subsubsection{Objective Estimator in Case of a  Given Belief Tree} \label{sec:GivenTreeEstimator}
We turn to the setting of a given externally-constructed belief tree, e.g. by a SS algorithm.  
For the sake of clarity and to ease the explanation, we assume that the number of child posterior beliefs is constant at each nonterminal belief and denoted by $n_z$.  Relaxing this assumption is straightforward.
The Bellman form  representation of \eqref{eq:Value} using such an estimator is  
%
\begin{equation}
\label{eq:ValueSampleGivenTree}
\begin{gathered}
\hat{V}(b_k, \pi_{k+}) \!   =  \! \frac{1}{n_z} \sum_{i=1}^{n_z} \rho(b_{k}, \pi_k(b_k), z^i_{k+1}, b^i_{k+1}) + \\
 +\gamma \frac{1}{n_z} \sum_{i=1}^{n_z} \hat{V}(\psi(b_k, \pi_k(b_k), z^i_{k+1}), \pi_{(k+1)+}), 	
\end{gathered}
\end{equation}
and the corresponding estimator for \eqref{eq:Qbellman} under an optimal future policy reads 
\begin{equation}
\label{eq:QbellmanSampleGivenTree}
\begin{gathered}
\hat{Q}(b_k, a_k)   = \frac{1}{n_z} \sum_{i=1}^{n_z} \rho(b_{k}, a_k, z^i_{k+1}, b^i_{k+1}) +   \\
+\gamma \frac{1}{n_z} \sum_{i=1}^{n_z}  \hat{V}(\psi(b_k, a_k, z^i_{k+1}), \pi^*_{(k+1)+}), 	
\end{gathered}
\end{equation}
where $n_z$ is the number of children of $b_{\ell}$ under the execution policy $\pi_{\ell+}$ and $i$ is the child index.  

\subsubsection{Interchangeability Between the history and Belief}
The purpose of this section is to clarify why further we will use interchangeably belief and the history. The belief is merely a reinterpretation of the knowledge about the POMDP state stored in history in the form of a PDF.  The belief $b_k$ is a function of the history $h_k$. Therefore different histories may yield the same belief. To avoid ambiguity and relate the objectives and their position in the belief tree with some abuse of notation we sometimes switch the dependence on the belief to dependence on corresponding history. In general we can write $b_{\ell}(h_{\ell})$.   

\subsubsection{Coupled Action-Value Function Estimation and Belief Tree Construction} \label{sec:CoupledTreeEstimator}

The estimator presented above  leverages symmetric in terms of observations Bellman form. However in MCTS methods due to exploration driven by, for example, UCB \eqref{eq:ucb}, the estimators are assembled from laces of the returns.  In each simulation a single lace is added to the estimator at each posterior belief. Whenever a new posterior belief node is expanded, a rollout is commenced such that the lace is complemented to the whole horizon. 

MCTS repetitively descends down the tree, adding a  lace of cumulative rewards (or updates visitation counts of an existing lace) and ascends back to root. On the way down it selects actions according to an exploration strategy e.g., \eqref{eq:ucb}.  This results in a policy tree, that represents a stochastic policy represented by visitation counts $\frac{N(ha)}{N(h)}$. Further we will focus on UCB exploration strategy, however all derivations of our approach are general and are valid for any exploration strategy, e.g, P-UCT \citep{Auger13Sp}  or $\epsilon$-greedy exploration \citep{Sutton18book}. 

A UCB-based MCTS over a Belief-MDP (BMDP) \citep{Auer02ml, Sunberg18icaps}  constructs a policy tree by executing multiple simulations. Each simulation adds a single belief node to the belief tree or terminates  by terminal state or action. To steer towards more deeper and more beneficial simulations, MCTS selects an action $a^{\ast}$ at each belief node according to the following rule $a^{\ast} = \maxim{\argmax }{a \in \mathcal{A}} \ \text{UCB} (ha)$ where 
\begin{align}
	&\text{UCB}(ha) = \hat{Q}(ha) + c\cdot \sqrt{\nicefrac{\log(N(h))}{N(ha)}},  \label{eq:ucb}
\end{align}
where $N(h)$ is the visitation count of the belief node defined by history $h$, $N(ha)$ is the visitation count of the belief-action node, $c$ is the exploration parameter and,  $\hat{Q}(ha)$ is the  estimator of the action-value function $Q$ for node $ha$ obtained by simulations. The rule described by \eqref{eq:ucb} is a result of modelling exploration as Multi Armed Bandit (MAB) problem \citep{Kocsis06ecml, Munos2014book, Auger13Sp}.  
When the action is selected, a question arises either to open a new branch in terms of observation and posterior belief or to continue through one of the existing branches. In continuous action,  and observation spaces, this can be resolved by the Double Progressive Widening (DPW) technique \citep{Sunberg18icaps, Auger13Sp}. If a new branch is expanded, an observation $o$ is created from state $x$ drawn from the belief $b$.

Let the return, corresponding to lace $i$ starting from some belief $b^i_{\ell}$ at depth $\ell - k$, be $g(b^i_{\ell}, a_{\ell})$   for $\ell \in [k:k+L-1]$. More specifically, suppose the new posterior belief was expanded at depth $d^i$ of the belief tree such that $d^i > \ell$. We have that  $g(b^i_{\ell}, a_{\ell})$ is composed from two parts, the already expanded tree part and the rollout added part such that
\begin{align}
\!\!\! g(b^i_{\ell}, a_{\ell}) {=}\! &\underbrace{\rho(b^i_{\ell}, a_{\ell}, b^i_{\ell+1}) {+} \!\!\!\!\! \sum_{l=\ell+1}^{k+d^i}\!\!\!\! \gamma^{l -\ell} \rho(b^i_{l}, \pi^{*,i}_{l}(b_{l}), b^i_{l+1})}_{\text{belief tree}}\! {+}\!\!\! \label{eq:BeliefTreeLace}\\
&+\underbrace{\sum_{l=k+d^i}^{k+L-1}  \gamma^{l -\ell} \rho(b^i_{l}, \mu(b^i_{l}), b^i_{l+1})}_{\text{rollout}},  \label{eq:Rollout}
\end{align} 
where $L$ is the horizon (tree depth), $\pi^{*,i}$ is an  optimal tree policy depending on the number of the simulation $i$ through $\hat{Q}$ and visitation counts in \eqref{eq:ucb} and  $\mu$ is the rollout policy. Importantly, in rollout the observations are drawn randomly and since we are in continuous spaces the beliefs in the rollouts are unique.  
A new belief node is added for $l = k+d^i$.  If due to DPW no new belief node was added to the belief tree, the rollout depicted by \eqref{eq:Rollout} and the return sample takes the form of 
\begin{align}
	&\!\!\! g(b^i_{\ell},a_{\ell}) {=} \rho(b^i_{\ell}, a_{\ell}, b^i_{\ell+1})  {+} \!\!\!\!\! \sum_{l=\ell+1}^{k+L-1}\!\!\!\! \gamma^{l -\ell} \rho(b^i_{l}, \pi^{*,i}_{l}(b_{l}), b^i_{l+1}).\!\!\!\!
\end{align} 
The estimate for \eqref{eq:Qbellman} under optimal future policy is assembled from  laces in accordance to 
\begin{align}
&\hat{Q}(h_{\ell}a_{\ell}) = \frac{1}{N(h_{\ell}a_{\ell})} \sum_{i=1}^{N(h_{\ell}a_{\ell})} g(b^i_{\ell}, a_{\ell}), \label{eq:SampleQMCTS}
\end{align}
where each reward $\rho(b,a,b')$ in the belief tree appears the number of times according to the visitation count of the node $b'$, namely $N(h')$.  
We note that for both estimators \eqref{eq:QbellmanSampleGivenTree} and \eqref{eq:SampleQMCTS}, the formulation  in \eqref{eq:dissectQSample} holds.

Now we move to the details of our general approach. 

\section{Our Approach}\label{sec:GeneralApproach}
This section is the core of our general approach. We first describe bounds over the theoretical and the estimated objectives. We then endow the rewards bounds with discrete simplification levels. Finally, instead of calculating rewards, we calculate the bounds over them  and if they are not tight enough we tighten them so we can make faster decisions with bounds over the objectives instead of objectives themselves.  

\subsection{Theoretical Simplification Formulation}
Simplification is any kind of relaxation of  POMDP tuple \eqref{eq:pomdp}  elements, accompanied by guarantees that quantify the (worst-case or potential) impact of a particular simplification technique on planning performance. In this section, we present a general simplification framework that is applicable to any reward bounds that satisfy the assumptions stated in Section~\ref{sec:Adaptive}. 

Our framework applies without any change to parametric and non-parametric beliefs, and to  closed-form belief-dependent rewards (that can be calculated exactly, i.e.~analytically), as well as to estimated rewards.  Therefore, in this paper we do not differentiate between these cases and denote the belief-dependent reward by $\rho(b_{\ell}, a_{\ell}, z_{\ell+1}, b_{\ell+1})$, without using the notation $\hat{\square}$ for estimators. In other words, depending on the setting, $\rho()$ and $b_{\ell}$ can represent, respectively, an analytical reward and a parametric belief, or a reward estimator and a nonparametric belief. In all cases, if one can provide monotonically adaptive bounds on the reward, our framework will return an identical solution as if the decision making was performed with original reward calculations (i.e.~depending on the setting, either an analytical reward calculation or reward estimator calculation). In Section~\ref{sec:SpecificSimplification} we provide a specific incarnation of the framework considering non-parametric beliefs represented by a set of weighted samples and a reward estimator, and where the simplification corresponds to utilizing only a subset of the samples.


As mentioned, we aim to simplify the belief-dependent reward $\rho(b_{\ell}, a_{\ell}, z_{\ell+1}, b_{\ell+1})$ calculations. Namely, the original reward $\rho$ is bounded using the simplified belief $b^s$ instead of original belief $b$. This operation materializes in the form of following inequality  
\begin{equation}
\label{eq:BoundRewardPairConsecutive}
\begin{gathered} 
\underline{\rho}(b_{\ell}^s, b_{\ell},  a_{\ell}, z_{\ell+1}, b_{\ell+1}, b_{\ell+1}^s)  \leq	 \\
\leq \rho(b_{\ell}, a_{\ell}, z_{\ell+1}, b_{\ell+1})    \leq    \\
\leq \overline{\rho}(b_{\ell}^s, b_{\ell}, a_{\ell}, z_{\ell+1}, b_{\ell+1}, b_{\ell+1}^s),  
\end{gathered}
\end{equation}
where $\underline{\rho}$ and $\overline{\rho}$ are the corresponding lower and upper bounds, respectively. The superscript $s$ denotes the fact that the corresponding belief was simplified as we depict in Fig.~\ref{fig:BeliefTreeInplaceSimp}. Notice that in   \eqref{eq:BoundRewardPairConsecutive} the pair of consecutive beliefs, $b_{\ell}$ and $b_{\ell+1}$, can be simplified differently.

Henceforth, in order to avoid unnecessary clutter we will omit the dependence on the observation and  denote the bounds over the reward using simplified beliefs as follows    
\begin{equation}\label{eq:BoundRewardGeneral}
	\underline{\rho}^s(b,a, b') \leq \rho(b,a, b') \leq \overline{\rho}^{s}(b,a, b').
\end{equation}
It should be stressed that since in the  belief tree $b'$ always has a single parent $b$, the reader should think about such a reward as one corresponding to $b'$. 

A key requirement is reduced computational complexity of these bounds compared to the complexity of the original reward.
Instead of calculating the expensive reward $\rho(b,a,b')$ for each pair of beliefs $b,b'$, we first obtain the corresponding simplified beliefs $b^s$ and $b'^{s}$, as illustrated in Fig.~\ref{fig:BeliefTreeInplaceSimp}, and then formulate the bounds $\underline{\rho}^s$ and $\overline{\rho}^s$ from \eqref{eq:BoundRewardGeneral}. However, we note that the form \eqref{eq:BoundRewardGeneral} is actually more general and not limited to belief simplification. 

Further we formulate bounds over the value function and action-value function, both under the optimal policy. In fact, our bounds hold under an  arbitrary policy. We narrow the discussion to optimal polices solely for the clarity of the explanation and this is not a limitation of our approach.

Suppose inequality \eqref{eq:BoundRewardGeneral} holds for any possible pair of consecutive beliefs, e.g.~these are analytical bounds,  as opposed to \citep{Zhitnikov22ai}. A direct consequence of this fact, alongside the structure of \eqref{eq:Value}, is that
\begin{equation}\label{eq:BoundsValueGeneral}
\underline{V}(b_{\ell}, \pi^*_{\ell+}) \leq V(b_{\ell}, \pi^*_{\ell+}) \leq 	\overline{V}(b_{\ell},\pi^*_{\ell+}),
\end{equation}
holds for any belief $b_{\ell}$ and $ \ell \in[k,k+L-1]$.  Using the Bellman representation as in \eqref{eq:Vbellman} the bounds \eqref{eq:BoundsValueGeneral} take the form of 
\begin{equation}\label{eq:RecursiveBoundsTheoretical}
\begin{split}
&\!\!\!\!	\overline{V}\!(b_{\ell}, \pi^*_{\ell+}) {=\!\!\!}\underset{z_{\ell+1}}{\mathbb{E}}\! \Big[ \overline{\rho}^s(b_{\ell},\! \pi^*_{\ell}(b_{\ell}), b^i_{\ell+1}) \! + \!\!   \overline{V}(b^i_{{\ell}+1},\pi^*_{(\ell+1)+})\! \Big] \!\!\!\!\!\!\! \\
&\!\!\!\!	\underline{V}\!(b_{\ell}, \pi^*_{\ell+}) {=\!\!\!} \underset{z_{\ell+1}}{\mathbb{E}}\! \Big[ \underline{\rho}^s(b_{\ell},\! \pi^*_{\ell}(b_{\ell}), b^i_{\ell+1}) \! + \!\!   \underline{V}(b^i_{{\ell}+1},\pi^*_{(\ell+1)+})\! \Big].	 \!\!\!\!\!\!\!
\end{split}
\end{equation}
The bounds over the value function in \eqref{eq:RecursiveBoundsTheoretical} are initialized at the $L$th time step in the planning horizon as $\overline{V}(b_{k+L}, \pi_{k+L})= 0$ and $\underline{V}(b_{k+L}, \pi_{k+L})= 0$.
Similarly the bounds over the action-value function \eqref{eq:Qbellman} under an optimal future policy are 
\begin{equation}\label{eq:BoundActionValueGeneral}
\!\!\! \underline{Q}\Big(\!b_{\ell},\! \{a_{\ell}, \pi^{*}_{(\ell+1)+}\}\! \Big) {\leq } Q\big(b_{\ell},a_{\ell}\big) {\leq} \overline{Q}\Big(\! b_{\ell}, \!\{a_{\ell}, \pi^{*}_{(\ell+1)+}\}\! \Big),\!\!\!\!
\end{equation}
where the policy $\pi^{\ast}_{(\ell+1)+}$ is optimal.
Note, as we observe in  \eqref{eq:RecursiveBoundsTheoretical},
the simplification assumed herein does not affect the distribution of future observations with respect to which the expectation is taken.

\subsubsection*{Bounding the Belief Dependent Element of the Reward}\label{sec:QDecomp}
At this point, we want to recall that commonly, the state-dependent element \eqref{eq:StateReward}  is much easier to calculate than the belief dependent one.
Leveraging the structure manifested by \eqref{eq:dissectQ} the immediate bounds over \eqref{eq:InformationReward} induce bounds over $Q^{I}(\cdot)$  as such
\begin{align}
	\underline{Q}^I(b_k,a_k) \leq Q^I(b_k,a_k) \leq \overline{Q}^I(b_k,a_k), \label{eq:Qibounds}
\end{align}
and utilizing  \eqref{eq:dissectQ} we arrive at 
\begin{align}
	&\overline{Q}(b_k, a_k) = (1-\lambda)Q^{x}(b_k, a_k) + \lambda \overline{Q}^{I}(b_k, a_k) \label{eq:IinducedUpper}\\
	&\underline{Q}(b_k, a_k) = (1-\lambda)Q^{x}(b_k, a_k) + \lambda \underline{Q}^{I}(b_k, a_k). \label{eq:IinducedLower}
\end{align}
Importantly, the belief dependent element \eqref{eq:InformationReward} does not have to be information-theoretic. The simplification paradigm is general and works for any belief-dependent operator given appropriate bounds.

\subsection{Bounds over the Estimated Objective}
As we explained in section \ref{sec:EstObj} in practice the value and action-value function are estimated. Instead of using 
\eqref{eq:BoundsValueGeneral} and \eqref{eq:BoundActionValueGeneral} we have  
\begin{equation}\label{eq:BoundValueSampleGeneral}
	\!\!\! \underline{\hat{V}}\big(b_{\ell},\!  \pi^{*}_{\ell+}\big) \!\leq \! \hat{V}\big(b_{\ell}, \pi^{*}_{\ell+} \big) \!\leq\! \overline{\hat{V}}\big(b_{\ell}, \! \pi^{*}_{\ell+}\big),\!\!\!\!
\end{equation}
and
\begin{equation}\label{eq:BoundActionValueSampleGeneral}
	\!\!\! \underline{\hat{Q}}\Big(\! b_{\ell},\! \{a_{\ell}, \pi^{*}_{(\ell+1)+}\}\! \Big) {\leq } \hat{Q}\big(b_{\ell},a_{\ell}\big) {\leq} \overline{\hat{Q}}\Big(\! b_{\ell}, \!\{a_{\ell}, \pi^{*}_{(\ell+1)+}\}\! \Big),\!\!\!\!
\end{equation}
respectively. 

The bounds, in case of symmetric estimators from section \ref{sec:GivenTreeEstimator}, are 
\begin{equation}\label{eq:SampleVboundsBellmanGivenTree}
	\begin{gathered}
		\!\!\!\! \overline{\hat{V}}(b_{\ell}, \pi^{\ast}_{\ell+}) \!=\! \frac{1}{n_z}\sum^{n_z}_{i=1} \overline{\rho}^s(b_{\ell},\! \pi^*(b_{\ell}),b^i_{\ell+1} ) +\\ 
		    +\gamma \frac{1}{n_z}\sum^{n_z}_{i=1}\overline{\hat{V}}(b^i_{\ell+1}, \! \pi^{\ast}_{(\ell+1)+})\!\!\! \!\!\!\!\!\!\!\!\!\!\\
		\!\!\!\! \underline{\hat{V}}(b_{\ell},  \pi^{\ast}_{\ell+}) \!=\! \frac{1}{n_z}\sum^{n_z}_{i=1} \underline{\rho}^s(b_{\ell},\! \pi^*(b_{\ell}),b^i_{\ell+1} )   + \\
		+\gamma \frac{1}{n_z}\sum^{n_z}_{i=1}  \underline{\hat{V}}(b^i_{\ell+1}, \! \pi^{\ast}_{(\ell+1)+}),\!\!\! \!\!\!\!\!\!\!\!\!\!
	\end{gathered}
\end{equation}
where, to clarify we repeat that  $n_z$ is the number of children of $b_{\ell}$ under the execution policy $\pi_{\ell+}$ and $i$ is the child index. 
The bounds over the estimated value function in \eqref{eq:SampleVboundsBellmanGivenTree} are initialized at the $L$th time step in the planning horizon as $\overline{\hat{V}}(b_{k+L}, \pi_{k+L})= 0$ and $\underline{\hat{V}}(b_{k+L}, \pi_{k+L})= 0$. 

In a similar manner we define also bounds over  \eqref{eq:QbellmanSampleGivenTree} as such 
\begin{equation}\label{eq:SampleQboundsBellmanGivenTree}
	\begin{gathered}
		\!\!\!\! \overline{\hat{Q}}(b_{\ell}, \{a_{\ell},\! \pi^{\ast}_{(\ell+1)+}\}) \!=\! \frac{1}{n_z}\sum^{n_z}_{i=1}\overline{\rho}^s(b_{\ell},\! a_{\ell},b^i_{\ell+1} )  + \\
		 +\gamma \frac{1}{n_z}\sum^{n_z}_{i=1}   \overline{\hat{V}}(b^i_{\ell+1}, \! \pi^{\ast}_{(\ell+1)+})\!\!\! \!\!\!\!\!\!\!\!\!\!\\
		\!\!\!\! \underline{\hat{Q}}(b_{\ell}, \{a_{\ell},\! \pi^{\ast}_{(\ell+1)+}\}) \!=\! \frac{1}{n_z}\sum^{n_z}_{i=1}\underline{\rho}^s(b_{\ell},\! a_{\ell},b^i_{\ell+1} ) + \\
		+\gamma \frac{1}{n_z}\sum^{n_z}_{i=1} \underline{\hat{V}}(b^i_{\ell+1}, \! \pi^{\ast}_{(\ell+1)+}),\!\!\! \!\!\!\!\!\!\!\!\!\!
	\end{gathered}
\end{equation}
We emphasize that the superscript $i$ in \eqref{eq:SampleVboundsBellmanGivenTree} and \eqref{eq:SampleQboundsBellmanGivenTree} denotes the child posterior nodes of $b_{\ell}$. 

The bounds over MCTS estimator \eqref{eq:SampleQMCTS} are 
\begin{equation}\label{eq:SampleQboundsBellmanMCTS}
	\begin{gathered}
		\!\!\overline{\hat{Q}}(ha)\! = \! \frac{1}{N(ha)} \!\!\! \sum_{i=1}^{N(ha)} \!\!\!\Big(\overline{\rho}^s(b^i_{\ell}, a_{\ell}, b^i_{\ell+1})+ \\
		{+}\!\!\!\!\!\sum_{l=\ell+1}^{k+d^i}\!\!\! \gamma^{l -\ell} \overline{\rho}^s(b^i_{l}, \pi^{*,i}_{l}(b_{l}), b^i_{l+1})\! + \!\!\!\!\!\!
		\sum_{l=k+d^i}^{k+L-1} \!\!\!\!\!  \gamma^{l -\ell} \overline{\rho}^s(b^i_{l}, \mu(b^i_{l}), b^i_{l+1})\!\Big) \\
		\!\!\underline{\hat{Q}}(ha) \!= \! \frac{1}{N(ha)} \!\!\! \sum_{i=1}^{N(ha)}\Big(\underline{\rho}^s(b^i_{\ell}, a_{\ell}, b^i_{\ell+1})  + \\
		{+}\!\!\!\!\sum_{l=\ell+1}^{k+d^i} \!\!\! \gamma^{l-\ell} \underline{\rho}^s(b^i_{l}, \pi^{*,i}_{l}(b_{l}), b^i_{l+1}){+\!\!\!\!\!} \sum_{l=k+d^i}^{k+L-1} \!\!\!\!\!\! \gamma^{l -\ell} \underline{\rho}^s(b^i_{l}, \mu(b^i_{l}), b^i_{l+1})\!\Big).
	\end{gathered}
\end{equation}
Let us clarify again that in \eqref{eq:SampleQboundsBellmanMCTS} the superscript $i$ denotes the number of the simulation. Moreover, the reward bounds within the tree repeat in more than a single simulation according to the visitation count of the corresponding posterior belief. 
\begin{figure}[t]
	\centering         
	\includegraphics[width=0.4\textwidth]{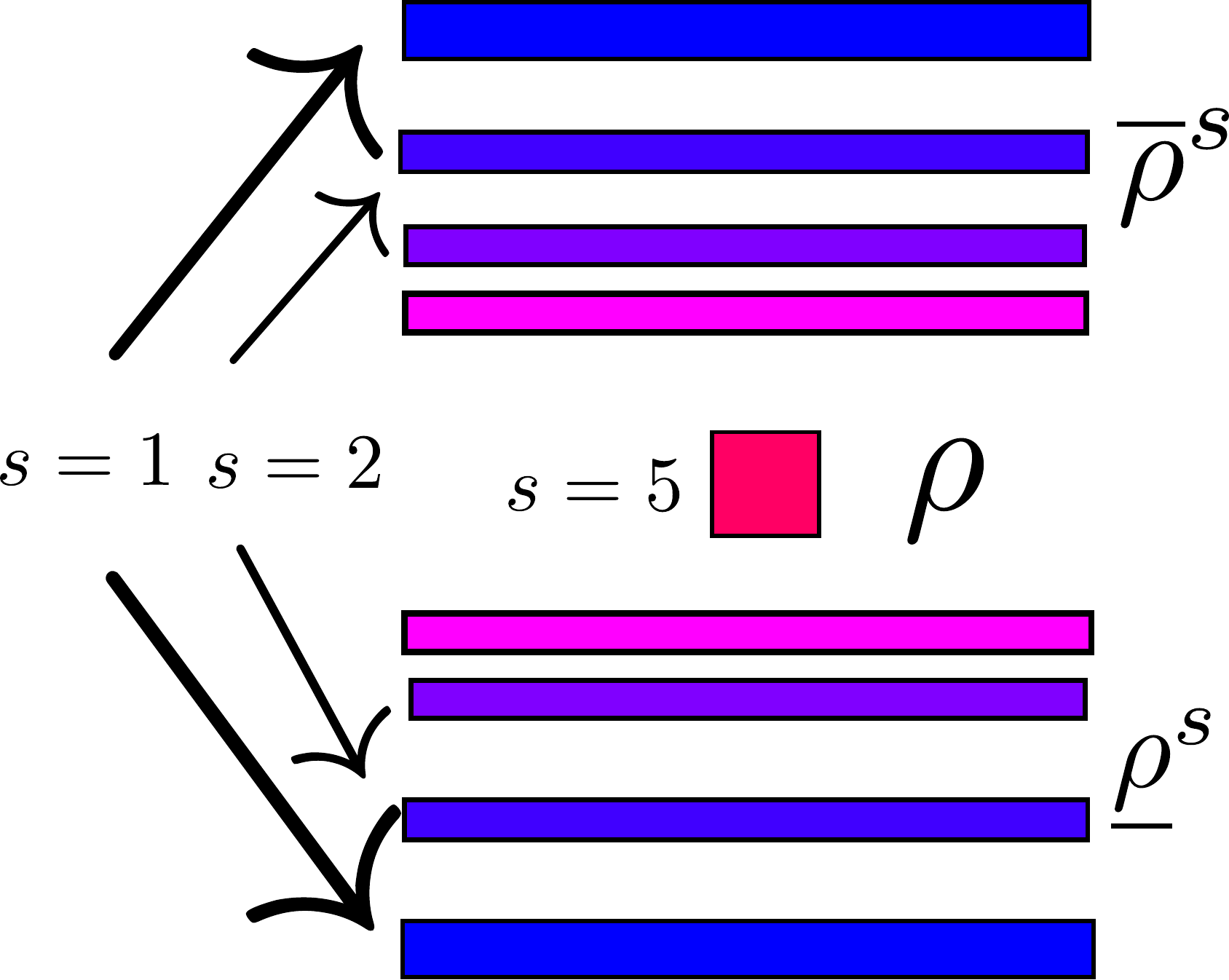}
	\caption{Reward bounds and different levels of the simplification. Here $n_{\mathrm{max}} = 5$. Warmer colors visualize tighter bounds. Whereas colder colors (blue) indicate looser bounds and cheaper to calculate.  }%
	\label{fig:BoundsReward}%
\end{figure}
Clearly, the decomposition displayed by Eq. \eqref{eq:IinducedUpper} and \eqref{eq:IinducedLower} is valid for both bounds \eqref{eq:SampleQboundsBellmanGivenTree} and \eqref{eq:SampleQboundsBellmanMCTS}. 
We have that
\begin{align}
	&\overline{\hat{Q}}(b_k, a_k) = (1-\lambda)\hat{Q}^{x}(b_k, a_k) + \lambda \overline{\hat{Q}}^{I}(b_k, a_k) \label{eq:IinducedUpperEstimated}\\
	&\underline{\hat{Q}}(b_k, a_k) = (1-\lambda)\hat{Q}^{x}(b_k, a_k) + \lambda \underline{\hat{Q}}^{I}(b_k, a_k). \label{eq:IinducedLowerEstimated}
\end{align}
\begin{figure*}[t]
	\begin{minipage}[t]{0.30\textwidth}   
		\centering         
		\includegraphics[width=\columnwidth]{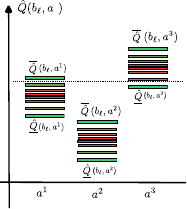}
		\subcaption{}
		\label{fig:ActionGraph}
	\end{minipage}
	\hfill
	\begin{minipage}[t]{0.30\textwidth}
		\centering
		\includegraphics[width=\columnwidth]{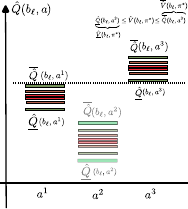}
		\subcaption{}	
		\label{fig:ActionGraphAdapted}	
	\end{minipage}
	\hfill
	\begin{minipage}[t]{0.30\textwidth}
		\centering
		\includegraphics[width=\columnwidth]{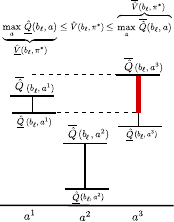}
		\subcaption{}	
		\label{fig:ValueGraph}	
	\end{minipage}
	\caption{In this illustration we have three candidate actions $\{a^1, a^2, a^3\}$ that can possibly be taken by the robot from the belief node $b_{\ell}.  $\textbf{(a)} We observe that $\overline{\hat{Q}}(b_{\ell}, a^1_{\ell}) > \underline{\hat{Q}}(b_{\ell}, a^3_{\ell})$  and prune action $a^2$. \textbf{(b)} After the resimplification no overlap an we can safely decide that $a^3_{\ell}$ is optimal. Moreover we prune the withered interval corresponding to the $a^2_{\ell}$. \textbf{(c)} Another situation where we are not concerned about optimal action, we solely want to send up to the tree the bounds over optimal value function.  }
	\label{fig:ConceptualDiagramGivenTree}
\end{figure*}  
\subsubsection*{Impact of the Information Weight $\lambda$} 
Allow us to linger on the $\lambda$  from eq.~\ref{eq:dissectQ} and \ref{eq:dissectQSample}. It is hard to predict how the objective will behave with various values of $\lambda$. Nevertheless, if the bounds are over the belief-dependent element of the reward, by subtracting \eqref{eq:IinducedLowerEstimated} from \eqref{eq:IinducedUpperEstimated}, we arrive at 
\begin{align}
&\overline{\hat{Q}}(b_k, a_k)\! - \! \underline{\hat{Q}}(b_k, a_k) {=}  \lambda \Big( \overline{\hat{Q}}^{I}(b_k, a_k)\! - \! \underline{\hat{Q}}^{I}(b_k, a_k) \Big).
\end{align}
The width of the bounds is monotonically increasing with $\lambda$. Of course, it will also happen to a theoretical analog of such a bounds displayed by eq.~\eqref{eq:IinducedUpper} and \eqref{eq:IinducedLower}.  We can envision more speedup from applying the simplification paradigm with lower values of $\lambda$ and will see it in the simulations.     

Further we will consider the estimated action-value or value functions and therefore omit the word ``estimated''.  We will also omit mentioning each time that our bounds are under the optimal policy.

\subsection{Multi-Level Simplification}\label{sec:Adaptive}
We now extend the definition of simplification as we envision it to be an \emph{adaptive paradigm}. We denote \textit{level of simplification} as how ``aggressive'' the suggested simplification is. 
Observe an illustration in Fig.~\ref{fig:BoundsReward}. 

With this setting, we can naturally define many discrete levels such that $s \in \{1, 2, \dots, n_{\mathrm{max}}\} $ represents the simplification level, where $1$ and $n_{\mathrm{max}}$ correspond to the coarsest and finest simplification levels, respectively. For instance, suppose the belief is represented by a set of samples (particles), as in Section \ref{sec:SpecificSimplification}. Taking a small subset of particles to represent the simplified belief corresponds to a \emph{coarse} simplification. If one takes many particles, this corresponds to a \emph{fine} simplification.

{\bf Remark:} From now on the superscript $s$ denotes the discrete simplification level. Importantly we always have a {\bf finite} number, denoted by $n_{\mathrm{max}}$, of simplification levels. 

Further, we assume bounds monotonically become tighter as the simplification level is increased and that the bounds for the finest simplification level $n_{\mathrm{max}}$ converge to the original reward without simplification. More formally, denote
$\overline{\Delta}^s(b,a, b') \triangleq \overline{\rho}^s(b, a,b') - \rho(b,a,b')$ and $\underline{\Delta}^s(b,a, b') \triangleq \rho(b,a,b') - \underline{\rho}^s(b, a,b')$. 
\begin{assumption}[Monotonicity]
	\label{assumption:monotonic} Let $n_{\mathrm{max}} \geq 2$, 
	$\forall s \in [1,n_{\mathrm{max}}-1]$ we get: $	\overline{\Delta}^s(b,a,b') \geq \overline{\Delta}^{s+1}(b,a,b')$ and $\underline{\Delta}^{s}(b,a,b') \geq \underline{\Delta}^{s+1}(b,a,b').$
\end{assumption}
%
\begin{assumption}[Convergence]
	\label{assumption:convergence}
	$\forall b,a,b'$ we get: $\overline{\rho}^{s = n_{\mathrm{max}}}(b,a, b') = \underline{\rho}^{s = n_{\mathrm{max}}}(b,a,b')= \rho(b,a,b').$
\end{assumption}
In Section  \ref{sec:SpecificSimplification}, we derive novel bounds on top of a particular simplification that takes a subset of belief samples instead of a complete set.  We prove that these bounds indeed satisfy both assumptions.

The simplification levels of the reward bounds for different posterior belief nodes in the belief tree  determine how tight the bounds over the value or action-value function are.  To tighten the bounds over the objective, we have the freedom to select any rewards the belief tree and tighten the bounds over these selected rewards by increasing their simplification levels; this, in turn, would  contract the bounds over the objective.

We call a particular algorithmic scheme to select the rewards a {\bf resimplification strategy}.  
A general valid resimplificaiton strategy is defined as follows.

\begin{defn}[Resimplification strategy]\label{def:ResimplificationStrategy}
	Given a pair of  lower $\underline{\hat{V}}(b_{\ell}, \pi_{\ell+}) $ ($ \underline{\hat{Q}}(b_{\ell},\! \{a_{\ell}, \pi^{*}_{(\ell+1)+}\})$) and upper bounds $\overline{\hat{V}}(b_{\ell}, \pi_{\ell+}) $ ($\overline{\hat{Q}}(b_{\ell},\! \{a_{\ell}, \pi^{*}_{(\ell+1)+}\})$) over the estimated objective, the resimplification strategy is a rule to promote one or more simplification levels of the rewards  in the the subtree rooted at $b_{\ell}$ and defined by the mentioned above estimated objective. If the resimplification does not promote the simplification level for any reward, so   $\overline{\hat{Q}}(b_{\ell},\! \{a_{\ell}, \pi^{*}_{(\ell+1)+}\}) - \underline{\hat{Q}}(b_{\ell},\! \{a_{\ell}, \pi^{*}_{(\ell+1)+}\})=0$.
\end{defn}

Note that, all the rewards within a subtree defined by  $\overline{\hat{Q}}(b_{\ell},\! \{a_{\ell}$, $\pi^{*}_{(\ell+1)+}\}), \underline{\hat{Q}}(b_{\ell},\! \{a_{\ell}, \pi^{*}_{(\ell+1)+}\})$ are being at the maximal simplification level  implies $\overline{\hat{Q}}(b_{\ell},\! \{a_{\ell}, \pi^{*}_{(\ell+1)+}\}) - \underline{\hat{Q}}(b_{\ell},\! \{a_{\ell}, \pi^{*}_{(\ell+1)+}\})=0$, but the inverse implication is not necessarily true. Once initiated, a {\bf valid} strategy can select no reward for simplification level promotion only if  $\overline{\hat{Q}}(b_{\ell},\! \{a_{\ell}, \pi^{*}_{(\ell+1)+}\}) - \underline{\hat{Q}}(b_{\ell},\! \{a_{\ell}, \pi^{*}_{(\ell+1)+}\})=0$. 
\begin{thm}[Monotonicity and Convergence of Estimated Objective Function Bounds]\label{thm:MonotonConvQ}
	If the bounds over the reward are monotonic (assumption \ref{assumption:monotonic}) and convergent (assumption \ref{assumption:convergence}), for both estimators  \eqref{eq:SampleQboundsBellmanGivenTree}  and \eqref{eq:SampleQboundsBellmanMCTS}, the bounds on the sample  approximation \eqref{eq:BoundActionValueSampleGeneral} are monotonic as a function of the number of resimplifications and convergent after at most $n_{{\mathrm{max}}} \cdot M$ resimplifications for {\bf any}  resimplification strategy. Here $M$ is the number of posterior beliefs  in  \eqref{eq:SampleQboundsBellmanGivenTree}  or \eqref{eq:SampleQboundsBellmanMCTS} .  Namely, if all the rewards are at the maximal simplifciation level $n_{\mathrm{max}}$ we have to reach
	\begin{align}
		\underline{\hat{Q}}(\cdot)  =  \hat{Q}(\cdot) = \overline{\hat{Q}}(\cdot). \label{sec:zerogapConv}
	\end{align}
	Similarly for Optimal value function  the equality	$\underline{\hat{V}}(\cdot)  =  \hat{V}(\cdot) = \overline{\hat{V}}(\cdot)$ holds.
\end{thm}%
\noindent 
The reader can find the proof in the Appendix~\ref{proof:MonotonConvQ}.
Theorem \ref{thm:MonotonConvQ} ensures that if the resimplification strategy is valid (Definition~\ref{def:ResimplificationStrategy}), we do not get stuck in an infinite loop of resimplifications if instead of $\hat{Q}(\cdot)$ we use its bounds. In particular, if \eqref{sec:zerogapConv} is reached, there is no reason to activate the resimplification routine. 

Importantly, as we discuss next and corroborate by simulations in many cases we can identify the optimal action before reaching the maximal number of resimplificaitons.

\subsection{Adaptive Simplification Mechanics} \label{sec:GeneralSimplificationMechanics}
Our adaptive simplification approach is based on two key observations.
The \emph{first key observation} is that we can compare bounds over \eqref{eq:BoundActionValueSampleGeneral} constituted by rewards at different levels of simplification. Our \emph{second key observation} is that we can reuse calculations between different simplification levels avoiding recalculation of the simplification from scratch.

Naturally we do not want  to reach \eqref{sec:zerogapConv}. Let us  begin by explaining how we determine an optimal action by using  bounds over the action-value function instead of  its explicit calculation  and obtain a significant speedup in planning  time. 
If there is no overlap between the intervals originated from the upper and lower bounds \eqref{eq:BoundActionValueSampleGeneral}  of each candidate action,  we can determine the optimal action and therefore  there is no reason to call the resimplification routine.  

Contemplate about some belief $b_{\ell}$ in the belief tree. We annotate by superscript $j$ candidate actions emanating from $b_\ell$, such that the index $j$ corresponds to the $j$th candidate action.  We first select a candidate action using the lower bound \eqref{eq:BoundActionValueSampleGeneral} over $\hat{Q}\big(b_{\ell},a^j_{\ell}\big)$ as 
\begin{align}
	&\!j^{\dagger}(b_{\ell}(h_{\ell})) {=} \argmax_{j}  \Big\{ \underline{\!\hat{Q}}(b_{\ell}(h_{\ell}),\! \{a^j_{\ell}, \pi^{\ast}_{({\ell}+1)+}\}){+}c^j(h_{\ell}a^j) \Big\},\label{eq:OptAction}
\end{align}
where $c^j$ is an action dependent constant. In case of a given belief tree $c^j = 0 \ \forall j$, whereas in case of MCTS, it is a constant originated from UCB as in \eqref{eq:ucb}. 

We then ask the question whether or not an overlap with another candidate action exists,  
\begin{equation}
\begin{gathered}\label{eq:AdaptCond}
	\!\underline{\hat{Q}}(b_{\ell},\!\{a^{j^{\dagger}}_{\ell} ,  \pi^{\ast}_{({\ell}+1)+}\} )\! + c^{j^{\dagger}} \overbrace{\geq}^{?}  \\
	\geq \max_{j \in \{1 \dots \} \setminus  \{ j^{\dagger}\}}\!\!\Big\{ \overline{\hat{Q}}(b_{\ell},\!\{a^j_{\ell}, \! \pi^{\ast}_{({\ell}+1)+}\} )+c^j\Big\}
\end{gathered}
\end{equation}
See a visualization in Fig.~\ref{fig:ActionGraph}.

\begin{figure}[t]
	\centering         
	\includegraphics[width=0.4\textwidth]{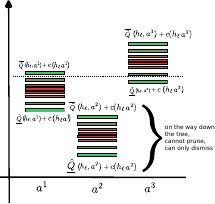}
	\caption{Demonstration of our approach in the setting of MCTS. In contrast to Fig.~\ref{fig:ActionGraphAdapted}, we cannot prune action $a^2$ and can only dismiss it to not participate in resimplifications. This is because, in the next tree queries, $a^2$ may be the best action for the robot to take. }%
	\label{fig:Dismissal}%
\end{figure}

If the  condition displayed by equation \eqref{eq:AdaptCond} is not fulfilled, as depicted in Fig.~\ref{fig:ActionGraph}, we shall tighten the bounds \eqref{eq:BoundActionValueSampleGeneral} by calling a {\bf resimplification strategy} . Importantly, in case of a given belief tree,  even if an overlap is present similar to branch-and-bound technique \citep{Kochenderfer22book} we can prune any subtree obtained with action $j$ satisfying  
\begin{align} 
	\!\underline{\hat{Q}}(b_{\ell},\!\{a^{j^{\dagger}}_{\ell} \!\!, \! \pi^{\ast}_{({\ell}+1)+}\} ) {+} c^{j^{\dagger}} {\geq}  \overline{\hat{Q}}(b_{\ell},\!\{a^j_{\ell}, \! \pi^{\ast}_{({\ell}+1)+}\} ) {+} c^j. \label{eq:DismissingGeneral}
\end{align}
We illustrated this aspect in Fig.~\ref{fig:ActionGraphAdapted}.  If the belief tree is constructed gradually as in MCTS based methods and anytime setting, instead of pruning, we still can use \eqref{eq:DismissingGeneral} to dismiss suboptimal, at current simulation of MCTS, actions (See Fig.~\ref{fig:Dismissal}). 

Once no overlap is present (the condition \eqref{eq:AdaptCond} is fulfilled)  we can declare that the selected action is optimal ($\pi^{\ast}_{\ell}(b_\ell)=a_{\ell}^{j^{\dagger}(b_{\ell})}$).  
Utilizing the optimal action we can bound  the \emph{optimal} value function $\hat{V}(b_{\ell}, \pi^{\ast}_{\ell+})$ as such 
\begin{align}
	&\overline{\hat{V}}(b_{\ell}, \{ \pi^{\ast}_{\ell}, \pi^{\ast}_{(\ell+1)+} \}) \triangleq \overline{\hat{Q}}(b_{\ell},\{a^{j^{\dagger}(b_{\ell})}_{\ell} , \pi^{\ast}_{({\ell}+1)+}\} ), \label{eq:LowerValue}\\
	& \underline{\hat{V}}(b_{\ell},\{ \pi^{\ast}_{\ell}, \pi^{\ast}_{(\ell+1)+})\} ) \triangleq \underline{\hat{Q}}(b_{\ell},\{a^{j^{\dagger}(b_{\ell})}_{\ell} , \pi^{\ast}_{({\ell}+1)+}\} ).\label{eq:UpperValue}
\end{align} 
Let us recite that the bounds \eqref{eq:LowerValue} and \eqref{eq:UpperValue} are conditioned on the fact that there is no overlap of the bounds intervals that correspond to different candidate actions, namely the condition \eqref{eq:AdaptCond} is met for \emph{each} belief $b_{\ell}$ in the belief tree. This situation is visualized in Fig.~\ref{fig:ActionGraphAdapted}.  

On the other hand, to identify the optimal immediate action $a_k^{*}$,  we require no overlap between bounds of different actions only at the root of the  belief tree (where the belief is $b_k$). This means that at each belief node $b_{\ell}$ in the tree, besides the root,  we only want to bound the value function for the optimal action (and under optimal future policy). While it is possible to do so by first determining the optimal action, as in \eqref{eq:LowerValue}  and \eqref{eq:UpperValue}, we can bypass this step and directly bound the value function over the optimal action as follows,
\begin{align}
	&\overline{\hat{V}}(b_{\ell}, \{ \pi^{\ast}_{\ell}, \pi^{\ast}_{(\ell+1)+} \}) \triangleq \max_j \overline{\hat{Q}}(b_{\ell},\{a^{j}, \pi^{\ast}_{({\ell}+1)+}\} ), \label{eq:LazyLowerValue}\\
	& \underline{\hat{V}}(b_{\ell},\{ \pi^{\ast}_{\ell}, \pi^{\ast}_{(\ell+1)+})\} ) \triangleq \max_j \underline{\hat{Q}}(b_{\ell},\{a^{j} , \pi^{\ast}_{({\ell}+1)+}\} ), \label{eq:LazyUpperValue}
\end{align}
i.e.~relaxing the requirement of no overlap between bounds for different actions  at any node  $b_{\ell}$ besides $b_k$. See illustration of \eqref{eq:LazyLowerValue} and \eqref{eq:LazyUpperValue} in Fig.~\ref{fig:ValueGraph}.  In turn, eliminating a single overlap at the root results in lower rewards simplification levels in the tree, although such a value bounds may be looser. As we shall see, this approach would typically yield more speedup.

Nevertheless, when we need a policy tree we still have to obtain an optimal action at each belief node within the tree. This requires no bounds overlap at each node, as in the former setting. This situation arises for example when the action and observation spaces are large but discrete. In this case the robot sometimes does not do re-planning at each time step. Instead the robot uses the policy tree as a representation of the policy and  selects an optimal action that corresponds to the received observation.  In addition, such a strategy  accommodates possible reuse calculations in such a solved belief tree \citep{Farhi19icra, Farhi21arxiv}.

To conclude this section let us summarize. As discussed,  we have the following two  variants:
\begin{itemize}
	\item The resimplification is initiated at each nonterminal posterior belief node $b_{\ell}$ up until no overlap between candidate actions is present and the optimal action $\pi^{\ast}_{\ell}(b_\ell)$ is selected. This way we bound the optimal value function of the descendant to $b_k$ nodes using an optimal action according to \eqref{eq:LowerValue} and \eqref{eq:UpperValue}. We named this approach Policy Tree (\texttt{PT}).
	\item The resimplification is commenced solely at the root $b_k$ of the whole belief tree. We eliminate the overlap and obtain an optimal action only at $b_k$.  This way we use  \eqref{eq:LazyLowerValue} and \eqref{eq:LazyUpperValue} to bound the optimal value function of the descendant to $b_k$ nodes.  We shall refer to this variant of our approach as \texttt{LAZY}. 
\end{itemize}

\subsection{Specific Resimplification Strategies}\label{sec:SpecifcStrategies}

In this paper we consider two specific resimplification strategies that are elaborated in the next sections: \texttt{Simplification Level} (\texttt{SL}) and \texttt{Gap}.  
We note that additional valid resimplification strategies exist and can be plugged-in into the above-proposed  general theory.  

{\bf Simplification level:} The resimplification strategy can be directly tied to the simplification level. 
In this situation the resimplifcation strategy promotes simplification level of the rewards inside the belief tree corresponding to bounds in \eqref{eq:BoundValueSampleGeneral} or \eqref{eq:BoundActionValueSampleGeneral} based on the simplification level itself. We provide further details in the setting of a given belief tree, considering a \texttt{PT} variant in Section \ref{sec:GivenTree}.

{\bf Gap:} Another possibility is that the resimplification is tied to the gap $\overline{\rho}^s - \underline{\rho}^s$. Such a resimplification promotes the simplification level if the reward bounds gap satisfies a certain condition.  We describe thoroughly this resimplification flavor  in the setting of a given belief tree, considering \texttt{LAZY} variant in Section \ref{sec:LazyVariantGivenTree}, and in MCTS setting, considering a \texttt{PT} variant in Section \ref{sec:resimplificationMCTS}.

Each of these strategies can be used in conjunction with any of the variants \texttt{PT} and \texttt{LAZY}. In the sequel,  we shall denote these combinations explicitly, e.g. \texttt{PT-SL}, \texttt{LAZY-Gap} and \texttt{PT-Gap}.

The preceding discussion raises the question of how do we actually incorporate the proposed bounds into online decision making. This brings us to the next section.
We first consider a given belief tree and then coupled belief tree construction and solution as in MCTS methods.  It shall be noted that further presented resimplification strategies are also suitable for static candidate action sequences, with minor modifications.

\section{Adaptive Simplification in the Setting of a Given Belief Tree} \label{sec:GivenTree}
We start with the assumption that the belief tree was generated in some way and that it is given, e.g, Sparse Sampling (SS) algorithm introduced by \cite{Kearns02jml}. In other words the belief tree construction is not coupled with rewards calculation and estimation of the objective.  

In this setting, we contribute two resimplification strategies. The first strategy is described in Section \ref{sec:SimplficiationLevel}.   The general idea is to break down recursively a given belief tree $\mathbb{T}$  into its sub-problems (subtrees), denoted as $\{\mathbb{T}^j\}_{j=1}^{|\mathcal{A}|}$ (each subtree $j$ at the root belief has a single action $j$), and solve each sub-problem with its own simplification level of the corresponding belief subtree. Ultimately this would lead to the solution of the entire problem via action-value function bounds \eqref{eq:SampleQboundsBellmanGivenTree}. This strategy is based on \texttt{Simplification Level} and it is a \texttt{PT} strategy. The action-value bounds should not overlap {\bf at each node} in the given belief tree. 

The second strategy is described in Section \ref{sec:LazyVariantGivenTree}. This resimplification strategy is based on \texttt{Gap} and it is a \texttt{LAZY} strategy.   Here, the general idea is to first substitute all the rewards in a given belief tree by bounds with the coarsest simplification level. We then 
eliminate an overlap between candidate actions only 
at the root belief node $b_k$ by a repetitive descending to the belief tree, promoting the simplification levels along a single lace chosen according to largest gap  and ascending back. We emphasize that in this setting, the action-value bounds should not overlap {\bf only at the root node} in the given belief tree.      
\begin{figure}[t]
	\centering         
	\includegraphics[width=\columnwidth]{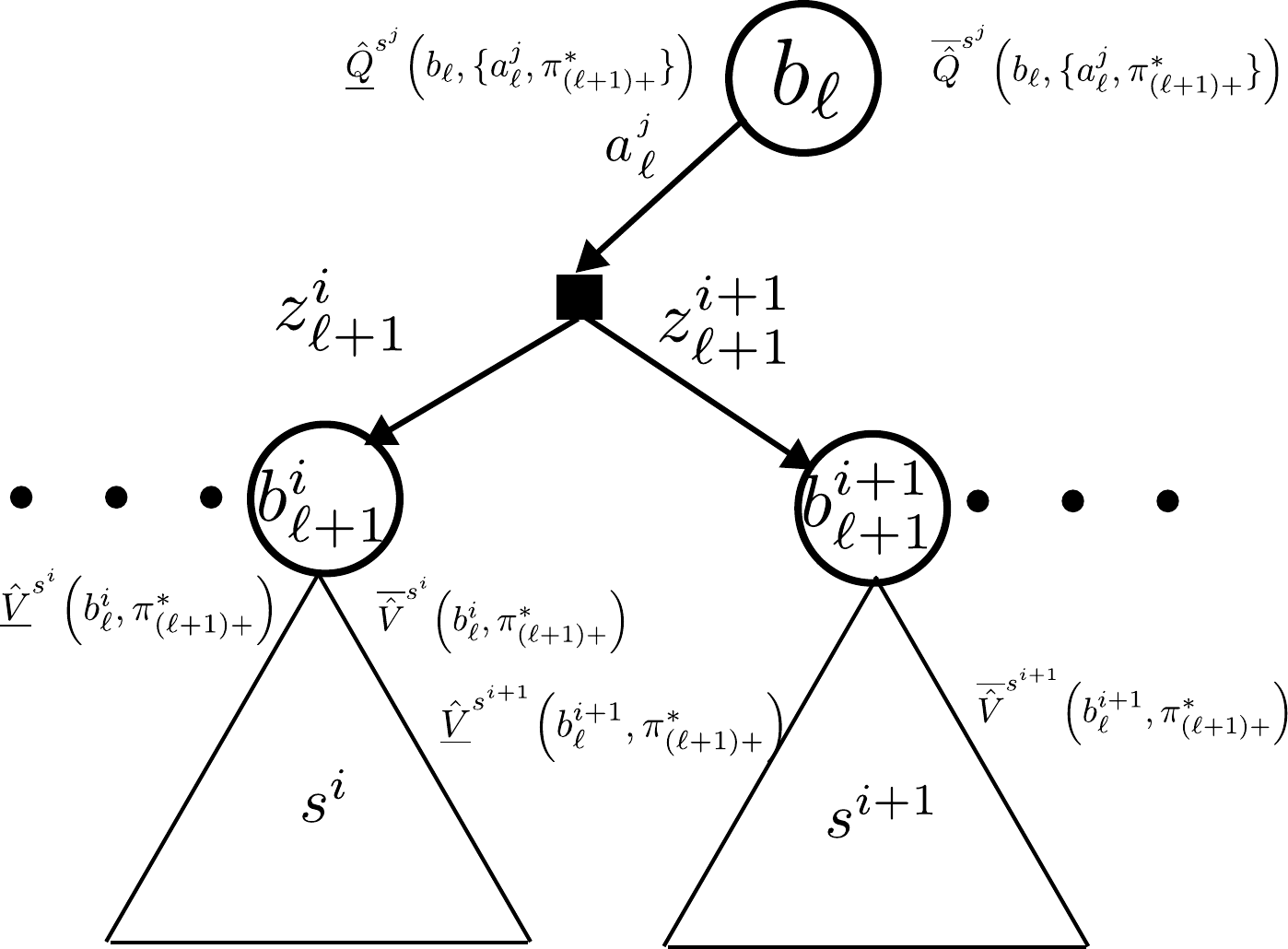}
	\caption{Pruning the subtrees by adaptively promoting the simplification levels of the rewards inside. Here the simplification levels of a subtrees are not equal. It is possible that  $s^i \neq s^{i+1}$. Note that here the superscripts are relative to $b_{\ell}$ as opposed to Fig.~\ref{fig:BeliefTreeInplaceSimp} and Fig.~\ref{fig:BoundBellmanExample}.}
	\label{fig:BoundBellman}
\end{figure}

As mentioned in the beginning of Section \ref{sec:GivenTreeEstimator}, only for simplicity we consider a symmetric setting in terms of sampled actions and the observations, but the approach is applicable without any limitations to any given belief tree.   
\subsection{Resimplification strategy: \texttt{PT-SL}}\label{sec:SimplficiationLevel}
This section presents our first resimplification strategy. We now turn to thorough description.

Not to be confused with {\bf policy tree} represented by the \eqref{eq:ValueSampleGivenTree} or \eqref{eq:QbellmanSampleGivenTree} the  { \bf given  belief tree} ($\mathbb{T}$) has more than a single action emanating from each belief node besides the leafs.

We now assign a simplification level to the bounds on the value and action value functions. 
Consider again some belief node $b_{\ell}$ in the belief tree, and assume recursively for \emph{each} of its children belief nodes $b_{\ell+1}$ we already calculated  the optimal policy $\pi_{(\ell+1)+}^{\ast}(b_{\ell+1})$ and the corresponding upper and lower bounds $\underline{\hat{V}}^{s}\big(b_{\ell+1}, \pi^{\ast}_{(\ell+1)+}\big)$ and $\overline{\hat{V}}^{s}\big(b_{\ell+1}, \pi^{\ast}_{(\ell+1)+}\big)$. In general, these bounds for each child sub-policy tree of $b_{\ell}$ can correspond to different simplification levels.

From now on let the superscript $s$ over the action-value function bounds from \eqref{eq:SampleQboundsBellmanGivenTree} and \eqref{eq:SampleVboundsBellmanGivenTree} denote the simplification level stemmed from pertaining reward bounds. The bounds previously described by Eqs.~\eqref{eq:SampleQboundsBellmanGivenTree} for belief node $b_{\ell}$, incorporating simplification level, are now modified to 
\begin{equation}\label{eq:BellmanQBoundsSimpLevel}
	\begin{gathered}
		\overline{\hat{Q}}^{s^j}(b_{\ell}, \{a^j_{\ell}, \pi^{\ast}_{(\ell+1)+}\}) = \frac{1}{n_z}\sum^{n_z}_{i=1}  \overline{\rho}^s(b_{\ell},\! a^j_{\ell}, b^i_{\ell+1})  + \\
		+\gamma \frac{1}{n_z}\sum^{n_z}_{i=1} \overline{\hat{V}}^{s^i} (b^i_{\ell+1},  \pi^{\ast}_{(\ell+1)+}) \\
		\underline{\hat{Q}}^{s^j} (b_{\ell}, \{a^j_{\ell},  \pi^{\ast}_{(\ell+1)+}\}) = \frac{1}{n_z}\sum^{n_z}_{i=1} \underline{\rho}^s(b_{\ell}, a^j_{\ell}, b^i_{\ell+1})  +\\
		+\gamma \frac{1}{n_z}\sum^{n_z}_{i=1} \underline{\hat{V}}^{s^i} (b^i_{\ell+1},  \pi^{\ast}_{(\ell+1)+}), \\
	\end{gathered}
\end{equation}
as illustrated in Fig.~\ref{fig:BoundBellman}. We shall pinpoint the abuse of notation here. In contrast to \eqref{eq:SampleQboundsBellmanGivenTree} the superscript $s$ over the immediate reward bounds denotes a specific  simplification level instead of indicating a general simplification.

Note  equation \eqref{eq:BellmanQBoundsSimpLevel} applies for each $a^j_{\ell}\in \mathcal{A}$, and as mentioned, each belief node $b^{i}_{\ell+1}$ (one for each observation $z^i_{\ell+1}$) has, in general, its own simplification level $s^{i}$.  In other words, for  each  $b^{i}_{\ell+1}$, $s^{i}$ is the simplification level that was sufficient for calculating the bounds $\Big\{  \overline{\hat{V}}^{s^i} \!\!\!\!(b^i_{\ell+1}, \! \pi^{\ast}_{(\ell+1)+}),  \underline{\hat{V}}^{s^i} \!\!\!(b^i_{\ell+1}, \! \pi^{\ast}_{(\ell+1)+})\Big\}$  and the corresponding optimal policy $\pi_{(\ell+1)+}^{\ast}$. Thus, when addressing belief node $b_\ell$ in \eqref{eq:BellmanQBoundsSimpLevel}, for each belief node $b^i_{\ell+1}$ and its corresponding simplification level $s^i$, these bounds are already available.  

Further, as seen in \eqref{eq:BellmanQBoundsSimpLevel}, the immediate reward and the corresponding bounds $\overline{\rho}$ and $\underline{\rho}$, in general, can be calculated with their own simplification level $s$. In particular, when starting calculations, $s$ could correspond to a default coarse simplification level, e.g.~coarsest level $s=1$.  Another possibility is to set $s=s^i$ for corresponding simplification level of value function bounds of the $i$-th child belief. 

To define simplification level $s^j$ of the bounds  \eqref{eq:BellmanQBoundsSimpLevel} we leverage the recursive nature  of the Bellman update and define 
\begin{equation}\label{eq:sj}
	s^j\triangleq \min\{\underbrace{s}_{\substack{ { \overline{\rho}^s} \\ {\underline{\rho}^s}}}, \underbrace{s^{i=1}, s^{i=2} \dots s^{i = {n_z}}}_{ \substack{ {\overline{\hat{V}}^{s^i}(b^i_{\ell+1}, \pi^{\ast}_{(\ell+1)+})} \\ {\underline{\hat{V}}^{s^i}(b^i_{\ell+1}, \pi^{\ast}_{(\ell+1)+})}}}\},
\end{equation}
where $\{s^{i=1}, s^{i=2}, \dots, s^{i = {n_z}}\}$ represent the (generally different) simplification levels of optimal value functions of  belief nodes $b^i_{\ell+1}$ considered in the expectation approximation in \eqref{eq:BellmanQBoundsSimpLevel}.
\begin{figure}[t]
	\centering         
	\includegraphics[width=0.7\columnwidth]{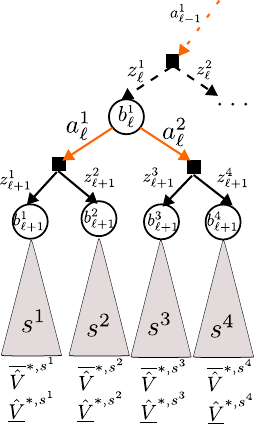}
	\caption{An example of the simplification paradigm. The superscript here denotes {\bf global} number of the belief, observation or action in the belief tree as opposed to equation \eqref{eq:BellmanQBoundsSimpLevel} and Fig~\ref{fig:BoundBellman}. }
	\label{fig:BoundBellmanExample}
\end{figure}

We now wish to decide which action $a_{\ell}^{j^{\dagger}(b_{\ell})}\in\mathcal{A}$ is optimal from belief node $b_{\ell}$; the corresponding optimal policy would then be $\pi_{\ell+}^{\ast} = \{a_{\ell}^{\ast}, \pi^{\ast}_{(\ell+1)+}\}$, where $\pi^{\ast}_{(\ell+1)+}$ is the already-calculated optimal policy for belief nodes $\{b^i_{\ell+1}\}_{i=1}^{n_z}$ that $a_{\ell}^{\ast}$ leads to.  See illustration in Fig.~\ref{fig:BoundBellman}.

Let us utilize now a general simplification approach described in section \ref{sec:GeneralSimplificationMechanics}. Overall in each belief node we have $n_a$ candidate actions indexed by superscript $j$ in \eqref{eq:BellmanQBoundsSimpLevel}.

 {\bf At each belief node } we first select an optimal action candidate according to \eqref{eq:OptAction} with a nullified action dependent constant ($\forall j \ c^j= 0$). Further, in any \texttt{PT} resimplification strategy there are  three possible scenarios. 
\begin{itemize}
	\item No overlap is present (\eqref{eq:AdaptCond} is satisfied) and we are at the root i.e.~$b_{\ell}=b_k$. In this case the optimal action shall be returned. 
	\item No overlap is present (\eqref{eq:AdaptCond} is satisfied) and we not at the root $b_k$. In this case, using the optimal action we bound optimal value function using the \eqref{eq:LowerValue} and \eqref{eq:UpperValue}.
	\item Eq.~\eqref{eq:AdaptCond} is not satisfied, meaning an overlap is present. In the presence of overlap we shall prune actions according to \eqref{eq:DismissingGeneral} and commence resimplification routine based on resimplification strategy.  
\end{itemize}

We now discuss how the simplification level is updated recursively from the simplification level of pertaining reward bounds, and revisit the process to calculate the optimal policy and the corresponding bounds.
For some belief node $b_{\ell}$ in the belief tree, consider the bounds $\overline{\hat{Q}}^{s^j}(b_{\ell}, \{a^j_{\ell},\pi^{\ast}_{(\ell+1)+}\})$ and $\underline{\hat{Q}}^{s^j}(b_{\ell}, \{a^j_{\ell},\pi^{\ast}_{(\ell+1)+}\})$ from \eqref{eq:BellmanQBoundsSimpLevel} for different actions $a^j_{\ell} \in \mathcal{A}$,  that partially overlap and therefore could not be pruned. Each policy tree corresponding to action $a^j_{\ell}$ can generally have its own simplification level $s^j$. We now iteratively increase the simplification level by $1$. This can be done for each of the branches, if $s^j$ is identical for all branches, or only for the branch with the coarsest simplification level. 

Consider now any such branch whose simplification level needs to be adapted from $s^j$ to $s^j+1$. Recall, that at this point, the mentioned bounds were already calculated, thus their ingredients, in terms of $\overline{\rho}(b^s_{\ell}, b_{\ell}, a^j_{\ell})$, $\underline{\rho}(b^s_{\ell}, b_{\ell}, a^j_{\ell})$ and  $\{ \overline{\hat{V}}^{s^i}(b^i_{\ell+1}, \pi^{\ast}_{(\ell+1)+}) , \underline{\hat{V}}^{s^i}(b^i_{\ell+1}, \pi^{\ast}_{(\ell+1)+})\}_{i=1}^{n_z}$, involved in approximating the expectation in  \eqref{eq:BellmanQBoundsSimpLevel}, are available. Recall also  \eqref{eq:sj}, i.e.~each element  in $\{s, s^{i=1}, s^{i=2}, \dots, s^{i = n_z}\}$ is either equal or larger than $s^j$. We now discuss both cases, starting from the latter.

As we assumed bounds to improve monotonically as simplification level increases, see Assump.~\ref{assumption:monotonic}, for any $s^i > s^j+1$ we already have readily available bounds $ \overline{\hat{V}}^{s^i}(b^i_{\ell+1}, \pi^{\ast}_{(\ell+1)+}) , \underline{\hat{V}}^{s^i}(b^i_{\ell+1}, \pi^{\ast}_{(\ell+1)+})$ 
which are tighter than those that would be obtained for simplification level $s^j+1$. Thus, we can \emph{safely skip} the calculation of the latter and use the existing bounds from level $s^i$ as is. 

For the former case, i.e.~$s^i = s^j$, we now have to adapt the  simplification level of a child tree $i$ to $s^j+1$ by calculating the bounds $ \overline{\hat{V}}^{s^i+1}(b^i_{\ell+1}, \pi^{\ast}_{(\ell+1)+}) , \underline{\hat{V}}^{s^i+1}(b^i_{\ell+1}, \pi^{\ast}_{(\ell+1)+})$. Here, our \emph{key insight} is that, instead of calculating these bounds from scratch, we can re-use calculations between different simplification levels, in this case, from level $s^i$. As the bounds from that level are available, we can identify only the incremental part that is ``missing'' to get from simplification level $s^i$ to $s^i+1$, and update the existing bounds $\overline{\hat{V}}^{s^i}(b^i_{\ell+1}, \pi^{\ast}_{(\ell+1)+}) , \underline{\hat{V}}^{s^i}(b^i_{\ell+1}, \pi^{\ast}_{(\ell+1)+})$ to recover $\overline{\hat{V}}^{s^i+1}(b^i_{\ell+1}, \pi^{\ast}_{(\ell+1)+}) , \underline{\hat{V}}^{s^i+1}(b^i_{\ell+1}, \pi^{\ast}_{(\ell+1)+})$ exactly. The same argument applies also for bounds over momentary rewards. In Section \ref{ReuseCalculationsEntropy} we apply this approach to a specific simplification and reward function.

We can repeat iteratively the above process of increasing the simplification level until we can prune all branches but one. This means each subtree will be solved maximum once, per simplification level. Since we assumed the reward bounds converge monotonically to the original reward for the finest level $s = n_{\mathrm{max}}$ (See Fig.~\ref{fig:BoundsReward}), from   Theorem~\ref{thm:MonotonConvQ}, we are guaranteed to eventually disqualify all sub-optimal branches.
Our described approach is summarized in Algs.~\ref{alg:sith-bsp} and \ref{alg:Prune}.

\subsubsection{Illustrative Example}\label{sec:IllustrativeExample}
We now illustrate the described above resimplification strategy in a toy example. Before we start this section, let us clarify that in the example the superscripts are global over the belief tree in contrast to previous section.  
Consider Fig.~\ref{fig:BoundBellmanExample} and  assume the subtrees to $b_{\ell}^1$ were solved using simplification levels that hold $s^2=s^1+1, s^2 < s^3,s^4$. Further assume the immediate reward simplification is $s=s^1$. According to definitions above this means that for subtree starting at $b_{\ell}^1$ and action $a^1_{\ell}$ the simplification level is   $\min\{s^{1}, s^{2}\}$ and for action  $a^2_{\ell}$  the simplification level is $\min\{s^{3}, s^{4}\}$. Now, we consider the case the existing bounds of the subtrees were not tight enough to prune, we adapt simplification level starting from $b_{\ell}^1$ and promote $ s  \leftarrow s^1+1$. Since $s^1 < s^1+1$ we re-simplify the subtree corresponding to simplification level of $s^1$ to simplification level $s^1+1$, i.e.~to a finer simplification.

However we do not need to re-simplify subtrees corresponding to $s^2,s^3,s^4$: The tree corresponding to  $s^2$ is already simplified to the currently desired level; thus we can use its existing bounds. For the two other trees, their current simplification levels, $s^3$ and $s^4$, are higher (finer) than the desired $s^1+1$ level, and since the bounds are tighter as simplification level increases we can use their existing tighter bounds without the need to ``go-back'' to a coarser level of simplification. If we can now prune one of the actions, we keep pruning up the tree. If pruning is still not possible, we need to adapt simplification again with simplification level $s^1+2$. 

\begin{algorithm}[t!]
	\caption{Simplified Information Theoretic Belief Space Planning (SITH-BSP)}
	\begin{algorithmic}[1]	
		\Procedure{SolveBeliefTree}{belief-tree: $\mathbb{T}$}
		\If { $\mathbb{T}$ is a leaf} 
		\State \textcolor{blue}{//Corresponds to a single belief node.}
		\State \textbf{return} $0, 0$
		\EndIf
		\ForAll{subtrees  $\mathbb{T}^j \in  \{\mathbb{T}^j\}_{j=1}^{|\mathcal{A}|}$} \textcolor{blue}{\Comment{Actions}}
		\State {\textcolor{blue}{//Observations}}
		\ForAll{subtrees  $\mathbb{T}^{j,i} \in  \{\mathbb{T}^{j,i}\}_{i=1}^{n_z}$} 
		\State \textcolor{blue}{//Returns Optimal Value bounds and prune suboptimal branches of $\mathbb{T}^{j,i}$.}
		\State \Call{SolveBeliefTree}{$\mathbb{T}^{j,i}$}  
		\State Set the simplification level of $\underline{\rho}^s(b, a^j, b'^{i})$ and $\overline{\rho}^s(b, a^j, b'^{i})$ as in \eqref{eq:sj}
		\EndFor
		\State Calculate $\underline{\hat{Q}}^{s^j},\overline{\hat{Q}}^{s^j}$ according to \eqref{eq:BellmanQBoundsSimpLevel}
		\EndFor
		\State \Call{Prune}{$\{ \underline{\hat{Q}}^{s^j},\overline{\hat{Q}}^{s^j}\}_{j=1}^{|\mathcal{A}|}$}  \textcolor{blue}{\Comment{Alg. \ref{alg:Prune}}} 
		\While{not all subtrees  $\mathbb{T}^j \in  \{\mathbb{T}^j\}_{j=1}^{|\mathcal{A}|}$ but $1$ pruned}
		\State Find minimal simplification level $s_{\mathrm{min}}$ between all  $\underline{\hat{Q}}^{s^j},\overline{\hat{Q}}^{s^j}$ corresponding to {\bf not} pruned $\mathbb{T}^j$
		\State \textcolor{blue}{// Can be more than single subtree}
		\State select subtree $s^j == s_{\mathrm{min}}$ 
		\State \Call{ResimplifyTree}{$\mathbb{T}^j$} 
		\State \Call{Prune}{$\{ \underline{\hat{Q}}^{s^j},\overline{\hat{Q}}^{s^j}\}_{j=1}^{|\mathcal{A}|}$}  \textcolor{blue}{\Comment{Alg. \ref{alg:Prune}}}
		\EndWhile
		\State \textbf{return} optimal action branch that left $a^{\ast}$  and $\underline{\hat{Q}}^{s^j},\overline{\hat{Q}}^{s^j}$. 
		\EndProcedure
		\Procedure{ResimplifyTree}{$\mathbb{T}^j$}
		\ForAll{subtrees  $\mathbb{T}^{j,i} \in  \{\mathbb{T}^{j,i}\}_{i=1}^{n_z}$}
		\State \Call{ResimplifyReward}{$\mathbb{T}^j$, $b$, $a^j$, $b^i$} \textcolor{blue}{\Comment{Alg.~\ref{alg:ResimplifyReward}}}
		\If{$b^i$ has children}
		\State \textcolor{blue}{// $s^i$ is a simplification level of corresponding optimal value function (policy tree) }
		\If{$s^i \leq s_{\mathrm{min}}$}
		\State  \textcolor{blue}{ // Alg.~\ref{alg:ResimplifySubtree}}
		\State \Call{ResimplifySubtree}{$\mathbb{T}^{j,i}$, $b$, $b^i$}
		\State $s^i \leftarrow s^i+1$
		\EndIf
		\EndIf
		\EndFor
		\State $s^j \leftarrow s^j+1$
		\EndProcedure
	\end{algorithmic}
	\label{alg:sith-bsp}
\end{algorithm}
\begin{algorithm}[t!]
	\caption{Pruning of trees}
	\begin{algorithmic}[1]
		\Procedure{Prune}{}\\
		\hspace*{\algorithmicindent} \textbf{Input:} (belief-tree root, $b$; bounds of root's children, $\{\underline{\hat{Q}}^j,\overline{\hat{Q}}^j \}_{j=1}^{n_a}$)\textcolor{blue}{\Comment{$n_a$ is the number of child branches (candidate actions) going out of $b$.}}
		\State $\underline{\hat{Q}}^{\ast} \leftarrow \maxim{max}{j} \{\underline{\hat{Q}}^j\}_{j=1}^{n_a}$
		\For{$j \in  1:n_a$ }
		\If{$\underline{\hat{Q}}^{\ast} > \overline{\hat{Q}}^j$}
		\State prune child $j$ from the belief tree
		\EndIf
		\EndFor
		\EndProcedure
	\end{algorithmic}
	\label{alg:Prune}
\end{algorithm}
\begin{algorithm}[t!]
	\caption{ResimplifyReward}
	\begin{algorithmic}[1]	
		\Procedure{ResimplifyRewad}{$\mathbb{T}^j$, $b$, $a$, $b'$}
		\State Obtain corresponding to the $\mathbb{T}^j$ bounds $\overline{\hat{V}}$, $\underline{\hat{V}}$
		\State  $\overline{\hat{V}} \leftarrow \overline{\hat{V}} -   \frac{\overline{\rho}^s(bab')}{n_z}$
		\State  $\underline{\hat{V}} \leftarrow \underline{\hat{V}} -   \frac{\underline{\rho}^s(bab')}{n_z}$
		\State Advance level of simplification of $b'$
		\State  $\overline{\hat{V}} \leftarrow \overline{\hat{V}} +   \frac{\overline{\rho}^s(bab')}{n_z}$
		\State  $\underline{\hat{V}} \leftarrow \underline{\hat{V}} +   \frac{\underline{\rho}^s(bab')}{n_z}$
		\EndProcedure
	\end{algorithmic}
	\label{alg:ResimplifyReward}
\end{algorithm}
\begin{algorithm}
	\caption{ResimplifySubtree}
	\begin{algorithmic}[1]	
		\Procedure{ResimplifySubtree}{$\mathbb{T}^{j,i}$,$b$ $b'$}
		\State  $\overline{\hat{V}}(b) \leftarrow \overline{\hat{V}}(b) -   \gamma \frac{\overline{\hat{V}}(b')}{n_z}$
		\State  $\underline{\hat{V}}(b) \leftarrow \underline{\hat{V}}(b) -    \gamma \frac{\underline{\hat{V}}(b')}{n_z}$
		\State \Call{ResimplifyTree}{$\mathbb{T}^{j,i}$} 
		\State  $\overline{\hat{V}} \leftarrow \overline{\hat{V}}(b) +   \gamma \frac{\overline{\hat{V}}(b')}{n_z}$
		\State  $\underline{\hat{V}} \leftarrow \underline{\hat{V}}(b) +    \gamma \frac{\underline{\hat{V}}(b')}{n_z}$
		\EndProcedure
	\end{algorithmic}
	\label{alg:ResimplifySubtree}
\end{algorithm}
\begin{algorithm}[h!]
	\caption{Lazy Simplified Information Theoretic Belief Space Planning (LAZY-BSP)}
	\begin{algorithmic}[1]
		\Procedure{Plan}{belief: $b$, belief-tree: $\mathbb{T}$}
		\State \Call{BoundOptimalValue}{belief: $b$, belief-tree: $\mathbb{T}$}
		\State	$a^* \leftarrow $ \Call{ActionSelection}{$b$, $L$} \textcolor{blue}{\Comment{Alg. \ref{alg:action-selection-lazy-sith-bsp}}}
		\State \Return $a^*$ 
		\EndProcedure
		\Procedure{BoundOptimalValue}{belief-tree: $\mathbb{T}$}
		\If { $\mathbb{T}$ is a leaf} 
		\State \textcolor{blue}{//Corresponds to a single belief node.}
		\State \textbf{return} $0, 0$
		\EndIf
		\ForAll{subtrees  $\mathbb{T}^j \in  \{\mathbb{T}^j\}_{j=1}^{|\mathcal{A}|}$} 
		\ForAll{subtrees  $\mathbb{T}^{j,i} \in  \{\mathbb{T}^{j,i}\}_{i=1}^{n_z}$} 
		\State $ \!\! \underline{\hat{V}}(b'), \overline{\hat{V}}(b') \!\! \leftarrow \!\!$ \Call{BoundOptimalValue}{$b$, $\mathbb{T}^{j,i}$}  
		\State Set the simplification level of $\underline{\rho}^s(b, a^j, b'^{i})$ and $\overline{\rho}^s(b, a^j, b'^{i})$ to coarsest possible
		\EndFor
		\State Calculate $\underline{\hat{Q}}^{j},\overline{\hat{Q}}^{j}$ 
		\EndFor
		\State $ \underline{\hat{V}}(b) \leftarrow \underset{j}{\max} \{\underline{\hat{Q}}^j\}$
		\State $ \overline{\hat{V}}(b) \leftarrow \underset{j}{\max} \{\overline{\hat{Q}}^j \}$
		\State \Return $\underline{\hat{V}}(b), \overline{\hat{V}}(b)$
		\EndProcedure
	\end{algorithmic}
	\label{alg:lazy-sith-bsp}
\end{algorithm}	
\begin{algorithm}[h!]
	\caption{Action Selection for Lazy Simplified Information Theoretic Belief Space Planning}
	\begin{algorithmic}[1]
		\Procedure{ActionSelection}{belief: $b$, horizon: $L$}
		\State \Call{Prune}{$\{ \underline{\hat{Q}}^{j},\overline{\hat{Q}}^{j}\}_{j=1}^{|\mathcal{A}|}$}  \textcolor{blue}{\Comment{Alg. \ref{alg:Prune}}}
		\State $a^{\dagger} \leftarrow \underset{a}{\argmax} \ \underline{\hat{Q}}\Big(b, \{a, \pi^{\ast}_{(k+1+)}\}\Big)$
		\State $\tilde{a} \leftarrow  \underset{a \in \mathcal{A} \setminus a^{\dagger}}{\argmax} \ \overline{\hat{Q}}\Big(b, \{a, \pi^{\ast}_{(k+1+)}\}\Big)$
		\State $\Delta \leftarrow  \Big(\overline{\hat{Q}} \big(b, \{\tilde{a}, \pi^{\ast}_{(k+1+)}\}\big){-} \underline{\hat{Q}}(b, \{a^{\dagger}, \pi^{\ast}_{(k+1+)}\})\Big)^+$ 
		\While {$\Delta > 0$}
		\State	$a^{\ast} \leftarrow $ \Call{LazyResimplify}{$b, L$} 
		\EndWhile
		\State \Return $a^*$
		\EndProcedure
		\Procedure{LazyResimplify}{belief: $b$}
		\If{$b$ is leaf}
		\State \Return $0, 0$ 
		\EndIf
		\State $\tilde{a} \!\leftarrow \! \underset{a}{\argmax} \ \overline{\hat{Q}}(b, \{a, \pi^{\ast}_{(k+1+)}\} {-} \underline{\hat{Q}}(b, \{a, \pi^{\ast}_{(k+1+)}\})$ \textcolor{blue}{\Comment{Gap as in \eqref{eq:gap}}}
		\If{$d == 1$ }
		\State  $b' \leftarrow \underset{b'}{\argmax} \quad   \overline{\rho}^s(b \tilde{a} b') - \underline{\rho}^s(b \tilde{a} b')$
		\Else 
		\State  $b' \leftarrow \underset{b'}{\argmax} \quad   \overline{\hat{V}}(b') - \underline{\hat{V}}(b')$
		\EndIf
		\State \Call{ResimplifyReward}{$\mathbb{T}$, $b$, $\tilde{a}$, $b'$}   \textcolor{blue}{\Comment{Alg.~\ref{alg:ResimplifyReward}}}
		\State  $\overline{\hat{Q}}(b, \{\tilde{a}, \pi^{\ast}_{(k+1+)}\} \leftarrow \overline{\hat{Q}}(b, \{\tilde{a}, \pi^{\ast}_{(k+1+)}\} -   \gamma \overline{\hat{V}}(b')$
		\State  $\underline{\hat{Q}}(b, \{\tilde{a}, \pi^{\ast}_{(k+1+)}\} \leftarrow \underline{\hat{Q}}(b, \{\tilde{a}, \pi^{\ast}_{(k+1+)}\} - \gamma \underline{\hat{V}}(b')$
		\State $\underline{\hat{V}}(b'), \overline{\hat{V}}(b')$ $\leftarrow$ \Call{LazyResimplify}{$b', d-1$}
		\State  $\overline{\hat{Q}}(b, \{\tilde{a}, \pi^{\ast}_{(k+1+)}\} \leftarrow \overline{\hat{Q}}(b, \{\tilde{a}, \pi^{\ast}_{(k+1+)}\} +   \gamma \overline{\hat{V}}(b')$
		\State  $\underline{\hat{Q}}(b, \{\tilde{a}, \pi^{\ast}_{(k+1+)}\} \leftarrow \underline{\hat{Q}}(b, \{\tilde{a}, \pi^{\ast}_{(k+1+)}\} + \gamma \underline{\hat{V}}(b')$
		\State $ \underline{\hat{V}}(b) \leftarrow \underset{a}{\max} \{\underline{\hat{Q}}(b, \{a, \pi^{\ast}_{(k+1+)}\} \}$
		\State $ \overline{\hat{V}}(b) \leftarrow \underset{a}{\max} \{\overline{\hat{Q}}(b, \{a, \pi^{\ast}_{(k+1+)}\} \}$
		\State \Return $\underline{\hat{V}}(b), \overline{\hat{V}}(b)$
		\EndProcedure
	\end{algorithmic}
	\label{alg:action-selection-lazy-sith-bsp}
\end{algorithm}

\subsubsection{A Detailed Algorithm Description}
Let us thoroughly describe  Alg.~\ref{alg:sith-bsp}. We are given a belief tree $\mathbb{T}$. First at the line $10$ Alg.~\ref{alg:sith-bsp} recursively descend to the leafs. When the line $11$ is hit for the first time the corresponding rewards are set to the initial simplification level or also possible that minimal level of child optimal value bounds is used. In our simulations we used minimal reward level. Further the algorithm calculates bounds over action-value function represented by \eqref{eq:BellmanQBoundsSimpLevel}. This happens in line $15$ of Alg.~\ref{alg:sith-bsp}. The next step is to try to prune all subtrees but one utilizing the Alg.~\ref{alg:sith-bsp}. Note, at this point all the subtrees $\mathbb{T}^j$ are already policy trees, namely only a single action emanating from each posterior belief. In there is more that single action left after pruning,  at the line  $20$ the Algorithm~\ref{alg:sith-bsp} calls routine \texttt{ResimplifyTree} to initiate { \bf resimplification} for selected subtree corresponding to action $a^j$.  The simplification level of a single step ahead reward is always have to be promoted as we do in line $27$. Further, Alg.~\ref{alg:sith-bsp} treats similarly subtrees, if they are present.

\subsection{Resimplification strategy: \texttt{LAZY-Gap}} \label{sec:LazyVariantGivenTree}
The \texttt{PT} resimplification strategy from previous section assure that no overlap is present (Fig.~\ref{fig:ActionGraphAdapted}) at  each non-leaf posterior belief and we know the optimal action to take. However, it can inflict a redundant computational burden. We can handle the overlap only at the root of the belief tree and use the bounds over optimal value function according to \eqref{eq:LazyLowerValue} and \eqref{eq:LazyUpperValue}. Since we already presented the resimplification strategy based on the simplification levels, our second resimplification strategy will be based on the distance between reward bounds.  However, the bounds \eqref{eq:LazyLowerValue} and \eqref{eq:LazyUpperValue} can be utilized directly also with the resimplifcation strategy based on simplification levels. Yet, this is out of the scope of this paper. 

In this section we present a lazy variant of the resimplifcation strategy. In a \texttt{LAZY} variant, the overlap is checked solely at the root $b_k$ of the whole belief tree. In this approach three scenarios can be encountered at each belief node.     
\begin{itemize}
	\item The belief node is not root. We bound optimal value according to \eqref{eq:LazyLowerValue} and \eqref{eq:LazyUpperValue}.
	\item At the root $b_k$ we shall check for overlap. If no overlap is present (\eqref{eq:AdaptCond} is satisfied) we prune all suboptimal actions according to Alg.~\ref{alg:Prune} and return an optimal action as described in Section \ref{sec:GeneralSimplificationMechanics}.   
	\item In the presence of an overlap at the root $b_k$ (Eq.~\eqref{eq:AdaptCond} is not satisfied), we shall prune actions according to \eqref{eq:DismissingGeneral} and Alg.~\ref{alg:Prune} and commence a resimplification routine for the non pruned actions based on the resimplification strategy.   
\end{itemize} 
Having presented general steps of  any \texttt{LAZY} variant of resimplification strategy, we are ready to delve into specific gap driven resimplfication strategy. 
Let us introduce the following notation 
\begin{align}
G(ha) \triangleq \overline{\hat{Q}}(ha)- \underline{\hat{Q}}(ha). \label{eq:gap}
\end{align}
We remind the reader that sometimes, for simplicity of explanation, we will make the gap dependent on belief and an action, and denote $G(ba)$.  We use this gap to steer the resimplification procedure towards more promising lace. The lace with actions inducing largest gap \eqref{eq:gap} at each belief action node along the lace will be selected to resimplification. In fact we use similar gap for value function to select observations along the lace. Now let us proceed to the detailed algorithm description.      

\subsubsection{A Detailed Algorithm Description}
 This approach is summarized in Alg.~\ref{alg:lazy-sith-bsp}
When we apply this resimplification strategy, we first use the lowest simplification level for each pair of consecutive beliefs in the given belief tree. In other words, the Alg.~\ref{alg:lazy-sith-bsp} first descends to the leaves of the given belief tree. Then it bounds each optimal value function using the initial simplification level using   \eqref{eq:LazyLowerValue} and \eqref{eq:LazyUpperValue}. This initial passage over the given belief tree is enclosed by routine \texttt{BoundOptimalValue}. In the procedure \texttt{ActionSelection} we increase the simplification level of the reward bounds in the given tree until there is no overlap at the {\bf root}, as in Fig.~\ref{fig:ActionGraphAdapted}. In this way, we can prune entire given subtrees at the root, corresponding to candidate actions. The procedure \texttt{LazyResimplify} descends back to some leaf through the lace with largest gaps on the way. It select action in line $15$. It then select observation/belief according to largest gap of a single step ahead rewards if these rewards are leafs (line $17$) or the largest gap of the optimal value function bounds (line $19$).

\section{Adaptive Simplification in the Setting of MCTS} \label{sec:MCTS}

In the previous sections, we described the application of the adaptive simplification paradigm when the belief tree is given or its construction is not coupled with the solution. 
We now turn to an anytime setting where the belief tree is not given. Instead, the belief tree construction is coupled with the estimation of the action-value function \eqref{eq:SampleQMCTS} at each belief action node. Such an approach is commonly used in Monte Carlo tree search (MCTS) methods based on an exploration strategy, e.g. Upper Confidence Bound (UCB) as in \eqref{eq:ucb}. Our goal is to suggest a resimplifcation strategy so that exactly the same belief tree as without simplification would be constructed.  Also the same optimal action is identified with and without simplification. To support general belief-dependent rewards we select PFT-DPW as the baseline, as mentioned in Section \ref{sec:RelatedWork}.

\begin{figure*}[t]
	\centering
	\includegraphics[width=\textwidth]{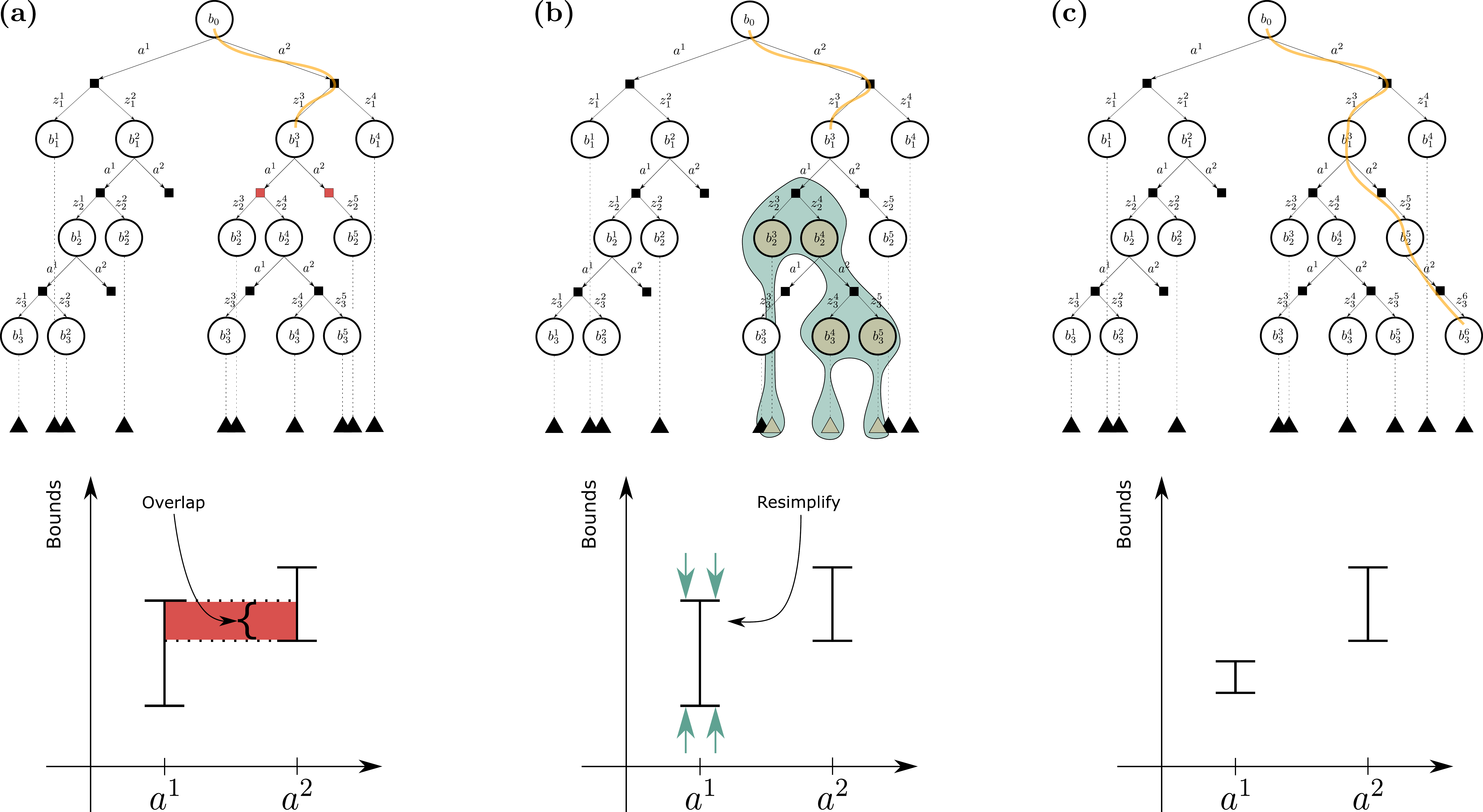}
	\caption{Illustration of our approach. The circles denote the belief nodes, and the rectangles represent the belief-action nodes. Rollouts, emanating from each belief node, are indicated by dashed lines finalized with triangles.  \textbf{(a)} The simulation starts from the root of the tree, but at node $b_1^3$ it can not continue due to an overlap of the child nodes (colored red) bounds. \textbf{(b)} One of the red colored belief-action nodes is chosen, and resimplification is triggered from it down the tree to the leaves (shaded green area in the tree). The beliefs and rollouts inside the green area (colored by light brown)  undergo resimplification if decided so. This procedure results in tighter bounds. \textbf{(c)} After the bounds got tighter, nothing prevents the SITH-PFT from continuing down from node $b_1^3$ guaranteeing the Tree Consistency. If needed, additional resimplifications can be commenced.}  
	\label{fig:ResimplIllustrationMCTS}
\end{figure*}
\begin{algorithm}[t!]
	\caption{SITH-PFT}
	\begin{algorithmic}[1]
		\Procedure{Plan}{belief: $b$}
		\For{$i \in 1:n$ or timeout} 
		\State $h \leftarrow \emptyset$ 
		\State	\Call{Simulate}{$b$, $d_{\text{max}}$, $h$}
		\EndFor
		\State \Return \Call{ActionSelection}{$b$, $h$} \textcolor{blue}{\Comment{called with nullified exploration constant $c$}}
		\EndProcedure
		\Procedure{Simulate}{belief: $b$, depth: $d$, history: $h$}
		\If { $d=0$ }
		\State \Return $0$
		\EndIf
		\State $a \leftarrow $ \Call{ActionSelection}{$b$, $h$}
		\If {$|C(ha)| \leq k_o N(ha)^{\alpha_{o}}$} 
		\State $o  \leftarrow$ sample $x$ from $b$, generate $o$ from $(x,a)$
		\State $b'  \leftarrow G_{\text{PF}(m)}(bao)$ 
		\State Calculate initial $\overline{\rho}^s, \underline{\rho}^s$ for $b, b'$ based on $s\leftarrow 1$ \textcolor{blue}{\Comment{minimal simp. level}}
		\State $C(ha) \leftarrow C(ha) \cup \{(\overline{\rho}^s, \underline{\rho}^s, b', o)\} $ 
		
		\State $L, U \leftarrow \overline{\rho}^s, \underline{\rho}^s + \gamma$ \Call{Rollout}{$b'$, $hao$, $d-1$}
		\Else 
		\State $(\overline{\rho}^s, \underline{\rho}^s, b', o) \leftarrow $ sample uniformly from $C(ha)$
		\State $L, U \leftarrow \overline{\rho}^s, \underline{\rho}^s + \gamma$ \Call{Simulate}{$b'$, $hao$, $d-1$}
		\EndIf
		\If {deepest resimplification depth $< d$ } \textcolor{blue}{\Comment{accounting for updated deeper in the tree bounds. See section~\ref{sec:sith-pft-description}}}
		\State reconstruct $\underline{\hat{Q}}(ha), \overline{\hat{Q}}(ha)$ 
		\EndIf 
		\State $N(h) \leftarrow N(h)+1$
		\State $N(ha) \leftarrow N(ha)+1$
		\State $\overline{\hat{Q}}(ha) \leftarrow \overline{\hat{Q}}(ha) + \frac{U-\overline{\hat{Q}}(ha)}{N(ha)}$ 
		\State $\underline{\hat{Q}}(ha) \leftarrow \underline{\hat{Q}}(ha) + \frac{L-\underline{\hat{Q}}(ha)}{N(ha)}$
		\State \Return $L, U$
		\EndProcedure
	\end{algorithmic}
	\label{alg:sith-pft}	
\end{algorithm}	
Common exploration strategies conform to the structure presented in \eqref{eq:OptAction}. Without loosing generality we focus on the most advanced to our knowledge exploration strategy, named UCB and  portrayed by \eqref{eq:ucb}.  
\subsection{UCB bounds}\label{sec:UCBBounds}
With this perspicuity in mind, we now introduce bounds over \eqref{eq:ucb}
\begin{align}
	&\!\!\! \uucb(ha) \triangleq \overline{\hat{Q}}(ha) + c \! \cdot \! \sqrt{\nicefrac{\log(N(h))}{N(ha)}}, \label{eq:ucbupper} \\
	&\!\!\!  \lucb(ha) \triangleq \underline{\hat{Q}}(ha) + c \! \cdot \! \sqrt{\nicefrac{\log(N(h))}{N(ha)}}. 	\label{eq:ucblower}
\end{align}
Similar to the given belief tree setting we now proceed to the explanation how the reward bounds \eqref{eq:BoundRewardGeneral} yield \eqref{eq:ucbupper} and \eqref{eq:ucblower}.

\subsection{Guaranteed Belief Tree Consistency}\label{sec:consistency}
\begin{algorithm}[t!]
	\caption{Action Selection for SITH-PFT}
	\begin{algorithmic}[1]
		\Procedure{ActionSelection}{$b$, $h$}
		\If{$|C(h)| \leq k_a N(h)^{\alpha_{a}}$} \textcolor{blue}{\Comment{action Prog. Widening}}
		\State  $a \leftarrow$ \Call{NextAction}{$h$}
		\State $C(h) \leftarrow  C(h) \cup \{a\}$
		\EndIf 
		\While{true}
		\State Status, $a \leftarrow$ \Call{SelectBest}{$b$, $h$}
		\If{Status}
		\State break
		\Else
		\ForAll {$b', o \in C(ha)$} 
		\State \Call{Resimplify}{$b'$, $hao$}
		\EndFor
		\State reconstruct $\overline{\hat{Q}}(ha), \underline{\hat{Q}}(ha)$ 
		\EndIf
		\EndWhile
		\State return $a$				
		\EndProcedure
		
		\Procedure{SelectBest}{$b$, $h$}
		\State Status $\leftarrow$ true
		\State $\tilde{a} \leftarrow \maxim{\argmax}{a}\{\lucb(ha)\}$ 
		\State gap $\leftarrow 0$ 
		\State child-to-resimplify $\leftarrow \tilde{a}$
		\ForAll{$ha$ children of $b$}
		\If{$\lucb(h\tilde{a}) < \uucb(ha)  \land a \neq \tilde{a}$ } 
		\State Status $\leftarrow$ false
		\If {$ \overline{\hat{Q}}(ha)- \underline{\hat{Q}}(ha) >$ gap}
		\State gap $\leftarrow  \overline{\hat{Q}}(ha)- \underline{\hat{Q}}(ha) $
		\State child-to-resimplify $\leftarrow$ $a$ 
		\EndIf
		\EndIf
		\EndFor
		\State \textbf{return} Status, child-to-resimplify
		\EndProcedure
	\end{algorithmic}
	\label{alg:ActionDis}
\end{algorithm}	

Since the simplification paradigm substituted  UCB \eqref{eq:ucb} by the bounds \eqref{eq:ucbupper} and \eqref{eq:ucblower}, the belief tree construction is coupled with these quantities, as opposed to the situation with the given belief tree. If there is an overlap between bounds on UCB for different actions, we can no longer guarantee the same belief tree will be constructed with and without simplification.

In this and the following sections we address this key issue. Specifically, we define the notion of Tree Consistency and prove the equivalence of our algorithm to our baseline PFT-DPW. 
\begin{defn}[Tree consistent algorithms]\label{def:TreeConsis}
	Imagine two algorithms, constructing a belief tree. Assume every common sampling operation for the two algorithms uses the same seed. 
	The two algorithms are \emph{tree consistent} if two belief trees constructed by the algorithms are identical in terms of actions, observations, and visitation counts. 	
\end{defn}
Our approach relies on a specific procedure for selecting actions within the tree. Since in each simulation the MCTS descends down the tree with a single return lace as in \eqref{eq:SampleQMCTS}, on the way down it requires the action maximizing UCB \eqref{eq:ucb} we shall eliminate overlap at each belief node as described in section \ref{sec:GeneralSimplificationMechanics}.  Further we restate the action selection procedure described in section \ref{sec:GeneralSimplificationMechanics} with particular action dependent constant from  eq.~ \eqref{eq:OptAction} and \eqref{eq:AdaptCond} rendering the UCB bounds from \eqref{eq:ucbupper} and \eqref{eq:ucblower}.

 Our action selection is encapsulated by Alg.~\ref{alg:ActionDis}, which is different from the procedure used in PFT-DPW. On top of DPW as in \cite{Sunberg18icaps} with parameters $k_a$ and $\alpha_{a}$, instead of selecting an action maximizing the UCB \eqref{eq:ucb},
at every belief node we mark as a candidate action the one that maximizes the lower bound $\lucb$ as such 
\begin{align}
	\tilde{a} = \maxim{\argmax }{a \in C(h)}  \ \lucb(ha).
\end{align}
If  $\forall a \neq \tilde{a}$, $\lucb(h\tilde{a}) \geq \uucb(ha)$, 
there is no overlap (Fig.~\ref{fig:ResimplIllustrationMCTS} \textbf{(c)}) 
and we can declare that $\tilde{a}$  is identical to $a^{\ast}$, i.e.,~the action that would be returned by PFT using \eqref{eq:ucb} and the tree consistency has not been affected. Otherwise, the bounds must be tightened, so ensure the tree consistency. We examine the $ha$ siblings of $h\tilde{a}$, which satisfy   $ a \neq \tilde{a}:\lucb(h\tilde{a}) < \uucb(ha)$ (Fig.~\ref{fig:ResimplIllustrationMCTS} \textbf{(a)}). 
Our next step is to tighten the bounds by resimplification (Fig.~\ref{fig:ResimplIllustrationMCTS} \textbf{(b)}) 
until there is no overlap using the valid resimplification strategy according to Definition \ref{def:ResimplificationStrategy}. 

{\bf Remark: }Note that here we cannot use the ``lazy variant'' from Section \ref{sec:LazyVariantGivenTree} due to the fact that the MCTS requires selecting an action going down to the tree, see line $12$ of  Algorithm~\ref{alg:sith-pft}. Therefore, if the UCB bounds do still overlap, we cannot assure that the same acton will be selected as in case of UCB itself.

\subsection{A Detailed Algorithm Description} \label{sec:sith-pft-description} 
Now we introduce our efficient variant of the Particle Filter Tree (PFT) presented in \cite{Sunberg18icaps}. We call our approach Simplified Information-Theoretic Particle Filter Tree (SITH-PFT). SITH-PFT (Alg.~\ref{alg:sith-pft}) incorporates the adaptive simplification into PFT-DPW.  We adhere to the conventional notations as in \cite{Sunberg18icaps} and denote by $G_{\text{PF}(m)}(bao)$ a generative model receiving as input the belief $b$, an action $a$ and an observation $o$, and producing the posterior belief $b'$. For belief update, we use a particle filter based on $n_x$ state samples. A remarkable property of our efficient variant is the consistency of the belief tree. In other words, PFT and SITH-PFT have the same belief tree constructed with \eqref{eq:ucb}, while SITH-PFT enjoys substantial acceleration.
By $C(ha)$ we denote the set of the children (posterior beliefs corresponding to the myopic observations) of the belief action node uniquely indexed by the history $h$ with concatenated action $a$.  Line 13 in Alg.~\ref{alg:sith-pft} is the DPW technique from \cite{Sunberg18icaps} with parameters $k_o$ and $\alpha_{o}$. The $N(\cdot)$ is the visitation count of belief or belief action nodes. 
In MCTS, the $Q$ estimate is assembled by averaging the laces of the returns over simulations see Eq.~\ref{eq:SampleQMCTS}. Each simulation yields a sum of discounted cumulative rewards.   Therefore, by replacing the reward with adaptive lightweights bounds \eqref{eq:BoundRewardGeneral}, we get corresponding discounted cumulative upper and lower bounds over the returns. Averaging the simulations (Alg.~\ref{alg:sith-pft} lines 28-29), yields the bounds over the action-value function and the UCB bounds used in the routine \texttt{ActionSelection}() to be explained in the next paragraph.

Consider a belief-action node $ha$ at level $d$ with  $\overline{\hat{Q}}(ha)$, $\underline{\hat{Q}}(ha)$. Suppose the algorithm selects it for bounds narrowing, as described in section~\ref{sec:consistency} and Alg.~\ref{alg:ActionDis} line $7$. All tree nodes of which $ha$ is an ancestor, contribute their immediate $\overline{\rho}^s, \underline{\rho}^s$ bounds to $\overline{\hat{Q}}(ha)$, $\underline{\hat{Q}}(ha)$ computation. Thus, to tighten $\overline{\hat{Q}}(ha), \underline{\hat{Q}}(ha)$, we can potentially choose any candidate node(s) in the subtree of $ha$. Each child belief node of $ha$ is sent to the resimplification routine (Alg.~\ref{alg:ActionDis} lines $11 - 13$), which performs the following tasks. 
First, 
it selects the action (Alg.~\ref{alg:resimpl} line 7) that will participate in the subsequent resimplification call and sends all its children beliefs nodes to the recursive call further down the tree (Alg.~\ref{alg:resimpl} line 8-10).  Secondly, 
It refines the belief node  $\overline{\rho}, \underline{\rho}$ according to the specific \emph{resimplification strategy} (Alg.~\ref{alg:resimpl} lines $3,4,12,18$). Thirdly, 
it reconstructs $\overline{\hat{Q}}(ha)$, $\underline{\hat{Q}}(ha)$ once all the child belief nodes of $ha$ have returned from the resimplification routine (Alg.~\ref{alg:resimpl} line 11) as we thoroughly explain in the next section.
Fourthly, 
it engages the rollout resimplification routine according to the specific \emph{resimplification strategy} (Alg.~\ref{alg:resimpl} lines 4, 13). Upon completion of this resimplification call initiated at $ha$, we obtain tighter immediate bounds of some of $ha$ descendant belief nodes (including rollouts nodes). Accordingly, appropriate descendant of $ha$ belief-action nodes bounds ($\overline{\hat{Q}}, \underline{\hat{Q}}$) shall be updated. 

Many resimplification strategies are possible, below we present our approach. In Section \ref{sec:LazyVariantGivenTree} we presented a resimplicifation strategy based on gap. Now we adapt it to the MCTS setting.

\subsection{Specific Resimplification Strategy: \texttt{PT-Gap} }\label{sec:resimplificationMCTS}
In this section, we explain the resimplification procedure in more detail. In particular we present a specific resimplification strategy and further show that this strategy is valid according to  Definition~\ref{def:ResimplificationStrategy}.  
When some sibling belief action nodes  have overlapping bounds (Fig.~\ref{fig:ActionGraph}, Fig.~\ref{fig:ResimplIllustrationMCTS}), we strive to avoid tightening them all at once since fewer resimplifications lead to greater acceleration (speedup).  
Thus, we choose  a single $ha$-node that causes the largest ``gap'', denoted by $G$, between its bounds (see Alg.~\ref{alg:ActionDis} lines 24-30), where $G$ is defined by \eqref{eq:gap}.
\begin{algorithm}[t]
	\caption{Resimplification}
	\begin{algorithmic}[1]
		\Procedure{Resimplify}{$b$, $h$}
		\If { $b$ is a leaf}
		\State \Call{RefineBounds}{$b$}
		\State \Call{ResimplifyRollout}{$b$, $h$}
		\State \textbf{return}
		\EndIf
		\State $\tilde{a} \leftarrow \maxim{\argmax}{a} \{N(ha)\cdot(\overline{\hat{Q}}(ha)-\underline{\hat{Q}}(ha))\}$ 
		\ForAll {$b', o \in C(h\tilde{a})$}
		\State \Call{Resimplify}{$b'$, $h\tilde{a}o$}
		\EndFor
		\State reconstruct $\overline{\hat{Q}}(h\tilde{a}), \underline{\hat{Q}}(h\tilde{a})$
		\State \Call{RefineBounds}{$b$}
		\State \Call{ResimplifyRollout}{$b$, $h$}
		\State \textbf{return} 
		\EndProcedure
		
		\Procedure{ResimplifyRollout}{$b$, $h$}
		\State $b^{\text{rollout}} \leftarrow$ find weakest link in rollout
		\State \Call{RefineBounds}{$b^{\text{rollout}}$}
		\EndProcedure
		
		\Procedure{RefineBounds}{$b$}
		\State if \eqref{eq:cond} holds for $b$, refine its $\overline{\rho}^{s+1}, \underline{\rho}^{s+1} $ and promote its simplification level
		\EndProcedure
	\end{algorithmic}
	\label{alg:resimpl}
\end{algorithm}
Further, we tighten the bounds down the branch of the chosen node (see Alg.~\ref{alg:ActionDis} lines 11-13) for each member of $C(ha)$, the set of children of $ha$.  Since the bounds converge to the actual reward (Assumption \ref{assumption:convergence}) we can guarantee that Alg.~\ref{alg:ActionDis} will pick a single action after a finite number of resimplifications; thus, tree consistency is assured. 


Specifically, we decide to refine $\overline{\rho}^s, \underline{\rho}^s$ of a belief node indexed by $h'$  at depth $d'$ within the subtree starting from a belief action node indexed by $ha$ at depth $d$ when 
\begin{align}
	\gamma^{d-d'} \cdot (\overline{\rho}^s- \underline{\rho}^s) \geq \frac{1}{d} G(ha), \label{eq:cond}
\end{align}
where
$G(ha)$ corresponds to the gap \eqref{eq:gap} of the belief-action node $ha$ that initially triggered resimplification in Alg.~\ref{alg:ActionDis} line 24. 

The explanation of resimplification strategy based on \eqref{eq:cond} is rather simple. 
The right hand side of \eqref{eq:cond} is the mean gap per depth/level in the sub-tree with $ha$ as its root and spreading downwards to the leaves.  Naturally, 
some of the nodes in this subtree have $\overline{\rho}^s-\underline{\rho}^s$  above or equal to the mean gap and some below. We want to locate and refine all those above or equal to it. For the left side of  \eqref{eq:cond};  the rewards are accumulated and discounted according to their depth. Thus, we must account for the relative discount factor. Note that the depth identified with the root is $d_{\max}$, as seen in Alg.~\ref{alg:sith-pft} line $4$, and the leaves are distinguished by depth $d=0$.
For each rollout originating from a tree belief node, we find the rollout node with the largest  $\overline{\rho}-\underline{\rho}$ satisfying \eqref{eq:cond} term locally in the rollout and resimplify it (Alg.~\ref{alg:resimpl} lines 4,13).
To choose the action to continue resimplification down the tree, we take the action corresponding to the belief action node with the largest gap, weighted by its visitation count (Alg.~\ref{alg:resimpl} line 7). With this strategy, we aim to keep the belief tree at the lowest possible simplification level while maintaining belieftree consistency.

If the action selection procedure triggers resimplification, it modifies the bounds through the tree. Since the resimplification works recursively, it reconstructs the belief-action node bounds coming back from the recursion (Alg.~\ref{alg:resimpl} line 11). Similarly, the action dismissal procedure reconstructs $\overline{\hat{Q}}$ and $\underline{\hat{Q}}$ of the belief-action node at which the action dismissal is performed (Alg.~\ref{alg:ActionDis} line 14).  Moreover, on the way back from the simulation, we shall update the ancestral belief-action nodes of the tree. Specifically, we need to reconstruct each $\overline{\hat{Q}}$ and $\underline{\hat{Q}}$ that is higher than the deepest starting point of the resimplification (Alg.~\ref{alg:sith-pft} line 23-25).
The reconstruction is essentially a double loop. To reconstruct $\overline{\hat{Q}}(ha),\underline{\hat{Q}}(ha)$ we first query for all belief children nodes $hao$. We then query all belief-action nodes that are children to the $hao$, i.e. $haoa'$. The possibly modified immediate bounds $\underline{\rho}$ and $\overline{\rho}$ are taken from $hao$ nodes and the $\overline{\hat{Q}}(\cdot)$, $\underline{\hat{Q}}(\cdot)$ bounds are taken from the $haoa'$ nodes. Importantly, each of the bounds is weighted according to the proper visitation count.

\subsection{Guarantees} \label{sec:guarantees}
In this section we first show that the resimplification strategy suggested in the previous section is valid. 
\begin{lem}[Validity of the suggested resimplification strategy]\label{lem:ValidStrategy}
	The resimplification strategy presented in Section~\ref{sec:resimplificationMCTS} promotes the simplification level of at least one reward in the rollout or belief tree. Alternatively, all the rewards are at the maximal simplification level $n_{\mathrm{max}}$. In other words the suggested resimplifcation strategy is valid.  
\end{lem}
We provide the complete proof in Appendix~\ref{proof:ValidStrategy}. Having proved the validity of the suggested resimplification strategy, we proceed to the monotonicity and convergence of UCB bounds from \eqref{eq:ucbupper} and \eqref{eq:ucblower}. 

\begin{lem}[Monotonicity and convergence of UCB bounds]\label{lem:UCBboundsmonotonicity}
	The UCB bounds are monotonic as a function of the number of resimplifications and after at most $n_{\mathrm{max}} \cdot M$ resimplifications we have that 
	\begin{align}
		 \uucb(ha) =  \lucb(ha) = \mathrm{UCB}(ha)
	\end{align}  
\end{lem}	
We provide the proof in Appendix~\ref{proof:UCBboundsmonotonicity}. Now, using Lemma~\ref{lem:UCBboundsmonotonicity}, we prove that SITH-PFT (Alg.~\ref{alg:sith-pft}) yields the same belief tree and the same best action as PFT.  
\begin{thm}\label{thm:TreeConsis}  
	SITH-PFT and PFT are Tree Consistent Algorithms for {\bf any} valid resimplification strategy. 
\end{thm} 
\begin{thm}\label{thm:SolConsis}
	SITH-PFT provides the same solution as PFT for {\bf any} valid resimplification strategy.
\end{thm}

We provide the full proofs of Theorems \ref{thm:TreeConsis} and \ref{thm:SolConsis} in  Appendix~\ref{sec:TreeConsis} and \ref{sec:SolConsis}, respectively. We showed that for any valid resimplification strategy SITH-PFT is guaranteed to construct the  same belief tree as PFT and select the same best action at the root. From Lemma~\ref{lem:ValidStrategy}, our resimplification strategy is valid. Thus, we achieved the desired result.

\section{Specific Simplification and Information-theoretic Bounds}
\label{sec:SpecificSimplification} 
In this section we focus on a specific simplification in the context of a continuous state space and nonparametric beliefs represented by $n_x$ weighted particles,
\begin{align}\label{eq:BeliefParticles}
b \bydef \{w^i, x^i\}^{n_x}_{i=1}.
\end{align}
\emph{Suggested Simplification:} Given the belief representation  \eqref{eq:BeliefParticles}, the simplified belief is a subset of $n^s_x$ particles, sampled from the original belief, where $n^s_x \leq n_x$. More formally:
\begin{equation}\label{eq:BeliefSimplification}
b^s_k \!\triangleq \!\!\big\{(x^i_k, w^i_k) \big| i \in A^s_k \subseteq \{1,2,\ldots, n_x\}, \! |A^s_k| = n_x^s \big\},
\end{equation}
where $A^s_k$ is the set of particle indices comprising the simplified belief $b^s_k$ for time $k$.

Increasing the level of simplification is done \emph{incrementally}. Specifically,
when resimplification is carried out, new indices are drawn from the sets $\{1,2,\ldots,n_x\} \setminus A^s_k$ and and included to the set  $A^s_k$. This operation promotes the simplification level to $s+1$ and defines  $A^{s+1}_k$.

\subsection{Novel Bounds Over Differential Entropy Estimator} \label{sec:Bounds}

As one of our key contributions, we now derive novel analytical bounds 
for the differential entropy estimator from \cite{Boers10fusion}. These bounds can then be used within our general simplification framework presented in the previous sections. 
To calculate differential entropy $$\mathcal{H}(b(x_k)) \bydef -\int b(x_k)\cdot \log\left(b(x_k)\right) \mathrm{d}x_k,$$ one must have access to 
the manifold representing the belief. In a nonparametric setting this manifold is out of reach. We have to resort to approximations. Several approaches exist. One of them is using Kernel Density Estimation (KDE) as done, e.g., by \cite{Fischer20icml}. Here, however, we consider the method proposed by \cite{Boers10fusion}. This method builds on top of usage of motion and observation models such that 
\begin{flalign}\label{eq:BoersDiffEnt}
	& \hat{\mathcal{H}}(b_{k}, a_k, z_{k+1}, b_{k+1})\triangleq \log \left[\sum_{i=1}^{n_x} \probd_Z (z_{k+1}| x_{k+1}^i) w_k^i\right]-\\
	&-\!\!\sum_{i=1}^{n_x} \! w_{k+1}^i \!\cdot \!\log \! \left[\probd_Z(z_{k+1}| x_{k+1}^i) \sum_{j=1}^{n_x}\probd_T(x_{k+1}^i| x_k^j, a_k)w_k^j\right]\!.\notag
\end{flalign}
One can observe this method requires access to samples representing both $b_k$ and $b_{k+1}$; thus, this corresponds to an information-theoretic reward of the form $r^I(b_k, a_k, z_{k+1}, b_{k+1})$. Note that as explained in Section \ref{sec:GeneralApproach} such a reward is tied to $b_{k+1}$. 

For the sake of clarity and to remove unnecessary clutter we apply an identical simplification described by \eqref{eq:BeliefSimplification} to both beliefs $b_k$ and $b_{k+1}$. The simplification indices  for both beliefs  are defined by $A^s_{k+1}$. However this is not an inherent limitation. One can easily maintain two sets of indices so as the theory presented below is developed to this more general setting.  Moreover, as mentioned in Section \ref{sec:GeneralApproach}, we have the same belief $b_{k+1}$ also participating in   $r^I(b_{k+1}, a_{k+1}, z_{k+2}, b_{k+2})$. In this reward, the simplification indices for $b_{k+1}$ will according to $A^s_{k+2}$ (and not according to $A^s_{k+1}$).  

Utilizing the chosen simplification \eqref{eq:BeliefSimplification}, we now introduce the following upper and lower bounds on \eqref{eq:BoersDiffEnt}. 
\begin{thm}[Adaptive bounds on differential entropy estimator]\label{theorem:LowerUpperReward}
	The estimator \eqref{eq:BoersDiffEnt} can be bounded  by
\begin{equation}
\label{eq:SpecificBounds}
\begin{gathered}
	\ell(b_{k}, a_k, z_{k+1}, b_{k+1}; A^s_k, A^s_{k+1}) \leq \\
	\leq -\hat{\mathcal{H}}(b_{k}, a_k, z_{k+1}, b_{k+1}) \leq \\
	\leq u(b_{k}, a_k, z_{k+1}, b_{k+1}; A^s_k, A^s_{k+1}),
\end{gathered}	
\end{equation}
where
	\begin{align}
		&u \bydef -\log \left[\sum_i^{n_x} \probd_Z(z_{k+1} |  x_{k+1}^i) w_k^i\right]  + \label{eq:immediate_bounds_l}\\
		&{+}\!\!\!\!\!\sum_{i \notin  A^s_{k+1}} \!\!\! w_{k+1}^i\cdot\log\left[ m \cdot \probd_Z(z_{k+1} |  x_{k+1}^i)\right] +\nonumber\\
		&{+}\!\!\!\!\!\sum_{i \in A^s_{k+1}} \!\!\! w_{k+1}^i \! \cdot \! \log\left[\probd_Z(z_{k+1} | x_{k+1}^i) \!\! \sum_j^{n_x}\probd_T(x_{k+1}^i | x_k^j, a_k)w_k^j\right] \nonumber \\
		& \ell \bydef -\log \left[\sum_i^{n_x} \probd_Z(z_{k+1} | x_{k+1}^i) w_k^i\right] +  \label{eq:immediate_bounds_u}\\
		&{+}\sum_i^{n_x} w_{k+1}^i \! \cdot \! \log\left[ \probd_Z(z_{k+1} | x_{k+1}^i)\!\!\! \sum_{j \in A^s_k} \probd_T(x_{k+1}^i | x_k^j, a_k)w_k^j\right] \nonumber
	\end{align}
and where superscript $s$ is the discrete level of simplification $s \in \{1, 2, \ldots, n_{\mathrm{max}}\}$, 	$m \triangleq \maxim{\max}{\substack{{x'} \\ x, a}} \ \probd_T(x'|x, a)$ and $A^s_k$,  $A^s_{k+1} \subseteq \{1,2,\ldots,n_x\}$.

\end{thm}
See proof in Appendix \ref{proof:LowerUpperReward}.
Theorem \ref{theorem:LowerUpperReward} accommodates different  sets $A^s_k \neq A^s_{k+1}$.  These sets denote sets of particle indices from $b_k$ and $b_{k+1}$ for simplification level $s$.  In  general, each of these sets can have its own simplification level. However, this is out of the scope of this paper. Here, both sets $A^s_k$, $A^s_{k+1}$ have the same simplification level, as well as the number of levels. Yet, the number of particles at each level can vary between $A^s_{k}$ and $A^s_{k+1}$.   Each subsequent level (low to high)  defines a larger set of indices such that higher levels of simplification (i.e.~more samples) correspond to tighter  and lower levels of simplification correspond to looser bounds. Note that  the bounds \eqref{eq:immediate_bounds_l} and \eqref{eq:immediate_bounds_u} actually use the original and simplified beliefs so it settles with Eqs.~\eqref{eq:BoundRewardPairConsecutive} and \eqref{eq:BoundRewardGeneral}.

Importantly, by caching the shared calculations of  both bounds in the same time instance,
we never repeat the calculation of these values and obtain maximal speedup. 
Without compromising on the solution's quality we are accelerating the online decision making process.

\begin{figure}[t]  
	\centering         
	\includegraphics[width=\columnwidth]{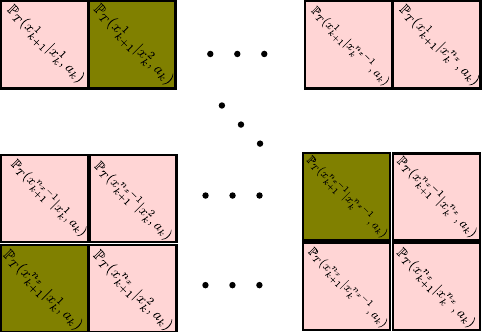}
	\caption{Schematic visualization of calculations reuse principle in bounds. We select {\bf columns } using indexes from set $A^s_{k}$  and rows by $A^s_{k+1}$. We marked by {\color{olive}{\bf olive}} color resulting constituents of the bounds. }
	\label{fig:reuse}
\end{figure}

\subsection{Bounds Properties and Analysis}
We now turn to analysis of the bounds and investigation of their properties. Allow us to start from computational complexity. We then examine monotonicity and convergence of the bounds and reuse of calculations. 
	
\subsubsection{Computational complexity}
Eqs.~\eqref{eq:immediate_bounds_l} and \eqref{eq:immediate_bounds_u} suggest that the bounds are cheaper to calculate than $\hat{\mathcal{H}}$ from \eqref{eq:BoersDiffEnt}, with complexity of $O(n_x^s\cdot n_x)$ instead of $O(n_x^2)$, where $n_x^s \triangleq | A^s_{k} |\equiv | A^s_{k+1} |$. 
Altogether, time saved for all belief nodes in the tree will result in the total speedup of
our approach.
\subsubsection{Monotonicy and Convergence }
\begin{thm}[Monotonicity and convergence]\label{theorem:MonotonicityConv}
	The bounds from \eqref{eq:SpecificBounds} are monotonic (Assumption \ref{assumption:monotonic}) and convergent (Assumption~\ref{assumption:convergence}) to \eqref{eq:BoersDiffEnt}. 
\end{thm}
\noindent See proof in Appendix \ref{proof:MonotonicityConv}.  
Finally,  bounding \eqref{eq:BoersDiffEnt} using Theorem \ref{theorem:LowerUpperReward} corresponds, in our general framework from Section \ref{sec:GeneralApproach}, to  \eqref{eq:BoundRewardPairConsecutive}. 
	
\subsubsection{Re-use of Calculations}\label{ReuseCalculationsEntropy}
The bounds can be tightened on demand incrementally without an overhead.
Moving from simplification level $s$ to level $s+1$, corresponds to adding some $m$ additional particles to $b^s$ to get $b^{s+1}$. 
For bounds calculation, we store the highlighted elements of the matrix in Fig.~\ref{fig:reuse}. This allows us to reuse the calculations when promoting the simplification level and {\bf between the lower and the upper bounds } in a particular time index. Namely, after a few bounds-contracting iterations they are just the reward itself and the entire calculation is roughly time-equivalent to calculating the original reward. This will happen in a worst-case scenario.

We provide the theoretical time complexity analysis using the specific bounds (from Section \ref{sec:Bounds}) in  Appendix~\ref{sec:TimeComplexity}. 
Now we are keen to present our simulations.

\section{Adaptation Overhead}
\label{sec:AdaptationOverhead}
Whereas the bounds presented in Section~\ref{sec:SpecificSimplification} are incremental repeated resimplifications may lead to actually slower decision-making. This overhead is caused by additional algorithmics introduced by the resimplification routine.  We can anticipate such scenarios when the actions are symmetrical in terms of the reward. However, as we empirically observed and will shortly present in the next section, in the setting of given belief  tree the cases where the simplification is beneficial prevail. Especially in the LAZY variant since there the Alg.~\ref{alg:lazy-sith-bsp} engages resimplification routine only at the root of the belief tree.  
	
In the setting of MCTS the situation is slightly more complicated. In UCB we cannot prune actions for eternity but only dismiss up until the next arrival to the belief node. This is because when MAB (defined in Section~\ref{sec:CoupledTreeEstimator}) converges it switches the current best action with arrivals to the belief node; such a behavior necessitates  our simplification approach to tighten the bounds for many candidate actions. As a result in a MCTS setting we obtain less speedup than in the setting of a given belief tree considering LAZY variant (Alg.~\ref{alg:lazy-sith-bsp}).  Nevertheless in some problems the simplification approach is invaluable, as for example, in the problem described in Section~\ref{sec:SafeAutonomousLocalization} and investigated in Section~\ref{sec:SafeLoc}. Importantly, we can further accelerate resimplification routines by parallelization. However, this is out of the scope of this paper. All our implementations are single threaded.

\section{Simulations and Results} \label{sec:SimsRes}
We evaluate our proposed framework and approaches in simulation considering the setting of nonparametric fully continuous POMDP.  Our implementation is built upon the JuliaPOMDP package collection \citep{egorov2017pomdps}.
For our simulations, we used  a $16$ cores 11th Gen Intel(R) Core(TM) i9-11900K with 64 GB of RAM working at 3.50GHz. 

First, we study empirically the specific simplification and bounds from Section \ref{sec:SpecificSimplification} and show that they become tighter as the number of particles increases. We, then benchmark our algorithms for planning in the setting of a given belief tree (Section \ref{sec:GivenTree}) and in an anytime MCTS setting (Section \ref{sec:MCTS}). In the former setting, we compare SITH-BSP and LAZY-BSP  against Sparse Sampling \citep{Kearns02jml}. In an anytime MCTS setting, we compare SITH-PFT with PFT-DPW \citep{Sunberg18icaps} and IPFT \citep{Fischer20icml}. This performance evaluation is conducted considering three problems, as discussed next.

\subsection{Problems under Consideration}\label{sec:Problems}
We proceed to the description of the evaluated problems.   In two first problems  the immediate reward for $b'$  is   
\begin{equation}
 \!\! \rho(b,a, z', b')=- (1 - \lambda )\!\!\!\underset{x' \sim b'}{\mathbb{E}}\Big[  r(x') \Big]\!- \! \lambda \hat{\mathcal{H}}(b,a, z', b'). \!\!\!
\end{equation}

\subsubsection{Continuous Light Dark}  \label{sec:LD}
Our first problem is \emph{2D continuous Light-Dark problem}. The robot starts at some unknown point $x_0\in \mathbb{R}^2$. In this world, there are spatially scattered beacons with known locations. Near the beacons, the obtained observations are less ``noisy''. The robot's mission is to get to the goal located at the upper right corner of the world. 
The state dependent reward in this problem  is $r(x) = -\|x - x^{\text{goal}}\|_2^2$. 
The initial belief is $b_0 = \mathcal{N}(\mu_0, I\cdot \sigma_0)$, where we select $x_0 = \mu_0$ for actual robot initial state. The motion and observation models are 
\begin{align}
\probd_T (x' | x, a)=\mathcal{N}(x+a, I\cdot \sigma_T), \label{eq:LDMotionModel}
\end{align}
and 
\begin{align}
\!\! \mathcal{O}=\probd_Z(z | x)=\mathcal{N}(x\! - \! x^b, I\!\cdot \! \sigma_{\mathcal{O}}\cdot \max\{d(x),d_{\mathrm{min}}\}), \label{eq:LDObsModel}
\end{align} 
respectively, where  $d(x)$ is the $\ell^2$ distance from robot's state  $x$ to the nearest beacon with known location denoted by $x^b$, and $d_{\mathrm{min}}$ is a tuneable parameter. 
\begin{figure*}[t!]
	\begin{minipage}[t]{0.49\textwidth}   
		\centering
		\includegraphics[width=\textwidth]{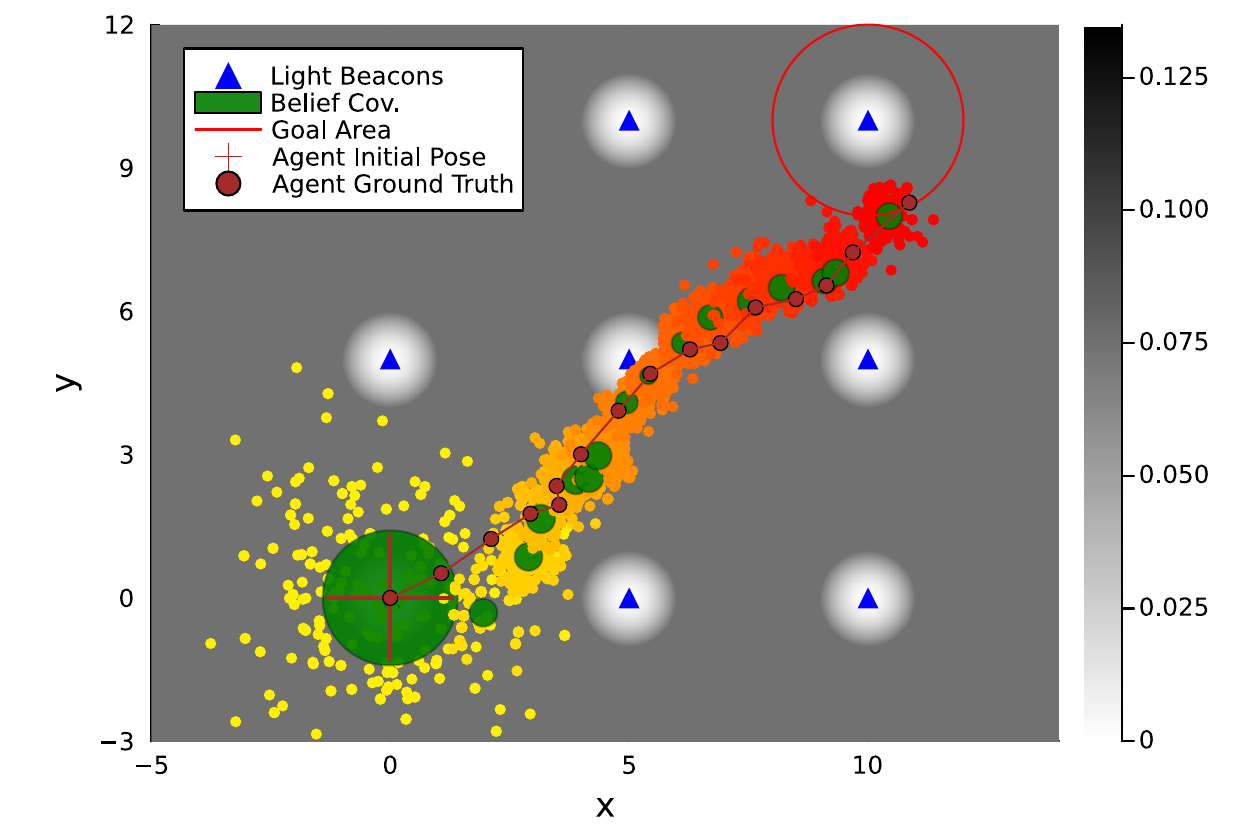}
		\subcaption{}	
		\label{fig:PassiveInferenceForBoundsStudy1}
	\end{minipage}
	\hfill
	\begin{minipage}[t]{0.49\textwidth}
		\centering
		\includegraphics[width=\textwidth]{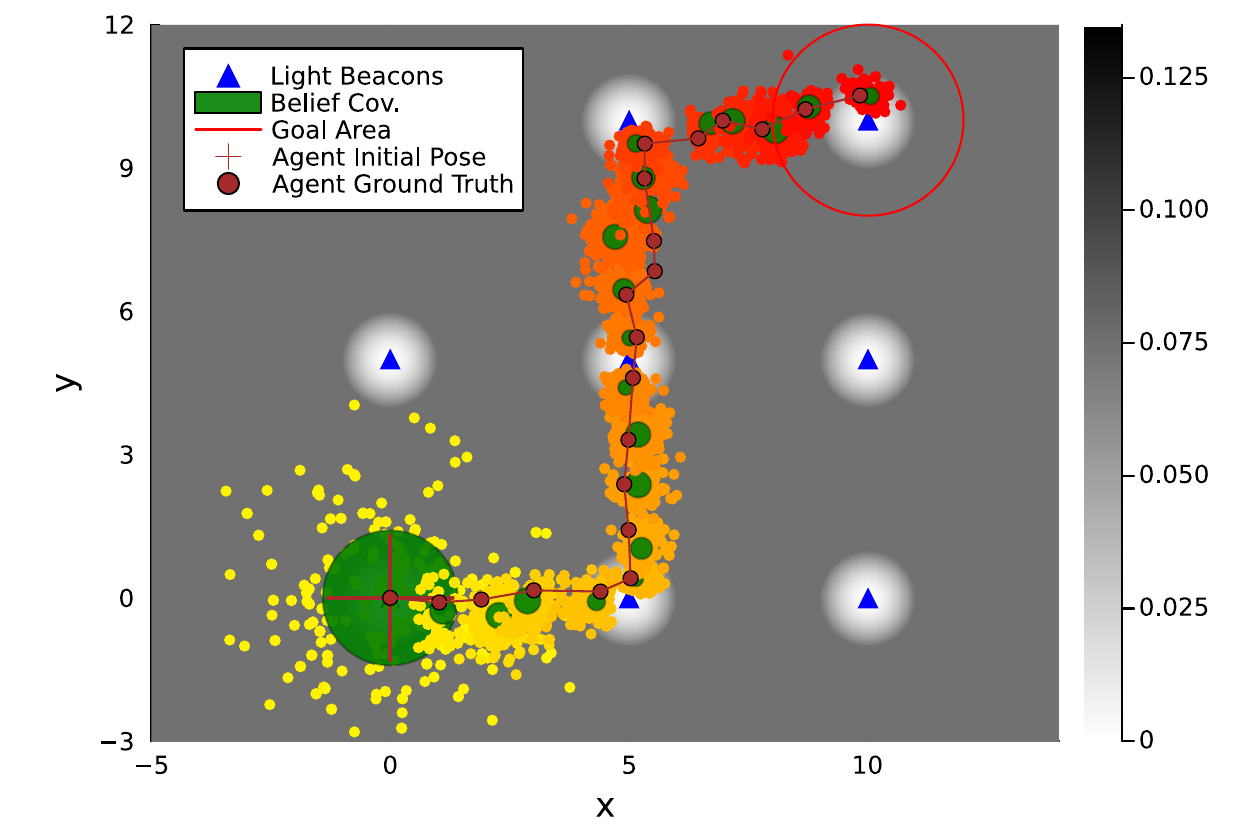}
		\subcaption{}
		\label{fig:PassiveInferenceForBoundsStudy2}	
	\end{minipage}
	\caption{The plot shows the evolution of belief in terms of sets of particles along the actual trajectory of the robot. The color of the particles from yellow to red illustrate the evolution of the belief over time. The green ellipses represent the parametric Gaussian belief covariances obtained from update by Kalman filter. The canvas color here is $\sigma_{\mathcal{O}}=\sigma_{T}=0.075$ as in equations \eqref{eq:LDMotionModel} and \eqref{eq:LDObsModel} respectively. 
	\textbf{(a)} Our first scenario. \textbf{(b)} Our second scenario.   }
\end{figure*}
\subsubsection{Target Tracking} \label{sec:TT}
Our second problem is \emph{2D continuous Target Tracking}. In this problem we have a moving target in addition to the agent. In this problem the belief is maintained over both positions, the agent and the target. The state dependent reward in this problem  is $r(x) = -\|x^{\text{agent}} - x^{\text{target}}\|_2^2$. 
The motion model of the target and the agent follows 
\begin{align}
&\probd_T(\cdot|x, a) = \mathcal{N}(x^{\text{agent}}\! + a^{\text{agent}}, \Sigma_T) \cdot \mathcal{N}(x^{\text{target}}\! + a^{\text{target}}, \Sigma_T), \nonumber
\end{align}	  
where by $x$ we denote the concatenated $\{x^{\text{agent}}, x^{\text{target}}\}$.   
For target actions we use a circular buffer with  $\{\uparrow, \uparrow,  \leftarrow \}$ action sequence of unit length motion primitives.  
For simplicity we assume that in inference as well as in the planning session the agent knows the target action sequence.   The observation model is also the multiplication of the observation model from the previous section with the additional observation model due to a moving target.  Thus, the overall observation model is 
\begin{equation}
	\nonumber
	\begin{gathered}
	\probd_Z (\cdot | x ; \{x^{b,i}\}_{i=1} )= \mathcal{N}(x^{\text{agent}}, \Sigma_\mathcal{O}(x^{\text{agent}}; \{x^{b,i}\}_{i=1}))\cdot \\
	\cdot \mathcal{N}(x^{\text{agent}} -  x^{\text{target}},\Sigma_\mathcal{O}(x^{\text{agent}},x^{\text{target}})), 
	\end{gathered}
\end{equation}
where $\Sigma_\mathcal{O}(x^{\text{agent}}; \{x^{b,i}\}_{i=1})$ conforms to the observation model covariance described in Section \ref{sec:LD} and 
\begin{align}
	&\Sigma_\mathcal{O}(x^{\text{agent}},x^{\text{target}}) =\!\!\\
	& \begin{cases}
	\!\sigma^2_{T}I   \| x^{\text{agent}} - x^{\text{target}}  \|_2, \text{ if } \| x^{\text{agent}} - x^{\text{target}}  \|_2 \geq \! d_{\text{min}} \\
	\!\sigma^2_{\mathcal{O}}I, \text{else}  \end{cases}.  \nonumber
\end{align}
Before the planning experiments we study of the entropy estimators and the bounds presented in Theorem~\ref{theorem:LowerUpperReward}.  
\subsubsection{Safe Autonomous Localization}
\label{sec:SafeAutonomousLocalization}
Our third problem is a variation of the problem presented in Section~\ref{sec:LD}. Here we change the reward to be the combination of localization reward and safety reward \citep{Zhitnikov22arxiv} 
\begin{equation}
\label{eq:SafeLocalizationReward}
\begin{gathered}
\rho(b,a, z', b')=\overbrace{- \hat{\mathcal{H}}(b,a, z', b')}^{\text{localization reward}} + \\
+\underbrace{s \Big(2 \cdot \mathbf{1}_{\{\mathrm{P}(\{x^{\prime} \in \mathcal{X}^{\mathrm{safe},\prime}\}|b^{\prime}) \geq \delta \}}(b^{\prime}) - 1 \Big)}_{\text{safety reward}}.
\end{gathered} 
\end{equation}
Such a safety reward divides the candidate actions into two sets, the safe set and the unsafe. If the safety parameter $s$ is sufficiently large to assure that safe action is selected, these two sets are detached enough in terms of safety reward and the unsafe set is substantially inferior such that there is no point to calculate localization reward precisely over this set of actions. There, we can, without any harm for decision-making outcome, substitute differential entropy by the bounds at the low simplification levels.                   
This aspect makes the simplification paradigm invaluable.     
\subsection{Entropy Estimators and Bounds Study}
In this section, we experiment with a passive case of the continuous 2D Light Dark problem from Section \ref{sec:LD}. Our goal is to study the various entropy estimators and our derived bounds from Section \ref{sec:SpecificSimplification} over the estimator developed in  \cite{Boers10fusion}. In this study, we manually supply the robot with an action sequence to conduct. This results in a single lace of the beliefs corresponding to observations that the robot actually obtained by executing a given externally action sequence. We also provide some attempt in this section to compare estimated reward with the exact analytical counterpart. 

Over this sequence of the beliefs, at each time instance of the sequence we calculate minus differential entropy estimator (information) in four ways. 
The first is the Boers estimator \citep{Boers10fusion} and our bounds from Theorem~\ref{theorem:LowerUpperReward}. The second is KDE approximation as done by \cite{Fischer20icml}. The third is the naive calculation of discrete entropy over the the particles weights: $\hat{\mathcal{H}}(b)=-\sum_i w^i \cdot \log w^i$. The fourth estimator is analytical and it requires additional explanation. If we make an unrealistic assumption that robot's ground truth state from which the observation has been taken is known, plug it into the covariance matrix of \eqref{eq:LDObsModel} and set prior belief to be Gaussian; the motion and observation models met all the requirement for the exact update by Kalman Filter (linear additive models). For the proof see \cite{Thrun05book}. In this case the belief stays Gaussian and the differential entropy has closed form solution.   

We have two scenarios. In the first scenario,    
the robot moves diagonally to the goal using a unit length action $\nearrow$ (Fig.~\ref{fig:PassiveInferenceForBoundsStudy1}) fifteen times. Along the way, it passes close-by two beacons. Consequentially, the robot's information about its state peaks twice.  In our second scenario the robot moves five times to the right $\rightarrow$ followed by ten times $\uparrow$ and again five times to the right $\rightarrow$ (Fig.~\ref{fig:PassiveInferenceForBoundsStudy2}). 
\begin{figure}[t!]
	\begin{minipage}[t]{\columnwidth}   
		\centering         
		\includegraphics[width=\columnwidth]{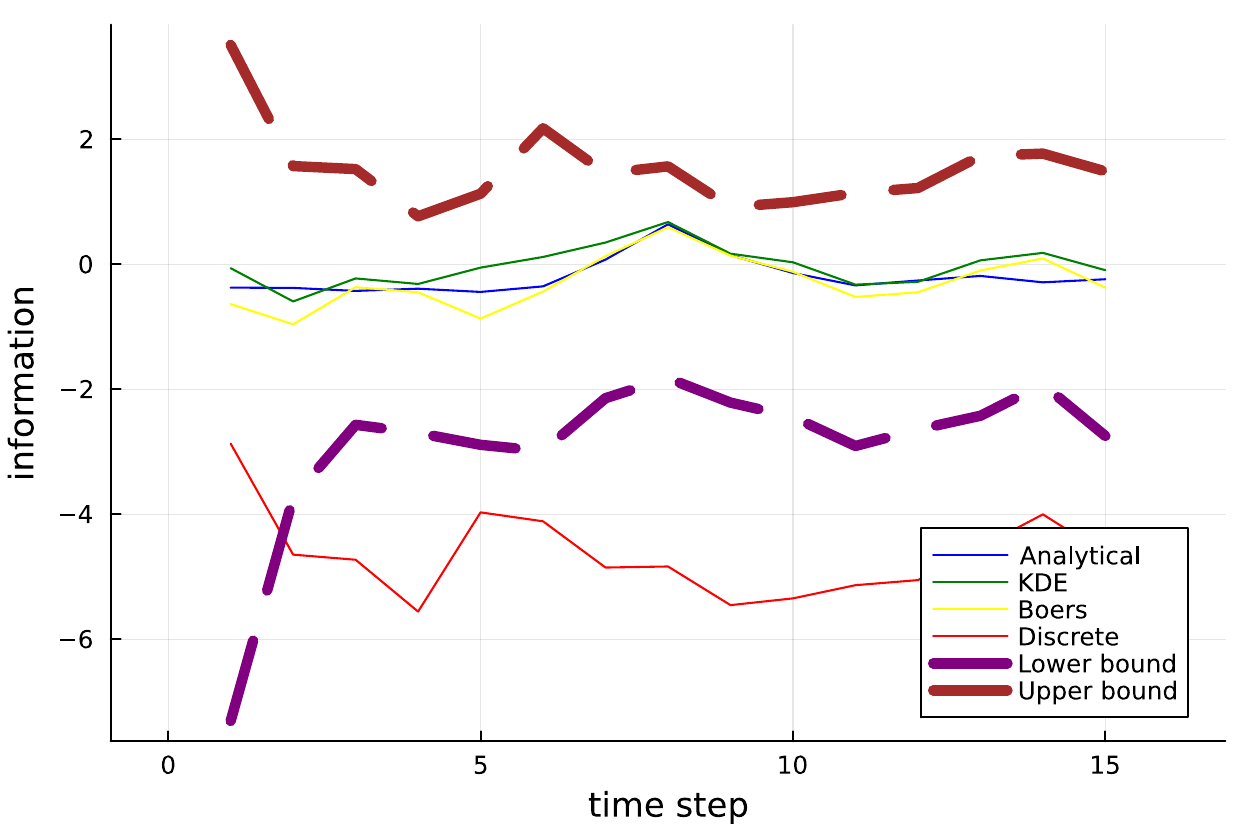}
		\subcaption{}
		\label{fig:bounds1scenario1}
	\end{minipage}
	\begin{minipage}[t]{\columnwidth}
		\centering
		\includegraphics[width=\textwidth]{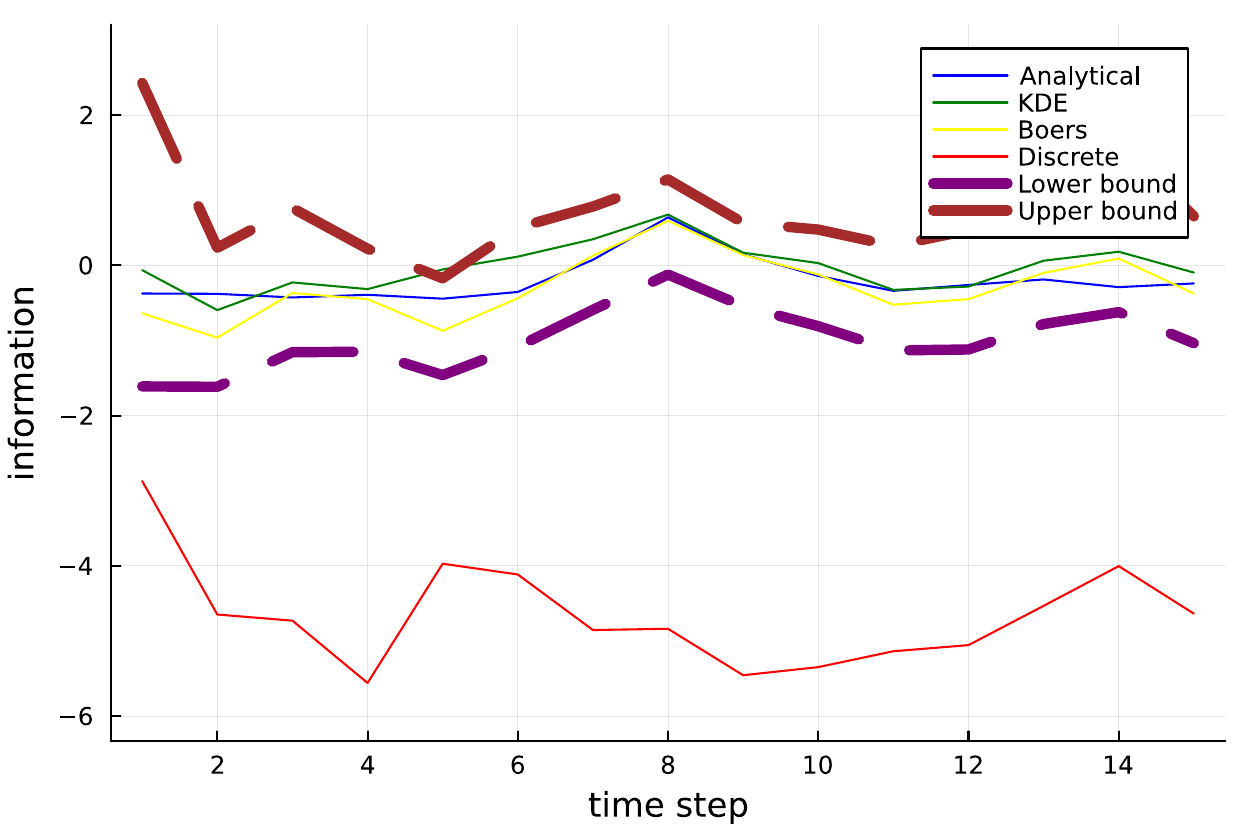}
		\subcaption{}	
		\label{fig:bounds2scenario1}	
	\end{minipage}
	\begin{minipage}[t]{\columnwidth}   
		\centering         
		\includegraphics[width=\textwidth]{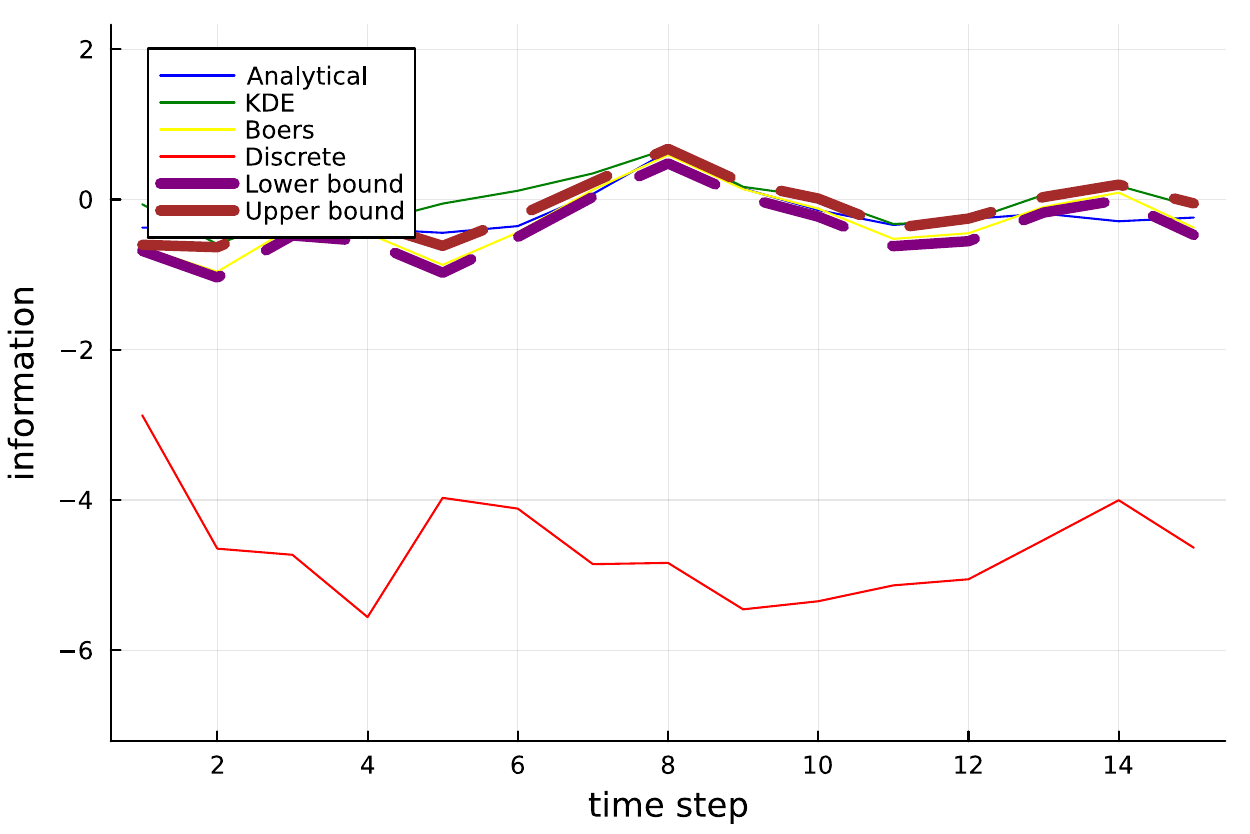}
		\subcaption{}
		\label{fig:bounds3scenario1}
	\end{minipage}
	\caption{Bounds convergence for our first scenario $n_x=300$ \textbf{(a)} $n^s_x=30$ particles    \textbf{(b)} $n^s_x=150$ particles  \textbf{(c)} $n^s_x=270$ particles. }
	\label{fig:BoundsStudyscenario1}
\end{figure}
\begin{figure}[t!]
	\begin{minipage}[t]{\columnwidth}   
		\centering         
		\includegraphics[width=\columnwidth]{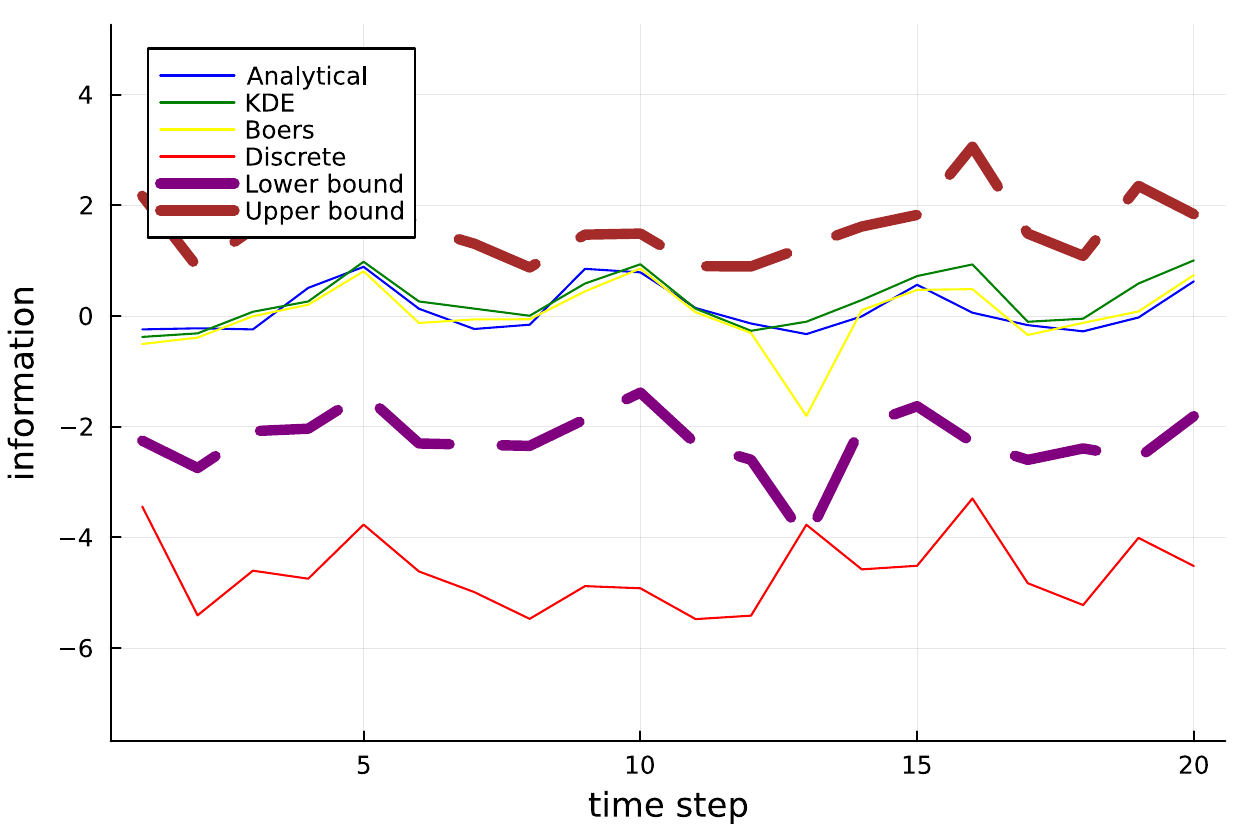}
		\subcaption{}
		\label{fig:bounds1scenario2}
	\end{minipage}
	\hfill
	\begin{minipage}[t]{\columnwidth}
		\centering
		\includegraphics[width=\textwidth]{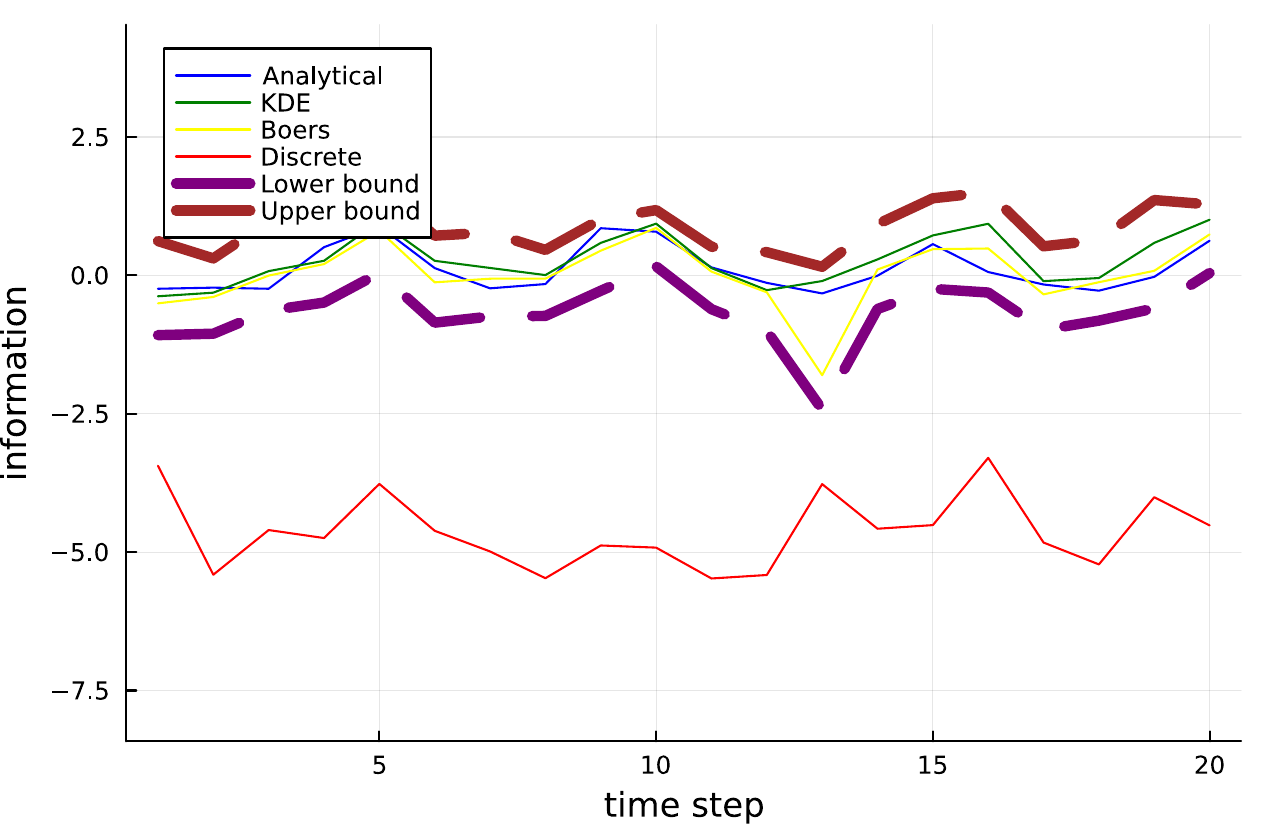}
		\subcaption{}	
		\label{fig:bounds2scenario2}	
	\end{minipage}
	\hfill
	\begin{minipage}[t]{\columnwidth}   
		\centering         
		\includegraphics[width=\textwidth]{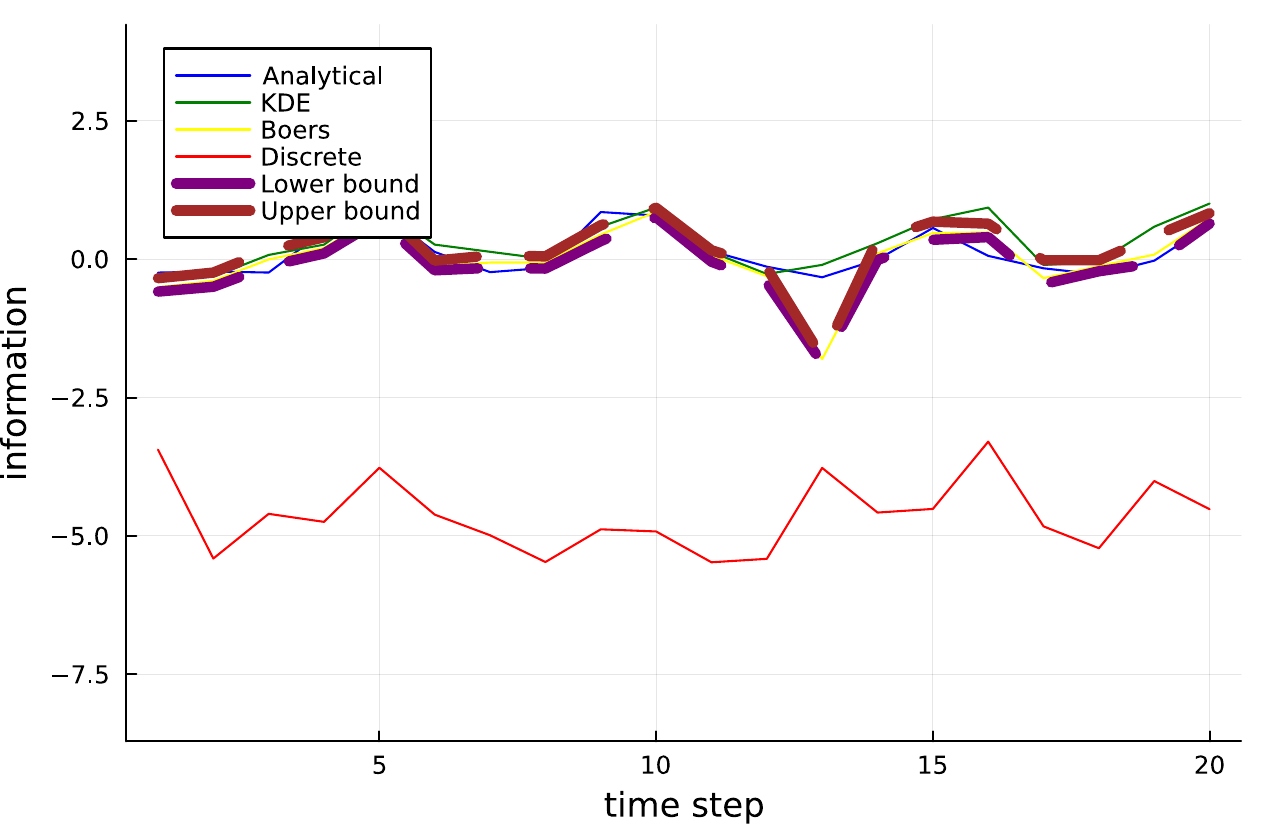}
		\subcaption{}
		\label{fig:bounds3scenario2}
	\end{minipage}
	\caption{Bounds convergence for our second scenario  $n_x=300$ \textbf{(a)} $n^s_x = 30$     \textbf{(b)} $n^s_x =150$ particles  \textbf{(c)} $n^s_x =270$ particles. } 
	\label{fig:BoundsStudyscenario2}
\end{figure} 

The prior belief in this setting follows a Gaussian distribution $b_0=\mathcal{N}\begin{pmatrix}\begin{pmatrix} 0.0 \\ 0.0 \end{pmatrix},  \begin{pmatrix} 2.0 & 0.0 \\ 0.0 & 2.0 \end{pmatrix}\end{pmatrix}$, the motion and observation models parameters are $\sigma_{\mathcal{O}}=\sigma_{T}=0.075, d_{\text{min}}= 0.0001$.   The number of unsimplified belief weighted particles is $n_x = 300$. For creating initial weighted particles we use the following proposal 
\begin{equation}
	\nonumber
	\begin{gathered}
	q = 0.25 \cdot \mathcal{N}\begin{pmatrix}\begin{pmatrix} 0.0 \\ 1.0 \end{pmatrix},  \begin{pmatrix} 2.0 & 0.0 \\ 0.0 & 0.2 \end{pmatrix}\end{pmatrix} + \\
	+ 0.25 \cdot \mathcal{N}\begin{pmatrix}\begin{pmatrix} 1.0 \\ 0.0 \end{pmatrix},  \begin{pmatrix} 2.0 & 0.0 \\ 0.0 & 0.2 \end{pmatrix}\end{pmatrix} + \\
    +0.25 \cdot \mathcal{N}\begin{pmatrix}\begin{pmatrix} -1.0 \\ 0.0 \end{pmatrix},  \begin{pmatrix} 2.0 & 0.0 \\ 0.0 & 0.2 \end{pmatrix}\end{pmatrix} + \\
    + 0.25 \mathcal{N}\begin{pmatrix}\begin{pmatrix} 1.0 \\ -1.0 \end{pmatrix},  \begin{pmatrix} 2.0 & 0.0 \\ 0.0 & 0.2 \end{pmatrix}\end{pmatrix}. \nonumber
\end{gathered}
\end{equation}
The initial weights are the ratio $w(x)=\frac{b_0(x)}{q(x)}$.    

To examine the bounds monotonical convergence with a growing number of simplified belief particles  we plot the bounds \eqref{eq:immediate_bounds_l} and \eqref{eq:immediate_bounds_u} for minus entropy estimator \eqref{eq:BoersDiffEnt} alongside  estimators  described above for the entire robot trajectory of the beliefs.

The results for the first and second scenarios are provided in Figs.~\ref{fig:BoundsStudyscenario1} and \ref{fig:BoundsStudyscenario2}, respectively.
For both scenarios we observe that the bounds become tighter as the number of particles of simplified belief $n^s_x$ increases. We also witness that all estimators vary but the overall trend is similar, putting aside the discrete entropy over the weights. The discrete entropy over the weights fails to adequately represent the uncertainty of the belief. This is an anticipated result. Let us proceed to the planning experiments.   
\subsection{Planning}
In this section we study and benchmark our efficient planning algorithms. In our algorithms \ref{alg:sith-bsp} and \ref{alg:lazy-sith-bsp} the tree is build by SS \citep{Kearns02jml} such that the given belief tree is obtained when the algorithm descends to the leafs. We first compare Alg.~\ref{alg:sith-bsp} and \ref{alg:lazy-sith-bsp} versus SS. We then proceed to simulations in an anytime MCTS setting.  

For all further experiments, the belief is approximated by a set of $n_x$ weighted samples as in \eqref{eq:BeliefParticles}. The robot does replanning after each executed action.  

\subsubsection{Acceleration measures}
Let us begin this section by describing our measures of acceleration. We report planning time speedup in terms of saved accesses to particles.

The following speedup is based on the final number of simplified beliefs particles required for planning   
\begin{align}
\frac{\sum_{i} \Big(n_{i,x}^2 - n_{i,x}^s n_{i,x}\Big)}{\sum_{i} n_{i,x}^2} \cdot 100, \label{eq:ParticlesSpeedup}
\end{align}
where the summation is over the future posterior beliefs in all the belief trees in a number of a consecutive planning sessions in particular scenario.     Eq.~\eqref{eq:ParticlesSpeedup} measures relative speedup without time spent on resimplifications. 
It is calculated at the end of several consecutive planning sessions.  To calculate speedup according to \eqref{eq:ParticlesSpeedup} one shall pick up the {\bf final} number of particles of simplified belief  $n^s_{i,x}$ used for the simplified reward for each belief node $i$, sum over all the nodes of the belief trees (given or constructed on the fly) from planning sessions, make a calculation portrayed by \eqref{eq:ParticlesSpeedup}.  Importantly, acceleration measure \eqref{eq:ParticlesSpeedup} assumes that time of evaluating the motion and observation models do not vary from one evaluation to another. If the number of belief particles is not not depending on the belief ($n_{i,x} = n_x$) we can further simplify the \eqref{eq:ParticlesSpeedup} to
\begin{align}
\frac{\sum_{i} \Big(n_{x} - n_{i,x}^s \Big)}{\sum_{i} n_{x}} \cdot 100.
\end{align}

To calculate planning time speedup we use the following metric 
\begin{align}
\frac{t_{\mathrm{baseline}} - t_{\mathrm{our}}}{t_{\mathrm{baseline}}} \cdot 100. \label{eq:speedup}
\end{align} 
If the quantities \eqref{eq:ParticlesSpeedup} and \eqref{eq:speedup}  are identical we can conclude that there will be no overhead from resimplifications and adapting the bounds. 
Note also that in the first place it is not clear how many particles $n_x$ for belief representation to take. The number of particles $n_x$ shall be as large as possible due to fact that we do not know when the belief represented by weighted particles will converge to the corresponding theoretical belief. 

To thoroughly study the acceleration yielded by our simplification paradigm  we calculate total speedup over a number of the consecutive planning sessions in terms of particles in accordance to \eqref{eq:ParticlesSpeedup} and in terms of time in accordance to \eqref{eq:speedup}.

\subsubsection{Results for 2D Continuous Light-Dark in the Setting of a Given Belief Tree}\label{sec:PlanningLightDark}
\begin{figure}[t!]  
	\centering         
	\includegraphics[width=0.5\textwidth]{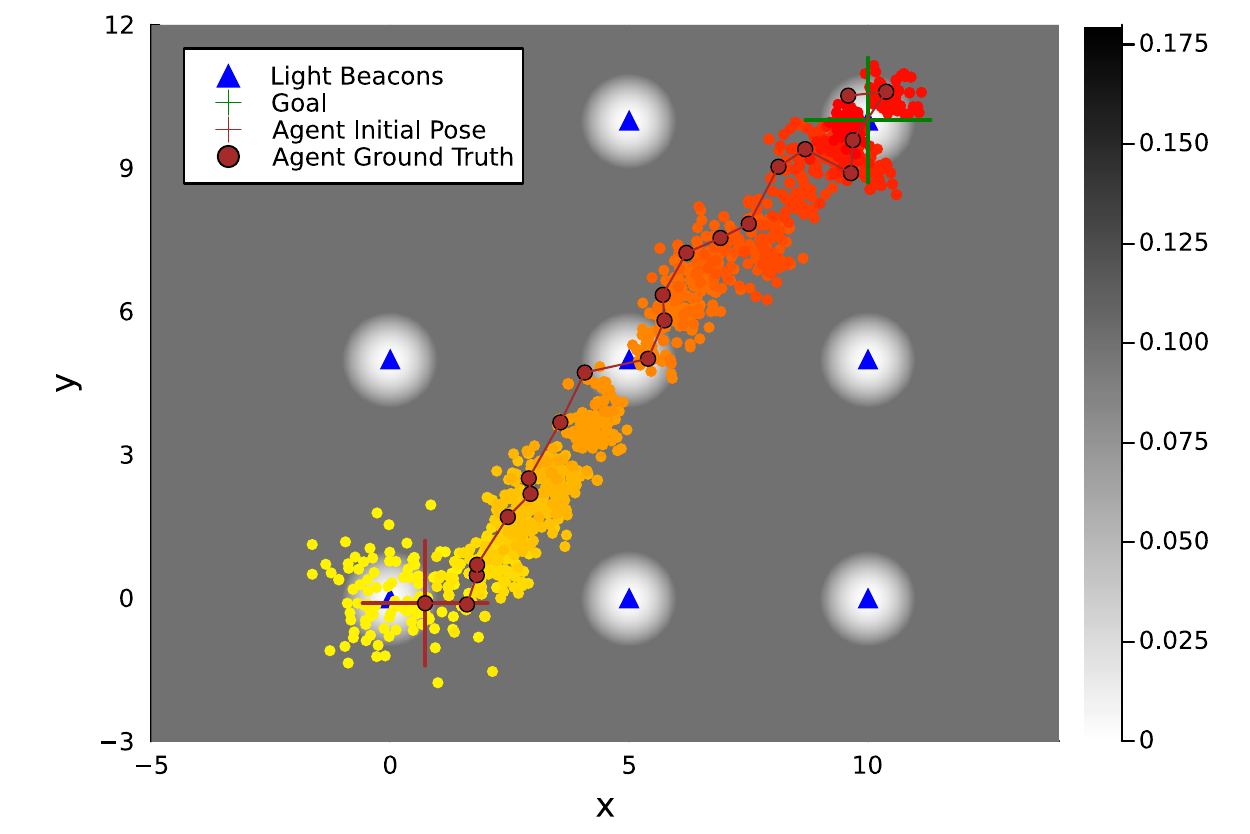}
	\caption{ Exemplary 2D Light Dark problem planning scenario. Here we present the first trial of configuration $\lambda=0.5$ of Table~\ref{tbl:AblationStudyL3StaticGoalLambda}.}
	\label{fig:PlanningLightDark}
\end{figure} 
\begin{table*}[t!]
	\setlength\extrarowheight{3pt}
	\caption{This table shows cumulative results of $20$ consecutive alternating planning and execution sessions of the Continuous Light Dark problem averaged over $15$ trials. Each planning session creates a single belief tree to perform a search for optimal action. This given belief tree has $4809$ belief nodes. Overall, in $20$ planning sessions, we have $96180$ belief nodes. The horizon in each planning session is $L=3$. The number of observations sampled from each belief action node is $n^1_z =1$, $n^2_z =3$, $n_z^3=3$ at the corresponding to superscripts depths $1,2,3$. This table examines the influence of various values of $\lambda$. }
	\centering
	\resizebox{\textwidth}{!}{
		\begin{tabular}{|c|c|c|c|c|c|c|c|c|}
			\hline
			BSP Alg.& $n_x$ &    $\lambda$   &particles speedup \eqref{eq:ParticlesSpeedup} & time speedup \eqref{eq:speedup} & resimpl. calls (recursive) & motion model calls & obs. model calls & return ($\hat{V}$)\\ 
			\hline  
			Alg~\ref{alg:sith-bsp} \ SITH &\multirow{3}{*}{$ 100 $}  & 
			\multirow{3}{*}{$ 0.1$} &  
			$ 78.76 \pm  0.20$ & $ 64.44 \pm 1.51 $ & 
			$ 2.05 \cdot 10^5 \pm  0.05 \cdot 10^5  $& 
			$ 3.13 \cdot 10^8 \pm  0.02 \cdot 10^8 $&
			$ 9.62 \cdot 10^6 \pm  0.0 $&
			$ -115.49 \pm  16.58 $  
			\\
			\cline{1-1} \cline{4-9} 
			Alg~\ref{alg:lazy-sith-bsp} \ LAZY & &&
			$ 85.46 \pm  1.22$ & $ 71.59 \pm 1.52 $ & 
			$ 10.71\cdot 10^5 \pm  4.61 \cdot 10^5 $ & 
			$ 2.38 \cdot 10^8 \pm  0.13 \cdot 10^8 $ &
			$ 9.62 \cdot 10^6 \pm  0.0 $ &
			$ -115.49 \pm  16.58 $  
			\\
			\cline{1-1} \cline{4-9}
			SS  &&&&&&
			$ 9.62 \cdot 10^8 \pm  0.0 $ &
			$ 9.62 \cdot 10^6 \pm  0.0 $ &
			$ -115.49 \pm  16.58 $   \\
			\hline  
			Alg~\ref{alg:sith-bsp} \ SITH &\multirow{3}{*}{$ 100 $} &  
			\multirow{3}{*}{$ 0.2$} &     
			$ 68.82 \pm  0.32$ & $ 53.59 \pm 2.05 $ & 
			$ 3.36 \cdot 10^5 \pm  0.06 \cdot 10^5  $& 
			$ 4.22 \cdot 10^8 \pm  0.03 \cdot 10^8 $&
			$ 9.62 \cdot 10^6 \pm  0.0 $&
			$ -103.51 \pm  14.91 $  
			\\
			\cline{1-1} \cline{4-9} 
			Alg~\ref{alg:lazy-sith-bsp} \ LAZY & &&
			$ 80.09 \pm  1.52$ & $ 65.01 \pm 1.88 $ & 
			$ 25.65 \cdot 10^5 \pm  6.17\cdot 10^5 $ & 
			$ 3.01 \cdot 10^8 \pm  0.18 \cdot 10^8 $ &
			$ 9.62 \cdot 10^6 \pm  0.0 $ &
			$ -103.51 \pm  14.91 $  
			\\
			\cline{1-1} \cline{4-9}
			SS &&&&&&
			$ 9.62  \cdot 10^8 \pm  0.0 $ &
			$ 9.62 \cdot 10^6 \pm  0.0 $ &
			$ -103.51 \pm  14.91 $   \\
			\hline 
			Alg~\ref{alg:sith-bsp} \ SITH &\multirow{3}{*}{$ 100 $} &  
			\multirow{3}{*}{$ 0.3$} &  
			$ 58.33 \pm  0.52$ & $ 42.76 \pm 2.96 $ & 
			$ 4.13 \cdot 10^5 \pm  0.05 \cdot  10^5 $& 
			$ 5.40 \cdot 10^8 \pm  0.01 \cdot 10^8 $&
			$ 9.62 \cdot 10^6 \pm  0.0 $&
			$ -91.86 \pm  13.88 $  
			\\
			\cline{1-1}   \cline{4-9}
			Alg~\ref{alg:lazy-sith-bsp} \ LAZY & &&
			$ 74.85 \pm  2.63$ & $ 58.94 \pm 3.04 $ & 
			$ 42.66 \cdot 10^5 \pm  9.80 \cdot 10^5$ & 
			$ 3.59  \cdot 10^8 \pm  0.29 \cdot 10^8 $ &
			$ 9.62 \cdot 10^6 \pm  0.0 $ &
			$ -91.86 \pm  13.88 $  
			\\
			\cline{1-1}  \cline{4-9}
			SS &&&&&& 
			$ 9.62 \cdot 10^8 \pm  0.0 $ &
			$ 9.62 \cdot 10^6 \pm  0.0 $ &
			$ -91.86 \pm  13.88 $   \\
			\hline 
			Alg~\ref{alg:sith-bsp} \ SITH &\multirow{3}{*}{$ 100 $} & 
			\multirow{3}{*}{$ 0.4$} &    
			$ 45.66 \pm  0.83$ & $ 29.33 \pm 4.78 $ & 
			$ 4.70 \cdot 10^5 \pm  0.04 \cdot 10^5  $& 
			$ 6.84 \cdot 10^8 \pm  0.08 \cdot 10^8 $&
			$ 9.62 \cdot 10^6 \pm  0.0 $&
			$ -80.44 \pm  11.77 $  
			\\
			\cline{1-1} \cline{4-9}
			Alg~\ref{alg:lazy-sith-bsp} \ LAZY & &&
			$ 69.94 \pm  1.89$ & $ 53.85 \pm 2.56 $ & 
			$ 59.05\cdot 10^5 \pm  8.76 \cdot 10^5 $ & 
			$ 4.16\cdot 10^8 \pm  0.22  \cdot 10^8 $ &
			$ 9.62 \cdot 10^6 \pm  0.0 $ &
			$ -80.44 \pm  11.77 $  
			\\
			\cline{1-1} \cline{4-9}
			SS &&&&&&
			$ 9.62 \cdot 10^8 \pm  0.0 $ &
			$ 9.62 \cdot 10^6 \pm  0.0 $ &
			$ -80.44 \pm  11.77 $   \\
			\hline 
			Alg~\ref{alg:sith-bsp} \ SITH &\multirow{3}{*}{$ 100 $} &
			\multirow{3}{*}{$ 0.5$} &     
			$ 34.46 \pm  0.79$ & $ 18.98 \pm 4.16 $ & 
			$ 5.27 \cdot 10^5 \pm  0.05 \cdot 10^5  $& 
			$ 7.92 \cdot 10^8 \pm  0.01 \cdot 10^8 $&
			$ 9.62 \cdot 10^6 \pm  0.0 $&
			$ -66.3 \pm  8.0 $  
			\\
			\cline{1-1} \cline{4-9}
			Alg~\ref{alg:lazy-sith-bsp} \ LAZY & &&
			$ 63.6 \pm  2.23$ & $ 46.67 \pm 2.81 $ & 
			$ 81.48 \cdot 10^5 \pm  8.52 \cdot 10^5 $ & 
			$ 4.87 \cdot 10^8 \pm  0.24 \cdot 10^8 $ &
			$ 9.62 \cdot 10^6 \pm  0.0 $ &
			$ -66.3 \pm  8.0 $  
			\\
			\cline{1-1} \cline{4-9}
			SS &&&&&&
			$ 9.62 \cdot 10^8 \pm  0.0 $ &
			$ 9.62 \cdot 10^6 \pm  0.0 $ &
			$ -66.3 \pm  8.0 $   \\
			\hline 
			Alg~\ref{alg:sith-bsp} \ SITH &\multirow{3}{*}{$ 100 $} &  
			\multirow{3}{*}{$ 0.6$} &    
			$ 25.09 \pm  0.89$ & $ 12.05 \pm 4.83 $ & 
			$ 5.85 \cdot 10^5 \pm  0.05 \cdot 10^5  $& 
			$ 8.64 \cdot 10^8 \pm  0.05 \cdot 10^8 $&
			$ 9.62 \cdot 10^6 \pm  0.0 $&
			$ -55.36 \pm  6.93 $  
			\\
			\cline{1-1} \cline{4-9} 
			Alg~\ref{alg:lazy-sith-bsp} \ LAZY & &&
			$ 56.32 \pm  2.72$ & $ 38.45 \pm 3.65 $ & 
			$ 113.26 \cdot 10^5 \pm  11.45\cdot 10^5 $ & 
			$ 5.71 \cdot 10^8 \pm  0.28 \cdot 10^8$ &
			$ 9.62 \cdot 10^6 \pm  0.0 $ &
			$ -55.36 \pm  6.93 $  
			\\
			\cline{1-1} \cline{4-9}
			SS &&&&&&
			$ 9.62 \cdot 10^8 \pm  0.0 $ &
			$ 9.62 \cdot 10^6 \pm  0.0 $ &
			$ -55.36 \pm  6.93 $   \\
			\hline 
		\end{tabular}
	}
	\label{tbl:AblationStudyL3StaticGoalLambda} 
\end{table*}
\begin{table*}[t!]
	\setlength\extrarowheight{3pt}
	\caption{This table shows cumulative results of $20$ consecutive alternating planning and execution sessions averaged over $15$ trials of Continuous Light Dark problem. The given belief tree in a single planning session has $4809$ belief nodes. Overall, in $20$ planning sessions, we have $96180$ belief nodes. The horizon in each planning session  is $L=3$. The number of observations sampled from each belief action node is $n^1_z =1$, $n^2_z =3$, $n_z^3=3$ at the corresponding to superscripts depths $1,2,3$. In this table we examine influence of various number of belief particles.}
	\centering
	\resizebox{\textwidth}{!}{
		\begin{tabular}{|c|c|c|c|c|c|c|c|c|}
			\hline
			BSP Alg.& $n_x$  &  $\lambda$   &  particles speedup \eqref{eq:ParticlesSpeedup} & time speedup \eqref{eq:speedup} & resimpl. calls (recursive) & motion model calls & obs. model calls & return ($\hat{V}$)\\ 
			\hline  
			Alg~\ref{alg:sith-bsp} \ SITH &\multirow{3}{*}{$  200$} &  
			\multirow{3}{*}{$ 0.5$} &  
			$ 34.1 \pm  0.8$ & $ 25.01 \pm 5.11 $ & 
			$ 5.30 \cdot 10^5 \pm  0.04 \cdot 10^5  $& 
			$ 31.80 \cdot 10^8 \pm  0.25 \cdot 10^8 $&
			$ 19.24 \cdot 10^6 \pm  0.0 $&
			$ -69.36 \pm  7.95 $  
			\\
			\cline{1-1} \cline{4-9} 
			Alg~\ref{alg:lazy-sith-bsp} \ LAZY &&&
			$ 64.0 \pm  2.98$ & $ 51.71 \pm 4.9 $ & 
			$ 83.95 \cdot 10^5 \pm  10.24 \cdot 10^5 $ & 
			$ 19.35 \cdot 10^8 \pm  1.29 \cdot 10^8 $ &
			$ 19.24 \cdot 10^6 \pm  0.0 $ &
			$ -69.36 \pm  7.95 $  
			\\
			\cline{1-1} \cline{4-9}
			SS &&&&&&
			$ 38.47 \cdot 10^8 \pm  0.0 $ &
			$ 19.24 \cdot 10^6 \pm  0.0 $ &
			$ -69.36 \pm  7.95 $   \\
			\hline 
			Alg~\ref{alg:sith-bsp} \ SITH &\multirow{3}{*}{$  300$} & 
			\multirow{3}{*}{$ 0.5$} &    
			$ 33.84 \pm  0.83$ & $ 18.67 \pm 2.02 $ & 
			$ 5.30 \cdot 10^5 \pm  0.04 \cdot 10^5 $& 
			$ 71.74 \cdot 10^8 \pm  0.58 \cdot 10^8 $&
			$ 28.85 \cdot 10^6 \pm  0.0 $&
			$ -68.29 \pm  8.42 $  
			\\
			\cline{1-1} \cline{4-9} 
			Alg~\ref{alg:lazy-sith-bsp} \ LAZY &&& 
			$ 63.39 \pm  3.44$ & $ 47.34 \pm 3.72 $ & 
			$ 84.09 \cdot 10^5 \pm  10.66 \cdot 10^5 $ & 
			$ 43.91 \cdot 10^8 \pm  3.09 \cdot 10^8 $ &
			$ 28.85 \cdot 10^6 \pm  0.0 $ &
			$ -68.29 \pm  8.42 $  
			\\
			\cline{1-1} \cline{4-9}
			SS &&&&&&
			$ 86.56 \cdot 10^8 \pm  0.0 $ &
			$ 28. 85 \cdot 10^6 \pm  0.0 $ &
			$ -68.29 \pm  8.42 $   \\
			\hline 
			Alg~\ref{alg:sith-bsp} \ SITH &\multirow{3}{*}{$  400$} &  
			\multirow{3}{*}{$ 0.5$} &  
			$ 33.97 \pm  0.85$ & $ 25.34 \pm 3.44 $ & 
			$ 6.65 \cdot 10^5 \pm  0.06 \cdot 10^5  $& 
			$ 181.50 \cdot 10^8 \pm  1.41  \cdot 10^8 $&
			$ 54.51 \cdot 10^6 \pm  0.0 $&
			$ -67.92 \pm  11.52 $  
			\\
			\cline{1-1} \cline{4-9} 
			Alg~\ref{alg:lazy-sith-bsp} \ LAZY &&& 
			$ 66.06 \pm  2.3$ & $ 53.74 \pm 3.4 $ & 
			$ 106.90\cdot 10^5 \pm  15.73 \cdot 10^5 $ & 
			$ 105.35 \cdot 10^8 \pm  5.75  \cdot 10^8 $ &
			$ 54.51 \cdot 10^6 \pm  0.0 $ &
			$ -67.92 \pm  11.52 $  
			\\
			\cline{1-1} \cline{4-9}
			SS &&&&&&
			$ 218.05 \cdot 10^8 \pm  0.0 $ &
			$ 54.51 \cdot 10^6 \pm  0.0 $ &
			$ -67.92 \pm  11.52 $   \\
			\hline
		\end{tabular}
	}
	\label{tbl:AblationStudyL3StaticGoalParticles} 
\end{table*}
We start from the problem described in Section \ref{sec:LD}.
Our action space is constituted by  motion primitives of unit length
$\mathcal{A} = \{  \rightarrow, \nearrow, \uparrow, \nwarrow,\leftarrow,  \swarrow,   \downarrow, \searrow\}$.
In this problem the selected parameters are  $\sigma_T = \sigma_\mathcal{O} = 0.1$, $d_{\text{min}} = 0.0001$, $\gamma = 0.95$.  We simulate $15$ trials of $20$ consecutive alternating planning and action execution sessions.  Fig.~\ref{fig:PlanningLightDark} shows an exemplary trial of $20$  executions of the best action identified by the robot.

We investigate the influence of the parameter $\lambda$ on speedup  in Table~\ref{tbl:AblationStudyL3StaticGoalLambda} and the impact of changing the number of particles in Table~\ref{tbl:AblationStudyL3StaticGoalParticles}.  In both tables we see the particles speedup (column $4$) and the time speedup (column $5$). As expected with increasing values of $\lambda$ (column $3$) the speedup diminishes. LAZY-BSP (Alg.~\ref{alg:lazy-sith-bsp}) produces larger speedup in terms of particles (column $4$) and time (column $5$) than SITH-BSP (Alg.~\ref{alg:sith-bsp}).  All three algorithms always selected the same optimal action. We observe that the return is always identical (column $9$).  Significant time speedup is obtained in the range of  $35\%-70 \%$  for LAZY-BSP depending on the values of $\lambda$. For the SITH-BSP  we see less time speedup ranging from $65\%$ to $10\%$ with increasing $\lambda$.    

In all tables the number of motion and observation model calls does not include belief update calls but only the calls for reward or bounds calculation.
The number of accesses to the observation model is always the same for all three algorithms (column $8$). This agrees with the structure of the bounds \eqref{eq:immediate_bounds_l}  and \eqref{eq:immediate_bounds_u}. For the baseline SS, up to rounding errors, the number of motion model accesses, as we anticipated, is the squared number of unsimplified belief particles multiplied by number belief nodes in the tree minus one for root belief, multiplied by number of planning sessions (column $7$ in the tables). This is in agreement with \eqref{eq:BoersDiffEnt}.  Also, for all three algorithms  the number of accesses to the observation model was the number of particles of unsimplified belief minus one for root belief, multiplied by the number of belief nodes in the tree, multiplied by the number of planning sessions.

\begin{figure*}[t!]
	\begin{minipage}[t]{0.49\textwidth}   
		\centering
		\includegraphics[width=\textwidth]{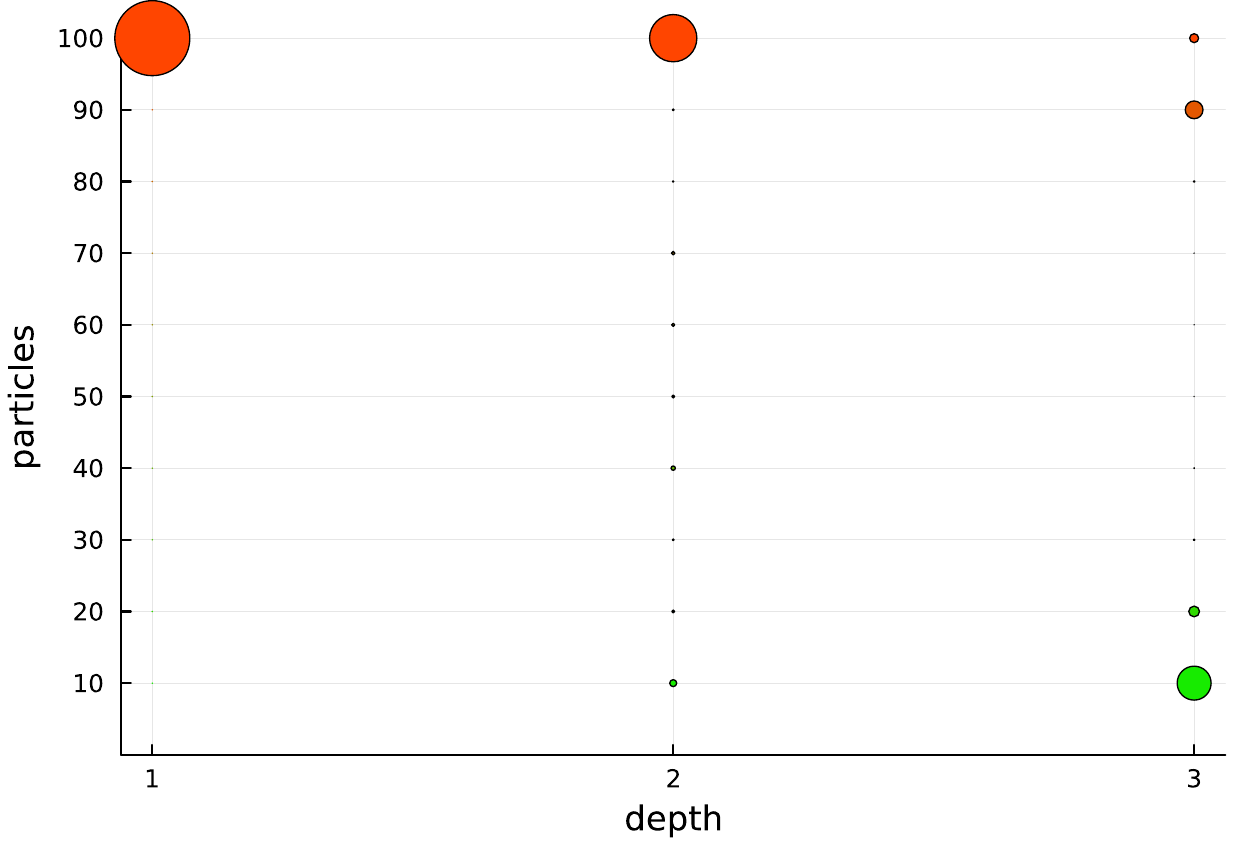}
		\subcaption{}	
		\label{fig:SimplifcationLevelsLAZY}
	\end{minipage}
	\hfill
	\begin{minipage}[t]{0.49\textwidth}
		\centering
		\includegraphics[width=\textwidth]{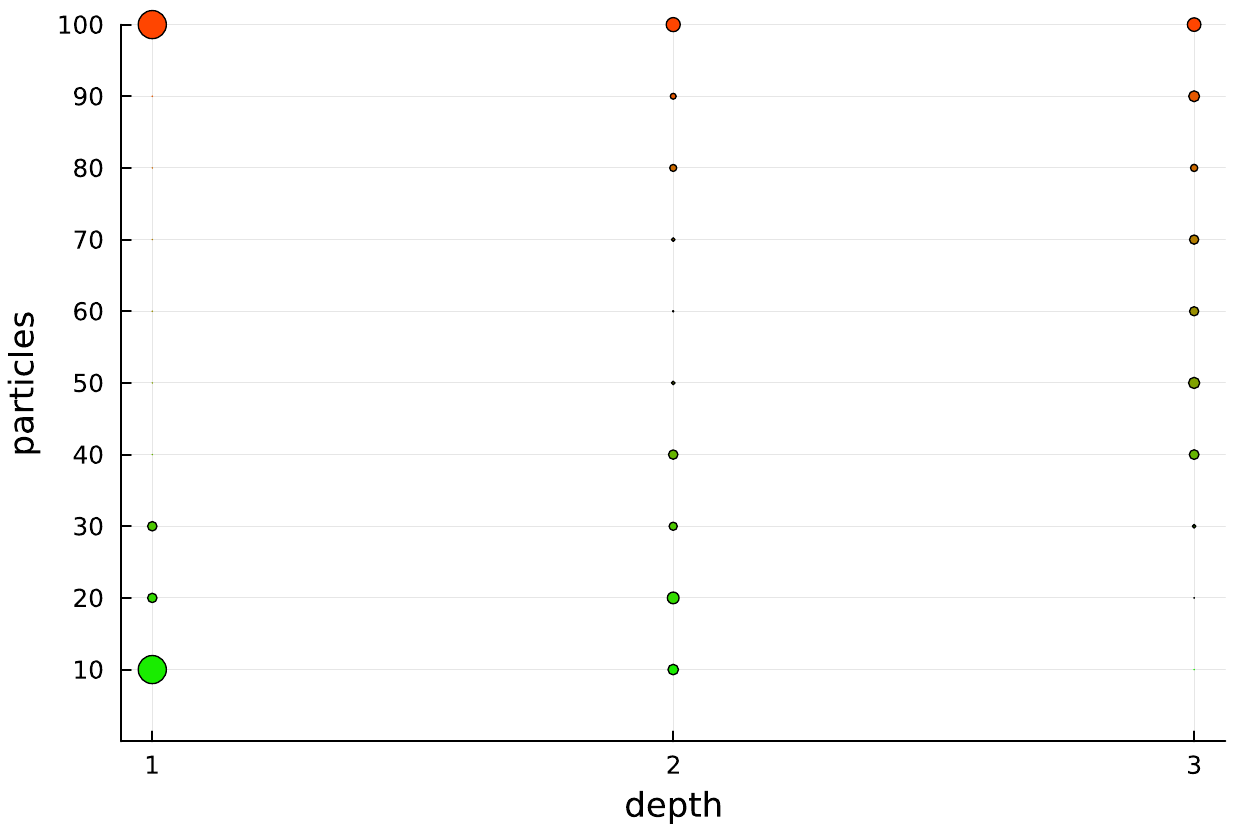}
		\subcaption{}
		\label{fig:SimplifcationLevelsSITH}	
	\end{minipage} 
	\caption{ Simplification levels at each depth of the given belief tree of Light Dark Problem (Section \ref{sec:LD}) after determining best action for one of the planning sessions. Here we present planning session $6$ of the first trial  of configuration $\lambda=0.5$ of Table~\ref{tbl:AblationStudyL3StaticGoalLambda}.  The radius of circles represents the fraction of all nodes at a particular depth that have a particular simplification level. This figure is associated with Table~\ref{tbl:SimplParticlesInTree}.  \textbf{(a)} LAZY-SITH-BSP Alg~\ref{alg:lazy-sith-bsp} \textbf{(b)} SITH-BSP Alg~\ref{alg:sith-bsp}.   }
	\label{fig:histLD}
\end{figure*}

We see that, while having larger particle speedup (column $3$),  LAZY-BSP makes more resimplification calls (column $6$) than SITH-BSP. Observing the histograms of simplification levels in Fig.~\ref{fig:histLD},  we understand that \texttt{LAZY} variant of resimplification strategy leads to lower simplification levels of the rewards at the deepest level of a given belief tree.  This was expected since the rewards at the upper levels of the belief tree participate in more laces and therefore their simplification level is promoted more times (See Alg.~\ref{alg:lazy-sith-bsp}). In addition at the lowest levels reside more beliefs and corresponding rewards. This fact is corroborated by  Table~\ref{tbl:SimplParticlesInTree}  where we witness that LAZY-BSP  yields more beliefs, in the  given in belief tree, with lower simplification levels than SITH-BSP. 
\begin{table*}[t!]
	\setlength\extrarowheight{4pt}
	\caption{This table displays the numbers of the beliefs at each  simplification level in a given tree after the identification of optimal action at the root $b_k$. Here we investigate Light Dark problem and belief tree as in Fig.~\ref{fig:histLD}. The given belief tree has $4809$ belief nodes.}
	\centering
	\resizebox{\textwidth}{!}{
		\begin{tabular}{|c|c|c|c|c|c|c|c|c|c|c|c|c|c|c|c|c|c|}
			\hline
			\multirow{2}{*}{BSP Alg.}& \multirow{2}{*}{$n_x$} &  \multirow{2}{*}{$n^1_z$} &  \multirow{2}{*}{$n^2_z$} &  
			\multirow{2}{*}{$n^3_z$} & \multirow{2}{*}{$\lambda$} &\multirow{2}{*}{$L$}  & \multicolumn{10}{|c|}{simpl. level, particles} \\
			\cline{8-17}
			&  & & & & & & 
			$\substack{ {s=1 } \\ { n^s_x=10}}$  & 
			$\substack{ {s=2 } \\ {n^s_x=20 }}$  & 
			$\substack{ {s =3 } \\ { n^s_x=30}}$  & 
			$\substack{ {s =4 } \\ { n^s_x=40}}$ & 
			$\substack{ { s =5 } \\ {n^s_x=50 }}$ & 
			$\substack{ { s =6 } \\ { n^s_x=60}}$ &
			$\substack{ {s =7  } \\ { n^s_x=70}}$& 
			$\substack{ {s =8  } \\ { n^s_x=80}}$&
			$\substack{ {s =9  } \\ { n^s_x=90}}$ & 
			$\substack{ {s =10  } \\ { n^s_x=100}}$ \\
			\hline 
			Alg~\ref{alg:lazy-sith-bsp} \ LAZY & \multirow{2}{*}{$  100 $} & \multirow{2}{*}{$1$} & \multirow{2}{*}{$3$}& \multirow{2}{*}{$3$} & 
			\multirow{2}{*}{$ 0.5 $} &  \multirow{2}{*}{$3$}&  
			$2103$ & 
			$666$ & 
			$91$ &
			$47$ &
			$14$ &
			$10$& 
			$15$&
			$88$&
			$1094$&
			$680$
			\\
			\cline{1-1} \cline{8-17}
			Alg~\ref{alg:sith-bsp} \ SITH  & &&&&&&
			$30$ & 
			$61$ & 
			$241$ &
			$618$ &
			$696$ &
			$567$& 
			$576$&
			$465$&
			$684$&
			$870$
			\\
			\hline 
		\end{tabular}
	}
	\label{tbl:SimplParticlesInTree} 
\end{table*}

\subsubsection{Results for 2D Continuous Target Tracking in the Setting of a Given Belief Tree}\label{sec:PlanningTargetTracking}
Our action space is 
$\mathcal{A} = \{  \rightarrow, \nearrow, \uparrow, \nwarrow,\leftarrow,  \swarrow,   \downarrow, \searrow,  \mathrm{Null}\}$, where action $\mathrm{Null}$ means that agent doesn't take any action. 
In this problem we selected the parameters to be $d_{\text{min}}=0.0001$, $\Sigma_T = I \cdot \sigma_T$ where $\sigma_T =0.1$ and $\sigma_{\mathcal{O}} =0.1$, $\gamma=0.95$. 

\begin{figure*}[t!]
	\begin{minipage}[t]{0.49\textwidth}   
		\centering
		\includegraphics[width=\textwidth]{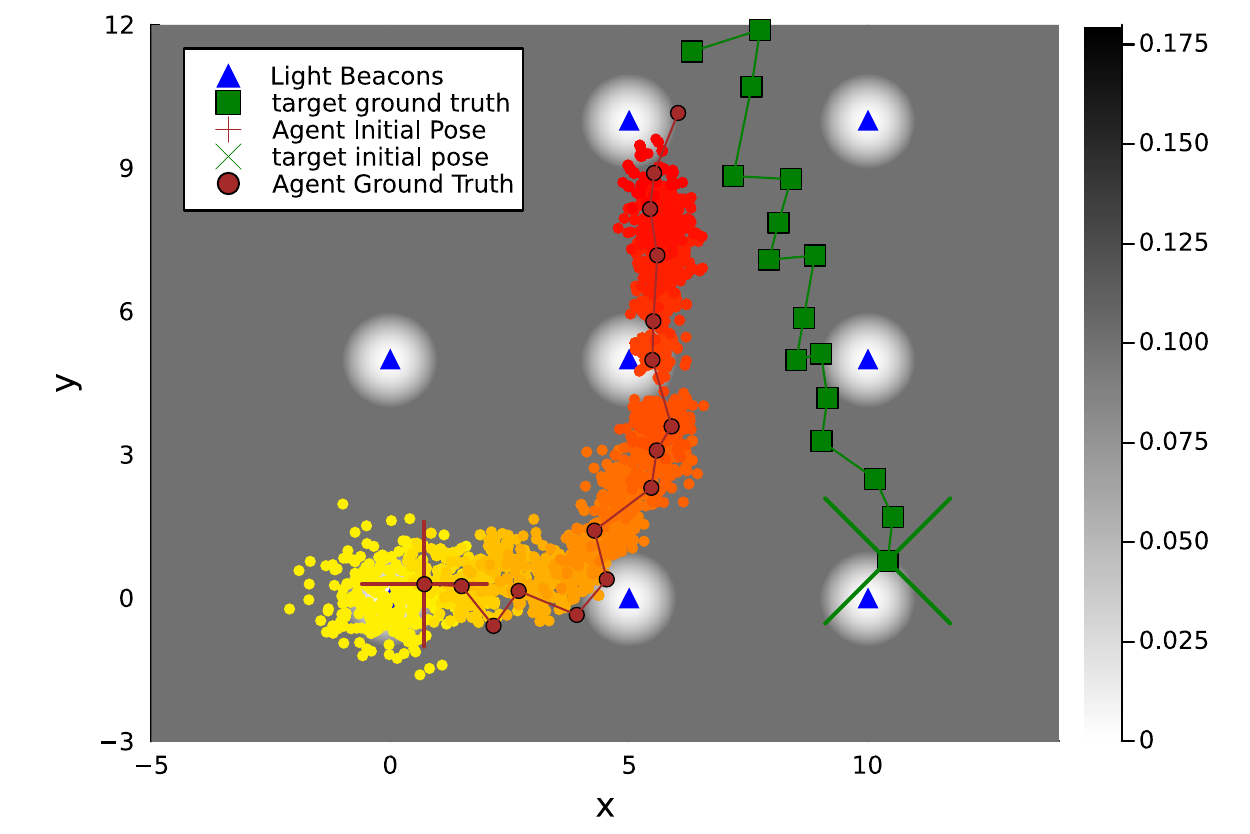}
		\subcaption{}	
		\label{fig:TargetTrackingScenarioAgent}
	\end{minipage}
	\hfill
	\begin{minipage}[t]{0.49\textwidth}
		\centering
		\includegraphics[width=\textwidth]{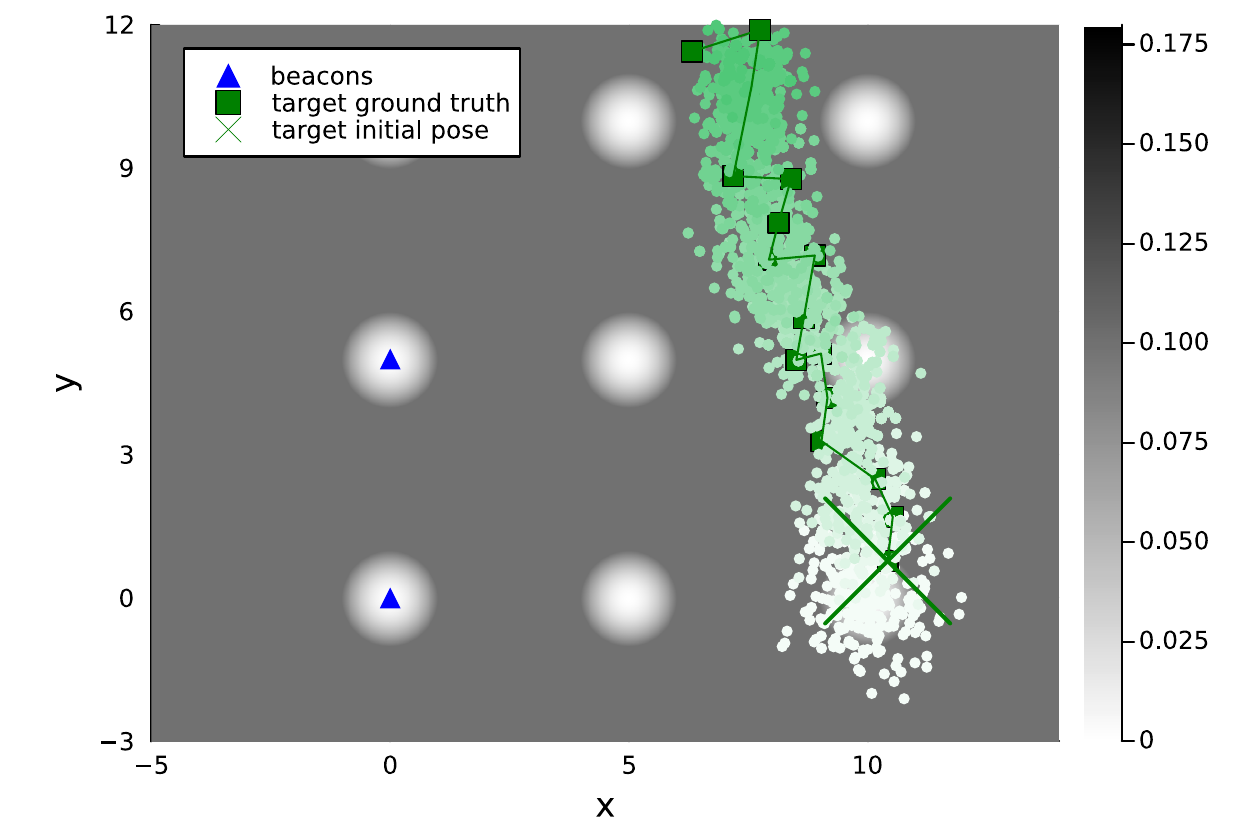}
		\subcaption{}
		\label{fig:TargetTrackingScenarioTarget}	
	\end{minipage}
	\caption{ In this illustration we show second trial of Table.~\ref{tbl:AblationStudyL3TargetTrackingParticles}, configuration $n_x=250$. The canvas color here is $\sigma_{\mathcal{O}}=\sigma_{T}=0.1$. \textbf{(a)} Agent particles \textbf{(b)} Target Particles. }
	\label{fig:PlanningTT}
\end{figure*}

We simulate $15$ trials of $15$ consecutive alternating planning sessions and the executions by the robot of the selected optimal action. Fig.~\ref{fig:PlanningTT} shows an exemplary trial. 
We show the agent particles in Fig.~\ref{fig:TargetTrackingScenarioAgent}  and the target particles in Fig.~\ref{fig:TargetTrackingScenarioTarget}.  Similar to the previous section, we study speedup with growing $\lambda$  in Table~\ref{tbl:AblationStudyL3TargetTrackingLambda} and as function of various amounts of belief particles in  Table~\ref{tbl:AblationStudyL3TargetTrackingParticles}. Again we observe that speedup diminishes with growing $\lambda$; LAZY-BSP (Alg.~\ref{alg:lazy-sith-bsp}) produces a larger speedup in terms of particles (column $4$) and time (column $5$) than SITH-BSP (Alg.~\ref{alg:sith-bsp});  accesses to motion and observation models are as expected; the return is identical for three algorithms. 

\begin{figure*}[t!]
	\begin{minipage}[t]{0.49\textwidth}   
		\centering
		\includegraphics[width=\textwidth]{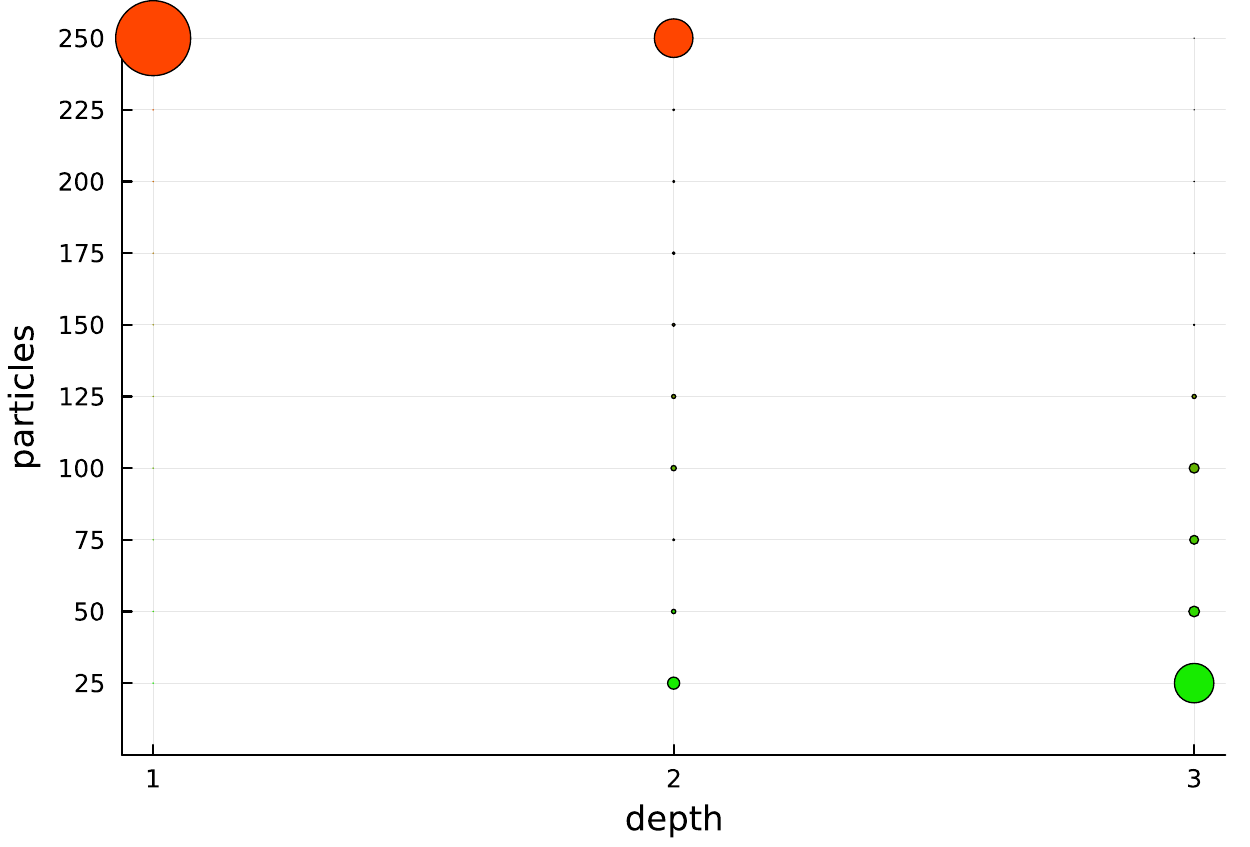}
		\subcaption{}	
		\label{fig:SimplifcationLevelsLAZYTargetTracking}
	\end{minipage}
	\hfill
	\begin{minipage}[t]{0.49\textwidth}
		\centering
		\includegraphics[width=\textwidth]{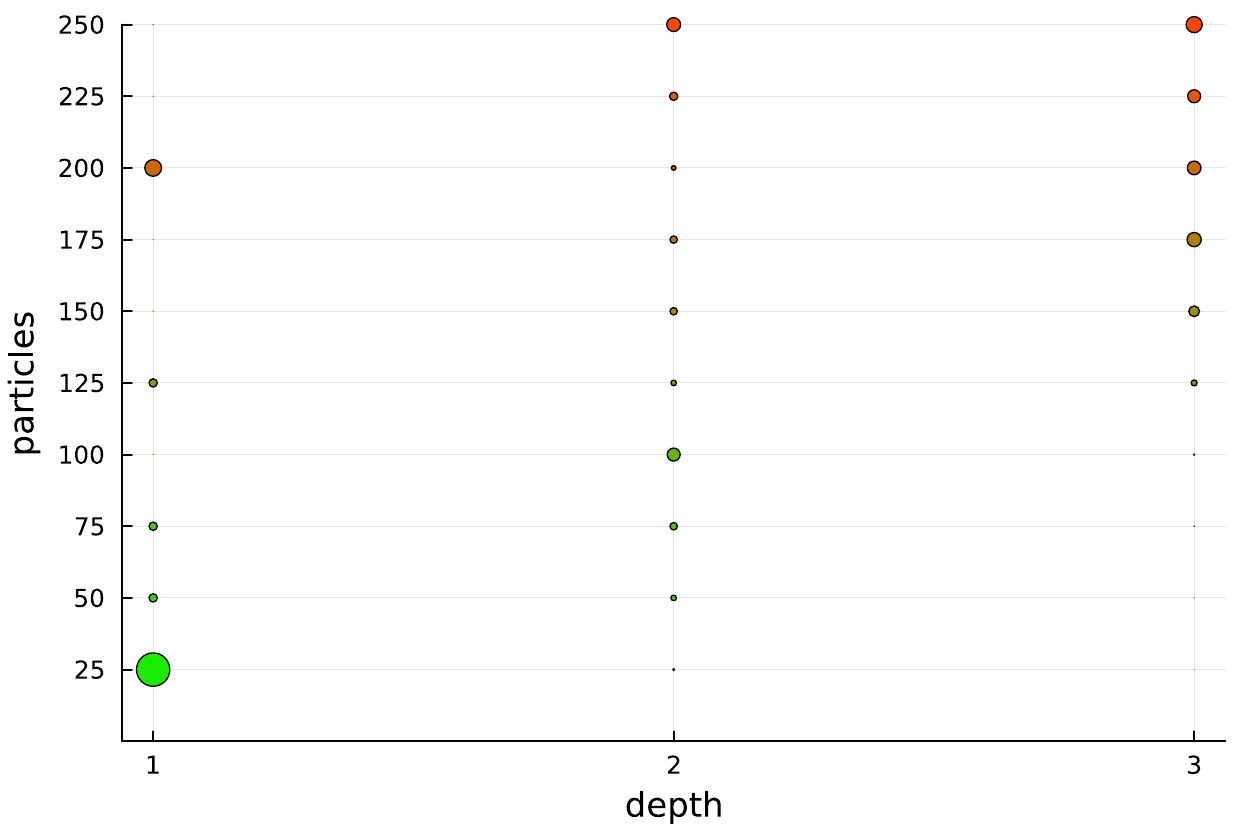}
		\subcaption{}
		\label{fig:SimplifcationLevelsSITHTargetTracking}	
	\end{minipage}
	\caption{ Simplification levels at each depth of the given belief tree of Target Tracking problem (Section~\ref{sec:TT}) after determining best action for one of the planning sessions.. Here we present planning session $6$ of the first trial  of configuration $n_x=250$ of Table~\ref{tbl:AblationStudyL3TargetTrackingParticles}.  The radius of circles represent the fraction of all nodes at particular depth that have a particular simplification level. This figure is associated with Table~\ref{tbl:SimplParticlesInTreeTT}.  \textbf{(a)} LAZY-SITH-BSP Alg~\ref{alg:lazy-sith-bsp} \textbf{(b)} SITH-BSP Alg~\ref{alg:sith-bsp}.   }
	\label{fig:histTT}
\end{figure*}

In  Fig.~\ref{fig:histTT}, which is associated with Table~\ref{tbl:SimplParticlesInTreeTT}, we observe that to select an optimal action LAZY-BSP  leaves more beliefs with lower simplification levels at the bottom of the given belief tree and produces more beliefs with lower simplification levels than SITH-BSP.  
A significant time speedup is obtained in the range of  $30\%-70 \%$  for LAZY-BSP  depending on the values of $\lambda$. For the SITH-BSP  we see less time speedup ranging from $60\%$ to $2\%$ with increasing $\lambda$. The same best action was identified by SITH-BSP, LAZY-BSP  and SS in all cases. 
Interestingly, in  configuration $n_x=350$ of Table.~\ref{tbl:AblationStudyL3TargetTrackingParticles}, for the first time we obtained that time speedup \eqref{eq:speedup} is larger than particle speedup \eqref{eq:ParticlesSpeedup}. This points to the fact that this run was so long due to large number of unsimplified belief particles $n_x =350$ so that the time of access to motion and observation models varied.

\begin{table*}[t!]
	\setlength\extrarowheight{3pt}
	\caption{This table shows cumulative results of $15$ consecutive  alternating planning and action execution sessions averaged over $15$ trials of Continuous Target Tracking problem. The given in a single planning session belief tree has $6814$ belief nodes. Overall, in $15$ planning sessions, we have $102210$ belief nodes. The horizon in each planning session  is $L=3$. The number of observations sampled from each belief action node is $n^1_z =1$, $n^2_z =3$, $n_z^3=3$ at corresponding to superscripts depths $1,2,3$. In this table we examine influence of various values of $\lambda$.}
	\centering
	\resizebox{\textwidth}{!}{
		\begin{tabular}{|c|c|c|c|c|c|c|c|c|}
			\hline
			BSP Alg.& $n_x$ &   $\lambda$ &     particles speedup \eqref{eq:ParticlesSpeedup} & time speedup \eqref{eq:speedup} & resimpl. calls (recursive) & motion model calls & obs. model calls & return ($\hat{V}$)\\ 
			\hline  
			Alg~\ref{alg:sith-bsp} \ SITH &\multirow{3}{*}{$ 100$} &  \multirow{3}{*}{$ 0.1$} &   
			$ 77.43 \pm  0.26$ & $ 60.3 \pm 2.21 $ & 
			$ 1.69 \cdot 10^5 \pm  0.04 \cdot 10^5  $& 
			$ 3.48 \cdot 10^8 \pm  0.03 \cdot 10^8 $&
			$ 10.22 \cdot 10^6 \pm  0.0 $&
			$ -79.87 \pm  9.69 $  
			\\
			\cline{1-1} \cline{4-9}
			Alg~\ref{alg:lazy-sith-bsp} \ LAZY &&&
			$ 86.97 \pm  1.28$ & $ 71.18 \pm 2.42 $ & 
			$ 7.44 \cdot 10^5 \pm  3.09 \cdot 10^5 $ & 
			$ 2.32 \cdot 10^8 \pm  0.16 \cdot 10^8 $ &
			$ 10.22 \cdot 10^6 \pm  0.0 $ &
			$ -79.87 \pm  9.69 $  
			\\
			\cline{1-1} \cline{4-9}
			SS &&&&&&
			$ 10.22 \cdot 10^8 \pm  0.0 $ &
			$ 10.22 \cdot 10^6 \pm  0.0 $ &
			$ -79.87 \pm  9.69 $   \\
			\hline 
			Alg~\ref{alg:sith-bsp} \ SITH &\multirow{3}{*}{$ 100$}  & 
			\multirow{3}{*}{$ 0.2$} &  
			$ 64.64 \pm  0.57$ & $ 46.39 \pm 2.27 $ & 
			$ 2.60 \cdot  10^5\pm  0.04 \cdot 10^5   $& 
			$ 5.03 \cdot 10^8 \pm  0.07  \cdot 10^8 $&
			$ 10.22 \cdot 10^6 \pm  0.0 $&
			$ -73.38 \pm  9.8 $  
			\\
			\cline{1-1} \cline{4-9}
			Alg~\ref{alg:lazy-sith-bsp} \ LAZY &&&
			$ 83.52 \pm  1.7$ & $ 67.24 \pm 2.62 $ & 
			$ 16.52 \cdot 10^5 \pm  5.56 \cdot 10^5 $ & 
			$ 2.75 \cdot 10^8 \pm  0.22 \cdot 10^8 $ &
			$ 10.22 \cdot 10^6 \pm  0.0 $ &
			$ -73.38 \pm  9.8 $  
			\\
			\cline{1-1} \cline{4-9}
			SS &&&&&&
			$ 10.22 \cdot 10^8 \pm  0.0 $ &
			$ 10.22 \cdot 10^6 \pm  0.0 $ &
			$ -73.38 \pm  9.8 $   \\
			\hline 
			Alg~\ref{alg:sith-bsp} \ SITH &\multirow{3}{*}{$ 100$} & \multirow{3}{*}{$ 0.3$} & 
			$ 49.57 \pm  0.93$ & $ 29.25 \pm 2.39 $ & 
			$ 3.14 \cdot 10^5 \pm  0.05 \cdot 10^5  $& 
			$ 6.86 \cdot 10^8 \pm  0.10 \cdot 10^8 $&
			$ 10.44 \cdot 10^6\pm  0.0 $&
			$ -66.29 \pm  9.3 $  
			\\
			\cline{1-1} \cline{4-9} 
			Alg~\ref{alg:lazy-sith-bsp} \ LAZY &&&
			$ 79.83 \pm  2.55$ & $ 63.34 \pm 3.45 $ & 
			$ 26.61 \cdot 10^5 \pm  8.41 \cdot 10^5 $ & 
			$ 3.21 \cdot 10^8 \pm  0.30 \cdot 10^8 $ &
			$ 10.44 \cdot 10^6\pm  0.0 $ &
			$ -66.29 \pm  9.3 $  
			\\
			\cline{1-1} \cline{4-9}
			SS &&&&&&
			$ 10.22 \cdot 10^8 \pm  0.0 $ &
			$ 10.44 \cdot 10^6 \pm  0.0 $ &
			$ -66.29 \pm  9.3 $   \\
			\hline 
			Alg~\ref{alg:sith-bsp} \ SITH &\multirow{3}{*}{$ 100$} & 
			\multirow{3}{*}{$ 0.4$} &  
			$ 35.75 \pm  1.09$ & $ 14.45 \pm 2.85 $ & 
			$ 3.61 \cdot 10^5\pm  0.06 \cdot 10^5  $& 
			$ 8.33\cdot 10^8 \pm  0.09\cdot 10^8 $&
			$ 10.44 \cdot 10^6 \pm  0.0 $&
			$ -59.99 \pm  8.05 $  
			\\
			\cline{1-1} \cline{4-9} 
			Alg~\ref{alg:lazy-sith-bsp} \ LAZY &&&
			$ 74.38 \pm  3.5$ & $ 55.69 \pm 4.38 $ & 
			$ 42.74 \cdot 10^5 \pm  12.16 \cdot 10^5$ & 
			$ 3.90 \cdot 10^8 \pm  0.38 \cdot 10^8 $ &
			$ 10.44 \cdot 10^6 \pm  0.0 $ &
			$ -59.99 \pm  8.05 $  
			\\
			\cline{1-1} \cline{4-9}
			SS &&&&&&
			$ 10.22 \cdot 10^8 \pm  0.0 $ &
			$ 10.44 \cdot 10^6 \pm  0.0 $ &
			$ -59.99 \pm  8.05 $   \\
			\hline 
			Alg~\ref{alg:sith-bsp} \ SITH &\multirow{3}{*}{$ 100$}  & 
			\multirow{3}{*}{$ 0.5$} &     
			$ 25.51 \pm  1.04$ & $ 6.44 \pm 2.49 $ & 
			$ 4.05 \cdot 10^5 \pm  0.06 \cdot 10^5  $& 
			$ 9.18 \cdot 10^8 \pm  0.08 \cdot 10^8 $&
			$ 10.44 \cdot 10^6 \pm  0.0 $&
			$ -53.15 \pm  7.03 $  
			\\
			\cline{1-1} \cline{4-9} 
			Alg~\ref{alg:lazy-sith-bsp} \ LAZY &&& 
			$ 67.76 \pm  3.88$ & $ 47.94 \pm 5.08 $ & 
			$ 63.18 \cdot 10^5 \pm  15.71 \cdot 10^5 $ & 
			$ 4.75 \cdot 10^8 \pm  0.44 \cdot 10^8 $ &
			$ 10.44 \cdot 10^6 \pm  0.0 $ &
			$ -53.15 \pm  7.03 $  
			\\
			\cline{1-1} \cline{4-9}
			SS &&&&&&
			$ 10.22 \cdot 10^8 \pm  0.0 $ &
			$ 10.44 \cdot 10^6 \pm  0.0 $ &
			$ -53.15 \pm  7.03 $   \\
			\hline 
			Alg~\ref{alg:sith-bsp} \ SITH &\multirow{3}{*}{$ 100$}  & 
			\multirow{3}{*}{$ 0.6$} & 
			$ 18.06 \pm  1.0$ & $ 2.63 \pm 2.32 $ & 
			$ 4.43 \cdot 10^5 \pm  0.06\cdot 10^5  $& 
			$ 9.65\cdot 10^8 \pm  0.06 \cdot 10^8 $&
			$ 10.44 \cdot 10^6 \pm  0.0 $&
			$ -46.97 \pm  7.14 $  
			\\
			\cline{1-1} \cline{4-9} 
			Alg~\ref{alg:lazy-sith-bsp} \ LAZY &&& 
			$ 59.53 \pm  3.78$ & $ 38.14 \pm 4.69 $ & 
			$ 89.27\cdot 10^5\pm  15.03 \cdot 10^5 $ & 
			$ 5.77 \cdot 10^8 \pm  0.43 \cdot 10^8 $ &
			$ 10.44 \cdot 10^6 \pm  0.0 $ &
			$ -46.97 \pm  7.14 $  
			\\
			\cline{1-1} \cline{4-9}
			SS &&&&&&
			$  10.22 \cdot 10^8 \pm  0.0 $ &
			$ 10.44 \cdot 10^6\pm  0.0 $ &
			$ -46.97 \pm  7.14 $   \\
			\hline 
		\end{tabular}
	}
	\label{tbl:AblationStudyL3TargetTrackingLambda} 
\end{table*}
\begin{table*}[t!]
	\setlength\extrarowheight{3pt}
	\caption{This table shows cumulative results of $15$ consecutive alternating planning and execution sessions averaged over $15$ trials of Continuous Target Tracking problem. The given belief tree has $6814$ belief nodes. Overall, in $15$ planning sessions, we have $102210$ belief nodes. The horizon in each planning session  is $L=3$. The number of observations sampled from each belief action node is $n^1_z =1$, $n^2_z =3$, $n_z^3=3$ at corresponding to superscripts depths $1,2,3$. In this table we examine  various numbers of belief particles.}
	\centering
	\resizebox{\textwidth}{!}{
		\begin{tabular}{|c|c|c|c|c|c|c|c|c|}
			\hline
			BSP Alg.& $n_x$ &  $\lambda$ &  particles speedup \eqref{eq:ParticlesSpeedup} & time speedup \eqref{eq:speedup} & resimpl. calls (recursive) & motion model calls & obs. model calls & return ($\hat{V}$)\\ 
			\hline  
			Alg~\ref{alg:sith-bsp} \ SITH &\multirow{3}{*}{$ 150$} &  \multirow{3}{*}{$ 0.5$} &  
			$ 25.19 \pm  0.94$ & $ 8.72 \pm 2.4 $ & 
			$ 4.03 \cdot 10^5 \pm  0.03 \cdot 10^5  $& 
			$ 20.71 \cdot 10^8 \pm  0.15 \cdot 10^8 $&
			$ 15.33 \cdot 10^6 \pm  0.0 $&
			$ -54.0 \pm  8.16 $  
			\\
			\cline{1-1} \cline{4-9} 
			Alg~\ref{alg:lazy-sith-bsp} \ LAZY &&&
			$ 68.36 \pm  2.66$ & $ 50.23 \pm 3.2 $ & 
			$ 63.14 \cdot 10^5\pm  9.15 \cdot 10^5 $ & 
			$ 10.55 \cdot 10^8 \pm  0.65 \cdot 10^8 $ &
			$ 15.33 \cdot 10^6 \pm  0.0 $ &
			$ -54.0 \pm  8.16 $  
			\\
			\cline{1-1} \cline{4-9}
			SS &&&&&&
			$ 22.10 \cdot 10^8 \pm  0.0 $ &
			$ 15.33 \cdot 10^6 \pm  0.0 $ &
			$ -54.0 \pm  8.16 $   \\
			\hline 
			Alg~\ref{alg:sith-bsp} \ SITH &\multirow{3}{*}{$ 250$} &  \multirow{3}{*}{$ 0.5$} &    
			$ 23.87 \pm  0.98$ & $ 11.01 \pm 3.93 $ & 
			$ 4.11 \cdot 10^5 \pm  0.05 \cdot 10^5  $& 
			$ 58.10 \cdot 10^8 \pm  0.40 \cdot 10^8 $&
			$  25.55 \cdot 10^6 \pm  0.0 $&
			$ -55.57 \pm  9.59 $  
			\\
			\cline{1-1} \cline{4-9}
			Alg~\ref{alg:lazy-sith-bsp} \ LAZY &&&
			$ 66.18 \pm  3.35$ & $ 51.51 \pm 3.83 $ & 
			$ 70.02 \cdot 10^5 \pm  12.74 \cdot 10^5 $ & 
			$ 30.79 \cdot 10^8\pm  2.37 \cdot 10^8 $ &
			$  25.55 \cdot 10^6 \pm  0.0 $ &
			$ -55.57 \pm  9.59 $  
			\\
			\cline{1-1} \cline{4-9}
			SS &&&&&&
			$ 63.88 \cdot 10^8 \pm  0.0 $ &
			$ 25.55 \cdot 10^6 \pm  0.0 $ &
			$ -55.57 \pm  9.59 $   \\
			\hline 
			Alg~\ref{alg:sith-bsp} \ SITH &\multirow{3}{*}{$ 350$} & \multirow{3}{*}{$ 0.5$} &  
			$ 23.95 \pm  1.07$ & $ 40.18 \pm 10.29 $ & 
			$ 4.11 \cdot 10^5 \pm  0.03 \cdot 10^5 $& 
			$ 113.81 \cdot 10^8 \pm  0.89 \cdot 10^8 $&
			$  35.77 \cdot 10^6 \pm  0.0 $&
			$ -55.62 \pm  8.73 $  
			\\
			\cline{1-1} \cline{4-9}
			Alg~\ref{alg:lazy-sith-bsp} \ LAZY &&&
			$ 66.36 \pm  2.58$ & $ 67.17 \pm 4.86 $ & 
			$ 69.40 \cdot 10^5 \pm  10.08 \cdot 10^5 $ & 
			$ 60.19 \cdot 10^8 \pm  3.62 \cdot 10^8 $ &
			$ 35.77 \cdot 10^6 \pm  0.0 $ &
			$ -55.62 \pm  8.73 $  
			\\
			\cline{1-1} \cline{4-9}
			SS &&&&&&
			$ 125.21 \cdot 10^8 \pm  0.0 $ &
			$ 35.77 \cdot 10^6 \pm  0.0 $ &
			$ -55.62 \pm  8.73 $   \\
			\hline 
		\end{tabular}
	}
	\label{tbl:AblationStudyL3TargetTrackingParticles} 
\end{table*}
\begin{table*}[t!]
	\setlength\extrarowheight{7pt}
	\caption{This table displays the numbers of the beliefs at each  simplification level in given tree after the identification of optimal action at the root $b_k$. Here we investigate Target Tracking problem and belief tree as in Fig.~\ref{fig:histTT}. The size of given belief tree is $6814$ belief nodes.}
	\centering
	\resizebox{\textwidth}{!}{
		\begin{tabular}{|c|c|c|c|c|c|c|c|c|c|c|c|c|c|c|c|c|}
			\hline
			\multirow{2}{*}{BSP Alg.}& \multirow{2}{*}{$n_x$} &  \multirow{2}{*}{$n^1_z$} &  \multirow{2}{*}{$n^2_z$} &  
			\multirow{2}{*}{$n^3_z$} & \multirow{2}{*}{$\lambda$} &\multirow{2}{*}{$L$} &  \multicolumn{10}{|c|}{simpl. level, particles} \\
			\cline{8-17}
			&  & & & & & & 
			$\substack{ {s=1 } \\ { n^s_x=25}}$  & 
			$\substack{ {s=2 } \\ {n^s_x=50 }}$  & 
			$\substack{ {s =3 } \\ { n^s_x=75}}$  & 
			$\substack{ {s =4 } \\ { n^s_x=100}}$ & 
			$\substack{ { s =5 } \\ {n^s_x=125 }}$ & 
			$\substack{ { s =6 } \\ { n^s_x=150}}$ &
			$\substack{ {s =7  } \\ { n^s_x=175}}$& 
			$\substack{ {s =8  } \\ { n^s_x=200}}$&
			$\substack{ {s =9  } \\ { n^s_x=225}}$ & 
			$\substack{ {s =10  } \\ { n^s_x=250}}$ \\
			\hline 
			Alg~\ref{alg:lazy-sith-bsp} \ LAZY & \multirow{2}{*}{$  250 $} & \multirow{2}{*}{$1$} & \multirow{2}{*}{$3$}& \multirow{2}{*}{$3$} & 
			\multirow{2}{*}{$ 0.5 $} &  \multirow{2}{*}{$3$}& 
			$3487$ & 
			$949$ & 
			$776$ &
			$884$ &
			$379$ &
			$106$& 
			$48$&
			$34$&
			$11$&
			$139$
			\\
			\cline{1-1} \cline{8-17}
			Alg~\ref{alg:sith-bsp} \ SITH  & &&&&&& 
			$10$ & 
			$19$ & 
			$46$ &
			$144$ &
			$538$ &
			$966$& 
			$1266$&
			$1208$&
			$1164$&
			$1452$
			\\
			\hline 
	\end{tabular}}
	\label{tbl:SimplParticlesInTreeTT} 
\end{table*}
\subsubsection{Experiments with MCTS}
\label{sec:exp}
\begin{figure*}[t!]
	\centering
	\begin{subfigure}{0.3\textwidth}            
		\includegraphics[width=\columnwidth]{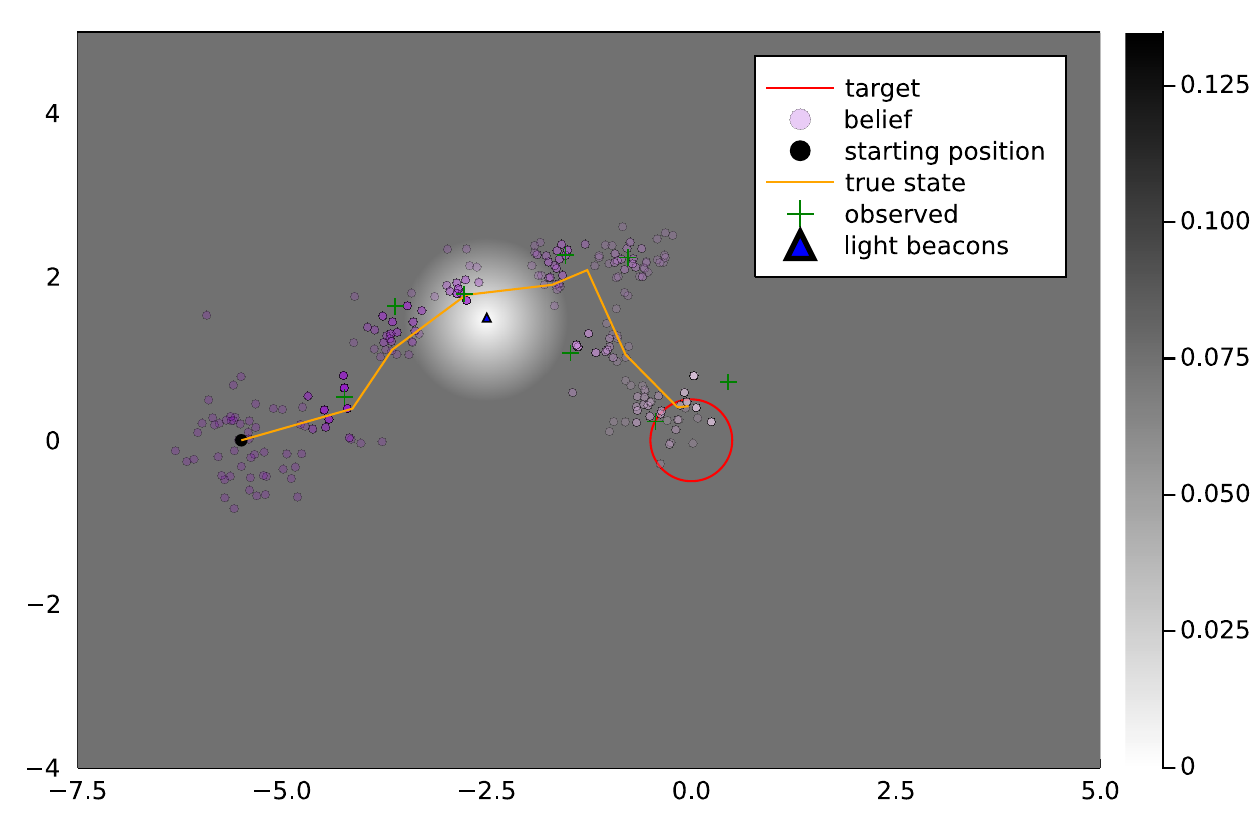}
		\caption{SITH-PFT}
		\label{fig:sith-pft_cl2d}
	\end{subfigure}%
	\begin{subfigure}{0.3\textwidth}
		\centering
		\includegraphics[width=\textwidth]{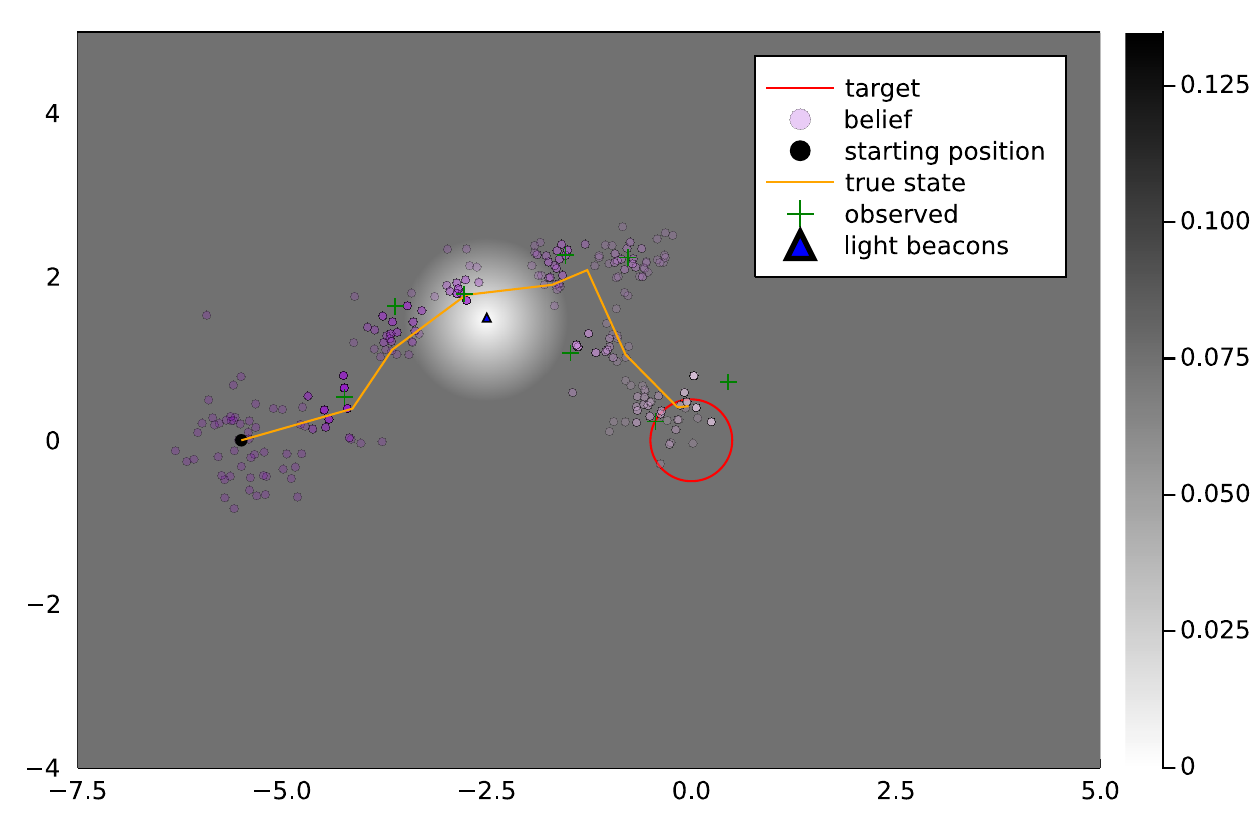}
		\caption{PFT-DPW}
		\label{fig:pft_cl2d}
	\end{subfigure}
	\begin{subfigure}{0.3\textwidth}
		\centering
		\includegraphics[width=\columnwidth]{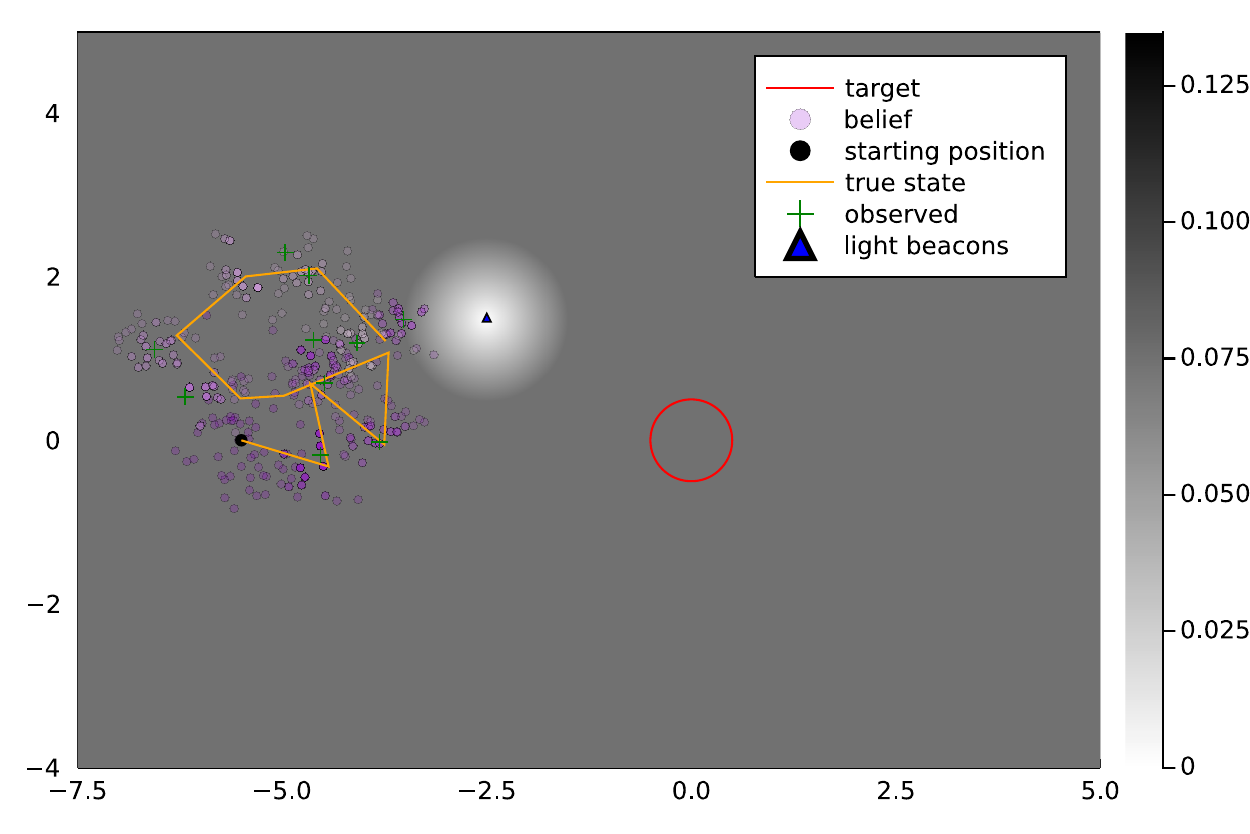}
		\caption{IPFT}
		\label{fig:ipft_cl2d}
	\end{subfigure}
	\caption{2D Continuous Light Dark. The agent starts from an initial unknown location and is given an initial belief. The goal is to get to location $(0,0)$ (circled in red) and execute the terminal action. Near the beacon (white light) the observations are less noisy. We consider multi-objective function, accounting for the distance to the goal and the differential entropy approximation (with the minus sign for reward notation). Executing the terminal action inside the red circle gives the agent a large positive reward but executing it outside it, will yield a large negative reward.}
	\label{fig:cl2d}
\end{figure*}
In an anytime setting of MCTS we focus on the $2D$-continuous light dark problem from Section \ref{sec:LD}. 
We place a single ``light beacon'' in the continuous world. Here we changed the reward. The agent's goal is to get to location $(0,0)$ and execute the terminal action - $\mathrm{Null}$. Executing it within a radius of $0.5$ from $(0,0)$ will give the agent a reward of $200$, and executing it outside the radius will yield a negative reward of $-200$. For all other actions the multi-objective reward function is $\rho(b,a,z,b') = -\underset{x\sim b'}{\mathbb{E}}[\|x \|_2] -  \hat{\mathcal{H}}(b,a,z,b')$.   The agent can move in eight evenly spread directions $\mathcal{A} = \{  \rightarrow, \nearrow, \uparrow, \nwarrow,\leftarrow,  \swarrow,   \downarrow, \searrow,  \mathrm{Null}\}$.  Motion and observation models, and the initial belief are $\probd_T(\cdot|x_,a) = \mathcal{N}(x + a, \Sigma_T)$, $\probd_Z(z|x)=\mathcal{N}(x, \min\{1, \parallel x-x^b\parallel^2_2\} \cdot \Sigma_\mathcal{O})$, $b_0 = \mathcal{N}(x_0, \Sigma_0)$ respectively. $x^b$ is the 2D location of the beacon and all covariance matrices are diagonal (i.e. $\Sigma = I\cdot \sigma^2$).  
\begin{table}[t!]
		\caption{Time speedup \eqref{eq:speedup} obtained SITH-PFT versus PFT-DPW. The rows are different configurations of the number of belief particles $n_x$, maximal tree depth $L$, and the number of iterations per planning session.  In all simulations SITH-PFT and PFT-DPW declared \emph{identical} actions as optimal and exhibited \emph{identical} belief trees in terms of connectivity and visitation counts.}
		\label{tbl:run_times}
		\centering
		\begin{tabular}{ |c| c |c| c|} 
			
			\toprule 
			
			($n_x$, $L$, \verb|#|iter.) & mean $\pm $std  & max. & min.   \\ 
			
			\midrule 
			(50, 30, 200) & $ 19.35 \pm 6.34$ & $ 30.17$ & $7.99$ \\ 
			(50, 50, 500) & $ 17.43 \pm 5.4$ & $ 33.49$ & $10.72$ \\ 
			(100, 30, 200) & $ 21.97 \pm 8.74$ & $ 49.24$ & $7.36$ \\ 
			(100, 50, 500) & $ 22.54 \pm 6.33$ & $ 36.09$ & $13.65$ \\ 
			(200, 30, 200) & $ 26.27 \pm 9.36$ & $ 42.43$ & $11.17$ \\ 
			(200, 50, 500) & $ 26.17 \pm 7.64$ & $ 44.31$ & $14.43$ \\ 
			(400, 30, 200) & $ 21.88 \pm 8.47$ & $ 37.04$ & $10.34$ \\ 
			(400, 50, 500) & $ 21.71 \pm 6.01$ & $ 32.69$ & $9.67$ \\ 
			(600, 30, 200) & $ 20.27 \pm 7.38$ & $ 32.95$ & $8.77$ \\ 
			(600, 50, 500) & $ 19.93 \pm 6.48$ & $ 31.26$ & $6.49$ \\
			\midrule 
		\end{tabular}
\end{table}%
\begin{table}[t!]
		\centering
		\caption{Total runtime of $25$ repetitions of two algorithms.}
		\label{tbl:total_run_times}
		\begin{tabular}{| c |c |c |} 
			
			\toprule 
			
			($n_x$, $L$, \verb|#|iter.) & Algorithm  & tot. plan. time [sec]   \\ 
			
			\midrule 
			\multirow{2}{*}{ (50, 30, 200)} & PFT-DPW & $ 49.7  $ \\ 
			& SITH-PFT &  $ 40.25  $ \\ 
			\hline 
			\multirow{2}{*}{ (50, 50, 500)} & PFT-DPW & $ 125.05  $ \\ 
			& SITH-PFT &  $ 103.71  $ \\ 
			\hline 
			\multirow{2}{*}{ (100, 30, 200)} & PFT-DPW & $ 185.47  $ \\ 
			& SITH-PFT &  $ 145.08  $ \\ 
			\hline 
			\multirow{2}{*}{ (100, 50, 500)} & PFT-DPW & $ 460.29  $ \\ 
			& SITH-PFT &  $ 357.57  $ \\ 
			\hline 
			\multirow{2}{*}{ (200, 30, 200)} & PFT-DPW & $ 709.66  $ \\ 
			& SITH-PFT &  $ 526.18  $ \\ 
			\hline 
			\multirow{2}{*}{ (200, 50, 500)} & PFT-DPW & $ 1755.08  $ \\ 
			& SITH-PFT &  $ 1298.86  $ \\ 
			\hline 
			\multirow{2}{*}{ (400, 30, 200)} & PFT-DPW & $ 2672.56  $ \\ 
			& SITH-PFT &  $ 2099.0  $ \\ 
			\hline 
			\multirow{2}{*}{ (400, 50, 500)} & PFT-DPW & $ 6877.24  $ \\ 
			& SITH-PFT &  $ 5403.91  $ \\ 
			\hline 
			\multirow{2}{*}{ (600, 30, 200)} & PFT-DPW & $ 6335.09  $ \\ 
			& SITH-PFT &  $ 5056.96  $ \\ 
			\hline 
			\multirow{2}{*}{ (600, 50, 500)} & PFT-DPW & $ 15682.47  $ \\ 
			& SITH-PFT &  $ 12602.09  $ \\ 
			\hline 
		\end{tabular} 
\end{table}
We selected the following parameters $x_0= \begin{pmatrix}
	-5.5 \\ 0.0
\end{pmatrix}, \Sigma_0 = \begin{pmatrix}
 0.2 & 0.0 \\ 0.0 & 0.2 
\end{pmatrix}, \sigma_T= \sigma_{\mathcal{O}} = 0.075$.
We experiment with $10$ different configurations (rows of Table~\ref{tbl:run_times}) that differ in $n_x$  (number of particles), $L$ (MCTS simulation depth), and $\#$iter (number of MCTS simulation iterations per planning session). Each scenario comprises $10$ planning sessions, i.e.~the agent performs up to $10$ planning action-executing iterations. The scenario stops if the best action determined in planning is $\mathrm{Null}$.  We repeat each experiment 25 times. In each such repetition we run PFT-DPW and SITH-PFT with the same seed and calculate the relative time speedup in percentage according to \eqref{eq:speedup} where $t_{\mathrm{PFT-DPW}}$ and $t_{\mathrm{SITH-PFT}}$ are running times of a baseline and our methods respectively.

In all different configurations, we obtained significant time speedup of approximately $20\%$ while achieving the exact same solution compared to PFT. In Table~\ref{tbl:run_times} we report the  mean and standard error  of \eqref{eq:speedup} as well as maximum and minimum value. Remarkably, we observe that we never slowdown the PFT-DPW with SITH-PFT. We also present total running times of $25$ repetitions of at most $10$ (the simulation stops if best identified action is $\mathrm{Null}$) planning sessions in Table~\ref{tbl:total_run_times}. Note that we divided the total planning time by the number of planning sessions in each repetition.

An illustration of evaluated scenario can be found in Fig.~\ref{fig:cl2d}. Note that SITH-PFT (Fig.~\ref{fig:sith-pft_cl2d}) yields an identical to PFT solution (Fig.~\ref{fig:pft_cl2d}) while IPFT demonstrates a severely degraded behavior. We remind the purpose of our work is to speedup the PFT approach when coupled with information-theoretic reward. Since the two algorithms produce identical belief trees and action at the end of each planning session, there is no point reporting the algorithms \emph{identical} performances (apart from planning time).

\subsubsection{Localization with Collision Avoidance Solved by MCTS}
\label{sec:SafeLoc}
In this section, we investigate the application of three algorithms, IPFT \citep{Fischer20icml}, PFT-DPW \citep{Sunberg18icaps}  and our SITH-PFT encapsulated by Alg.~\ref{alg:sith-pft}. The algorithmic implementation of IPFT boils down to making more simulations inside IPFT with substantially less number of belief particles subsampled from root belief. 

Further, we discuss the quality and speed of IPFT. Representation of the belief with a tiny amount of particles induces larger error in differential entropy estimation and other parts of the reward function such as, for example, soft safety reward component in \eqref{eq:SafeLocalizationReward}. The authors of \citep{Fischer20icml} claim that IPFT averages differential entropies calculated from tiny subsets subsampled from the particle belief. However, observing the \texttt{SIMULATE} routine (similar to our in Alg.~\ref{alg:sith-pft}) in \citep{Fischer20icml}, we see that in practice this average is obtained through more simulations, starting from a new subsample from the root belief, with less number of particles, thereby averaging entropies calculated from different beliefs with less number of particles, but same history of actions and observations. The parameter $K$ in \citep{Fischer20icml} in practice is the visitation count $N(b)$ of each belief in the belief tree. There is no direct control of this parameter. In other words, to make a proper comparison we shall increase the number of \texttt{SIMULATE} calls inside IPFT by a factor $K=\nicefrac{n_x}{m}$ where $m$ is the size of the subsample from a belief represented by $n_x$ particles. In such a way in both belief trees, built by IPFT and PFT-DPW,  there are the same number of total particles. 
This is in contrast to using the same number of calls to \texttt{SIMULATE} in both trees. If the number of calls to \texttt{SIMULATE} is the same the number of particles in the tree built by IPFT will be much smaller than in the tree built by PFT-DPW. 
Do note that we cannot assure that the  same $K$ will be for each future history due to the exploratory nature of MCTS.

The speed of IPFT is linked with the rollout policy of MCTS. As we mentioned above, when the belief is represented by particles we know that asymptotically when the number of particles tends to infinity this representation converges to the theoretical belief for any given belief \citep{Crisan02tsp}. Therefore, we shall take as many particles as possible for the belief representation. Given that the size of subsample $m$ in IPFT does not change, this will increase the parameter $K$ and therefore slowdown IPFT. Because when the new belief node is expanded in the belief tree there is always a rollout initiated, a more complex rollout policy will slowdown IPFT more, yet, this is ultimately the question of how big the parameter $K$ is.

%
\begin{figure*}[t!]
	\begin{minipage}[t]{0.33\textwidth}   
		\centering
		\includegraphics[width=\textwidth]{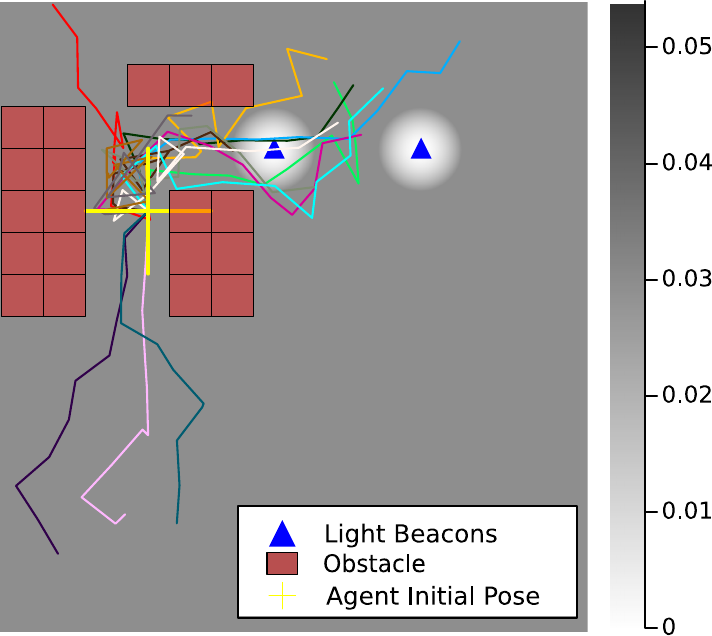}
		\subcaption{}	
		\label{fig:SafeIPFT}
	\end{minipage}
	\hfill
		\begin{minipage}[t]{0.33\textwidth}
		\centering
		\includegraphics[width=\textwidth]{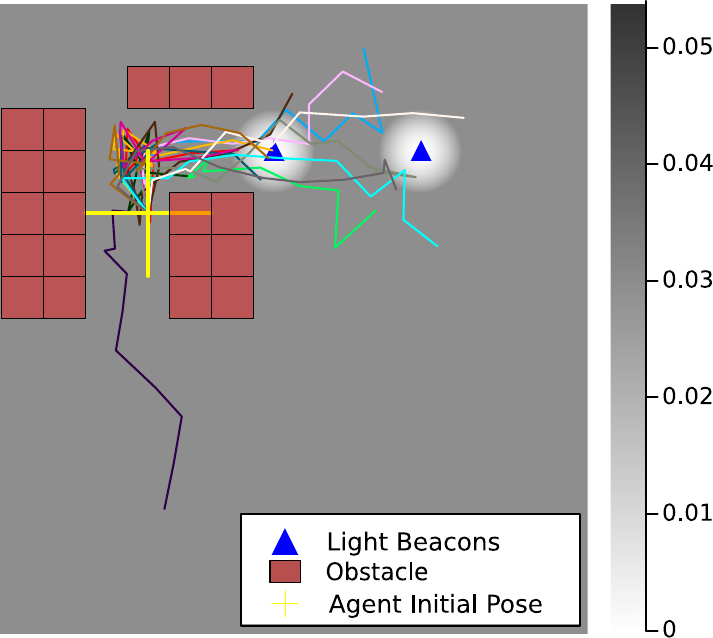}
		\subcaption{}
		\label{fig:SafePFT}		
	\end{minipage}
	\hfill
	\begin{minipage}[t]{0.33\textwidth}
		\centering
		\includegraphics[width=\textwidth]{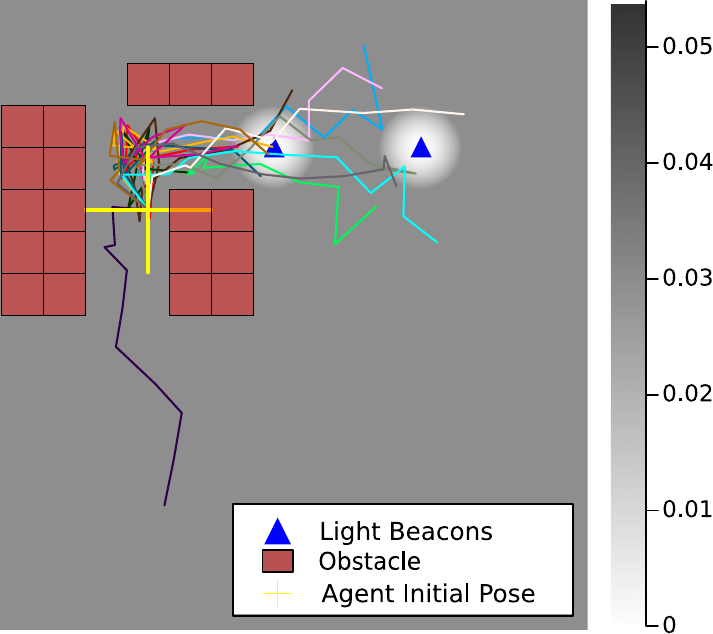}
		\subcaption{}
		\label{fig:SafeSITH}	
	\end{minipage}
	\caption{The plot shows  $15$ differently colored robot trajectories. Each such  trajectory comprises ten time steps. In each such step the robot performs re-planning and executes the best action selected by an appropriate BSP algorithm. The color of each trajectory matches  planning with the same seed in each plot. The canvas color here is $\sigma_{\mathcal{O}}=\sigma_{T}=0.03$ as in equations \eqref{eq:LDMotionModel} and \eqref{eq:LDObsModel} respectively. 
	The parameters are $n_x=300$, $m=20$, number of calls to \texttt{SIMULATE} of IPFT is $4500$, the number of calls to \texttt{SIMULATE} of PFT-DPW and SITH-PFT is $300$. In such a setting the constructed belief trees by these methods have the same number of total samples (see Section~\ref{sec:SafeLoc} for details).  \textbf{(a)} Safe IPFT. \textbf{(b)} Safe SITH-PFT (Alg~\ref{alg:sith-pft}), \textbf{(c)} Safe PFT-DPW.}
	\label{fig:IPFT_SITHTraj}
\end{figure*}
\begin{figure*}[t!]
	\begin{minipage}[t]{0.49\textwidth}   
		\centering
		\includegraphics[width=\textwidth]{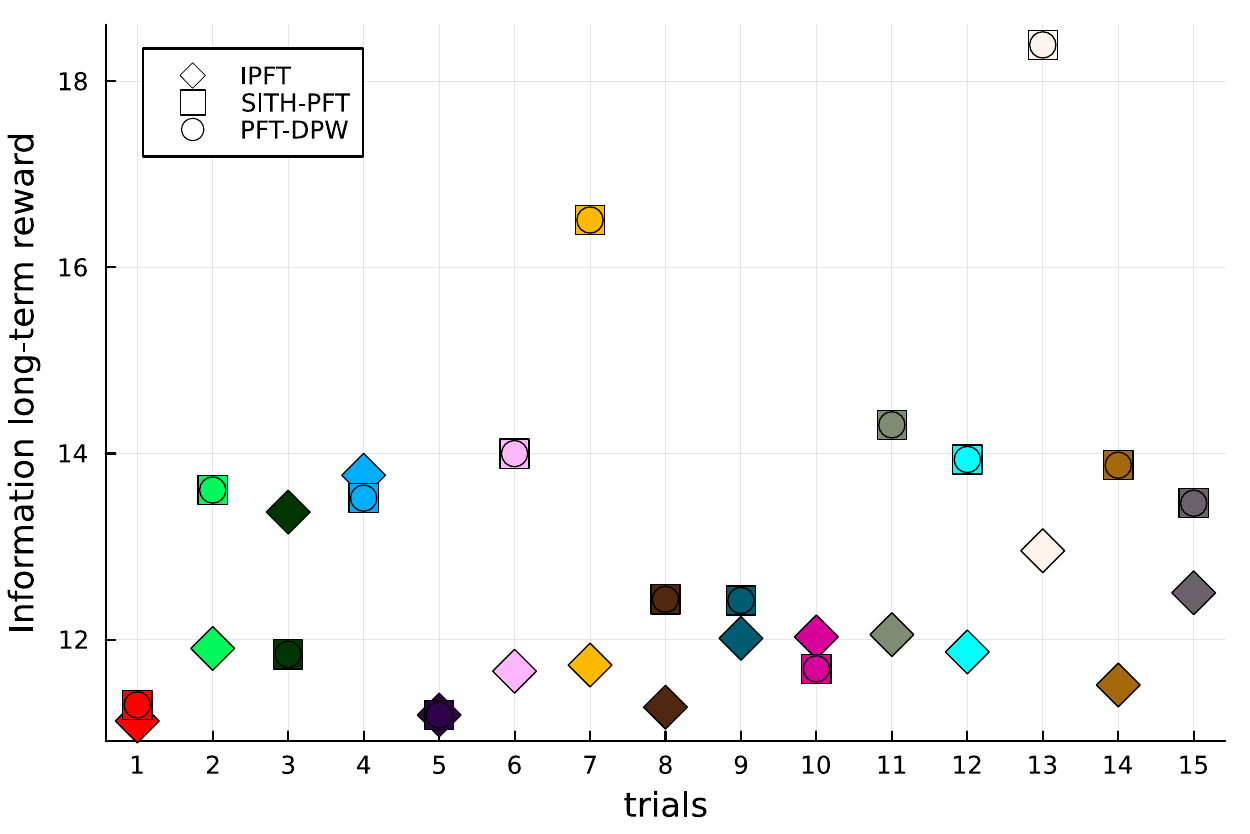}
		\subcaption{}	
		\label{fig:InformationSafeLocalization}
	\end{minipage}
	\hfill
	\begin{minipage}[t]{0.49\textwidth}
		\centering
		\includegraphics[width=\textwidth]{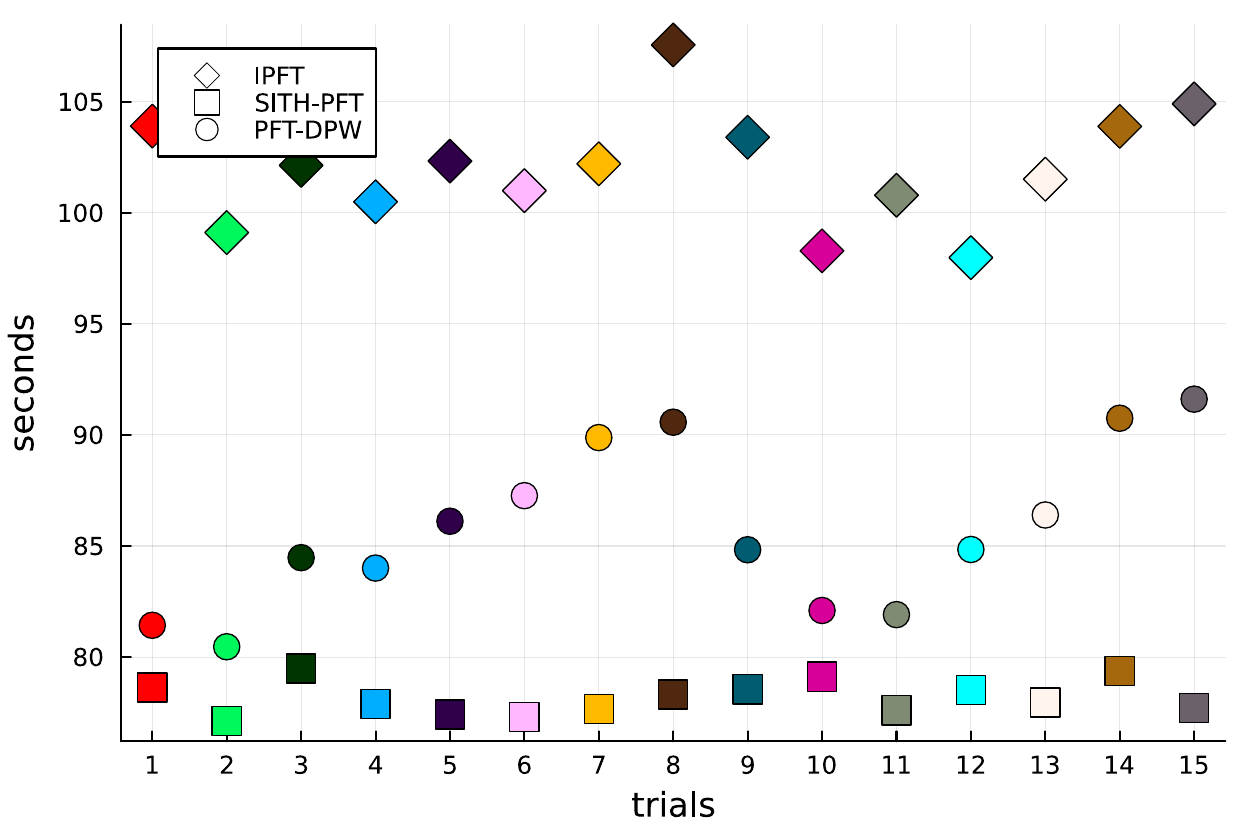}
		\subcaption{}
		\label{fig:RunningTimesSafeLocalization}	
	\end{minipage}
	\caption{This plot is associated with Fig.~\ref{fig:IPFT_SITHTraj}. Each color matches the corresponding trajectory in Fig.~\ref{fig:IPFT_SITHTraj}. The parameters are $n_x=300$, $m=20$, $K = \frac{300}{20}=15$ number of calls to \texttt{SIMULATE} of IPFT is $4500$, the number of calls to \texttt{SIMULATE} of PFT-DPW and SITH-PFT is $300$.  In such a setting the constructed belief trees by these methods have the same number of total samples (see Section~\ref{sec:SafeLoc} for details).  \textbf{(a)} Cumulative information reward as in \eqref{eq:SafeLocalizationReward} in the execution of the trajectory. Here the SITH-PFT curve 
	and the PFT-DPW curve overlap. This is because the rewards are identical since the same best action is calculated by SITH-PFT and PFT-DPW; \textbf{(b)} Average planning times of $10$ planning sessions in each trial.}
\end{figure*}

As we observe in Fig.~\ref{fig:IPFT_SITHTraj}, IPFT is less accurate compared to PFT-DPW and SITH-PFT  in spite of a much larger number of calls to \texttt{SIMULATE} routine compared to PFT-DPW and SITH-PFT. Clearly, better localization is closer to the beacons. In Fig.~\ref{fig:SafeIPFT} we see that more trajectories went to completely different from beacons directions as opposed to  Fig.~\ref{fig:SafeSITH} and Fig.~\ref{fig:SafePFT} displaying identical results.  From Fig.~\ref{fig:InformationSafeLocalization} we conclude that in $10$ from $15$ trials the information reward obtained in execution of the optimal action returned by IPFT was inferior to the corresponding reward obtained by SITH-PFT and PFT-DPW. From Fig.~\ref{fig:RunningTimesSafeLocalization} we see that IPFT is slowest from the three algorithms while SITH-PFT (Alg.~\ref{alg:sith-pft}) is the fastest \emph{in all trials}. 
\subsection{Discussion}
Although the speedup was significant and steady for all simulations, we did not observe growth in speed-up with growth of number of belief particles in any simulation. This can be explained by the fact that increasing number of particles of the belief ($n_x$) changes also the bounds because the parameter $n_x$ is present in the bounds as well.      
The limitation of our approach is that it leans on converging bounds, which are not trivial to derive and specific for a particular reward function. In addition, it requires slightly more caching than the baseline. Our simplification approach may still be ill-timed, since the resimplifications take an additional toll in terms of running time.

\section{Conclusions} \label{sec:concl}
We contributed a rigorous provable theory of adaptive multilevel simplification that accelerates the solution of belief-dependent fully continuous POMDP. Our theory always identifies the same optimal action or policy as the unsimplified analog. Our theoretical approach receives as input adaptive bounds over the belief-dependent reward.  Using the suggested theory and any bounds satisfying stated conditions we formulated three algorithms, considering a given belief tree and an anytime MCTS setting.  We also contributed a specific simplification for nonparametric beliefs represented by weighted particles and derived novel bounds over a differential entropy estimator.  These bounds are computationally cheaper than the latter. Our experiments demonstrate that our algorithms are paramount in terms of computation time while guaranteed to have the same performance as the baselines. In the setting of the given belief tree, we achieved a speedup up to $70\%$. In an anytime MCTS setting, our algorithm enjoyed the speedup of $20\%$.

\section{Funding}
This research was  supported by the Israel Science Foundation (ISF) and by 
a donation from the Zuckerman Fund to the Technion Artificial Intelligence Hub (Tech.AI).
\section{APPENDIX} \label{sec:App}

\subsection{Proof for Theorem \ref{thm:MonotonConvQ}}\label{proof:MonotonConvQ}
To shorten the notations we prove the theorem for value function under arbitrary policy. Note that by substituting the policy $\pi_{(\ell)+}$  by $\{\pi_{\ell}(b_{\ell}), \pi^*_{(\ell+1)+}\}$ where $a_{\ell} = \pi_{\ell}(b_{\ell})$ we always can obtain action-value function.  
Without loosing generality assume the resimplification hits an arbitrary belief action node. The new upper bound will be 
\begin{align}
	\!\!\overline{\hat{V}}(b_{\ell}, \pi_{\ell+}) \! + \! \frac{1}{M} \Bigg(\!\!	\underbrace{  \overline{\Delta}^{s+1}(b,a,b')\! - \! \overline{\Delta}^s(b,a,b')}_{\leq 0} \!\!\Bigg) \!\!\leq \! \overline{\hat{V}}(b_{\ell}, \pi_{\ell+})\!\!\!
\end{align}
The new lower bound will be
\begin{align}
		\!\underline{\hat{V}}(b_{\ell}, \pi_{\ell+})\! - \! \frac{1}{M} \Bigg(\!\!	\underbrace{  \underline{\Delta}^{s+1}(b,a,b') \!-\! \underline{\Delta}^s(b,a,b')}_{\leq 0} \!\! \Bigg) \!\! \geq \! 	\underline{\hat{V}}(b_{\ell},\! \pi_{\ell+})\!\!\!
\end{align}
where $M= n_z^{d}$ depending on the depth $d$ of resimplified reward bound. Moreover if  the inequalities involving increments are strict $   \overline{\Delta}^s(b,a,b') > \overline{\Delta}^{s+1}(b,a,b')$ and $\underline{\Delta}^s(b,a) > \underline{\Delta}^{s+1}(b,a,b')$ also the retracting the bounds over Value function inequalities are strict. In case of MCTS,  we have that $M= \frac{N(ha)}{N(h')}$ where history $ha$ corresponds to $b_{\ell}$ and action $a$, and $h'$ corresponds to $b'$. 
\qed \\
\subsection{Proof of Lemma \ref{lem:ValidStrategy}}\label{proof:ValidStrategy}
Recall that the bounds $\overline{\rho}, \underline{\rho}$ of belief nodes and ''weakest link'' rollout nodes are refined when the inequality \eqref{eq:cond} is encountered.

Assume in contradiction that the resimplification strategy does not promote any reward level and $G(ha) > 0$. This means that $\nicefrac{G(ha)}{d} > 0 $ and for all reward bounds the inequality  	$\gamma^{d-d'} \cdot (\overline{\rho}- \underline{\rho}) < \frac{1}{d} G(ha)$. This is not possible since $\nicefrac{G(ha)}{d} $ is the mean gap with respect to simulations of MCT and the depth of the belief tree, multiplied by the appropriate discount factor, over all the nodes that are the descendants to $ha$ . See equation \eqref{eq:SampleQboundsBellmanMCTS}.  \qed

\subsection{Proof of Lemma \ref{lem:UCBboundsmonotonicity}}\label{proof:UCBboundsmonotonicity}
Observe that
\begin{align}
\uucb(ha)\!- \!\lucb(ha) = \overline{\hat{Q}}(ha) - \underline{\hat{Q}}(ha).
\end{align} 
We already proved the desired for $\overline{\hat{Q}}(ha), \underline{\hat{Q}}(ha)$ in Theorem~\ref{thm:MonotonConvQ}. Using the convergence $
	\underline{\hat{Q}}(\cdot)  =  \hat{Q}(\cdot) = \overline{\hat{Q}}(\cdot)$ we obtain
	\begin{align}
	&\underline{\hat{Q}}(\cdot)+  c \! \cdot \! \sqrt{\nicefrac{\log(N(h))}{N(ha)}} =  \nonumber \\
	&\hat{Q}(\cdot) + c \! \cdot \! \sqrt{\nicefrac{\log(N(h))}{N(ha)}} =  \\
	&\overline{\hat{Q}}(\cdot)+ c \! \cdot \! \sqrt{\nicefrac{\log(N(h))}{N(ha)}} \nonumber.  
	\end{align} 
	The proof is completed.
\qed
\subsection{Proof of Theorem \ref{thm:TreeConsis}}\label{sec:TreeConsis}
We provide proof by induction on the belief tree structure.\\ 
\textbf{Base:} Consider an initial given belief node $b_0$. No actions have been taken and no observations have been made. Thus, both the PFT tree and the SITH-PFT tree contain a single identical belief node, and the claim holds. \\
\textbf{Induction hypothesis:} Assume we are given two identical trees with $n$ nodes, generated by PFT and SITH-PFT. The trees uphold the terms of \textbf{Definition \ref{def:TreeConsis}}.\\
\textbf{Induction step:} Assume in contradiction that different nodes were added to the trees in the next simulation (expanding the belief tree by one belief node by definition). Thus, we got different trees.\\
Two scenarios are possible:	
\begin{case}\label{enum:case_1}
	The same action-observation sequence $a_0, z_1, a_1, z_2... a_m$ was chosen in both trees, but different nodes were added.
\end{case}
\begin{case}\label{enum:case_2}
	Different action-observation sequences were chosen for both trees, and thus, we got different trees structure.
\end{case}
Since the Induction hypothesis holds, the last action  $a_m$ was taken from the same node denoted $h'$ shared and identical to both trees. Next, the same observation model is sampled for a new observation, and a new belief node is added with a rollout emanating from it. The new belief nodes and the rollout are identical for both trees since both algorithms use the same randomization seed and observation and motion models. Case~\ref{enum:case_2} must be true since we showed Case~\ref{enum:case_1} is false. There are two possible scenarios such that different action-observation sequences were chosen:
\begin{subcases}
	\begin{case}\label{enum:case_2.1}
		At some point in the actions-observations sequence, different observations $z_i, z'_i$ were chosen.
	\end{case}
	\begin{case}\label{enum:case_2.2}
		At some point in the actions-observations sequence, PFT chose action $a^{\dagger}$ while SITH-PFT chose a different action, $\tilde{a}$, or  got stuck without picking any action.
	\end{case}
	
	Case~\ref{enum:case_2.1} is not possible since if new observations were made, they are the same one by reasons contradicting Case~\ref{enum:case_1}. If we draw existing observations (choose some observation branch down the tree) the same observations are drawn since they are drawn with the same random seed and from the same observations ``pool''. It is the same ``pool'' since the Induction hypothesis holds. Case~\ref{enum:case_2.2} must be true since we showed Case~\ref{enum:case_2.1} is false, i.e., when both algorithms are at the identical node denoted as $h$ PFT chooses action $a^{\dagger}$, while SITH-PFT chooses a different action, $\tilde{a}$, or even got stuck without picking any action.
	Specifically, PFT chooses action $a^{\dagger} = \maxim{\argmax}{a}\ \text{UCB}$ and SITH-PFT's candidate action is $\tilde{a} = \maxim{\argmax }{a \in \mathcal{A}}  \ \lucb(ha)$. Three different scenarios are possible:
	
	\begin{subcases}
		\begin{case}\label{enum:case_2.2.1}
			the $\uucb, \lucb$ bounds over $h\tilde{a}$ were tight enough and $\tilde{a}$ was chosen such that $a^{\dagger} \neq \tilde{a}$.
		\end{case}
		\begin{case}\label{enum:case_2.2.2}
			SITH-PFT is stuck in an infinite loop. It can happen if the $\uucb, \lucb$ bounds over $h\tilde{a}$, and at least one of its sibling nodes $ha$, are not tight enough. However, all tree nodes are at the maximal simplification level. Hence, resimplification is triggered over and over without it changing anything.
		\end{case}
	\end{subcases}
\end{subcases}

Case~\ref{enum:case_2.2.1} is not possible as the bounds are analytical (always true) and converge to the actual reward ($\lucb=\text{UCB}=\uucb$) for the maximal simplification level. Case~\ref{enum:case_2.2.2} is not possible. If the bounds are not close enough to make a decision, resimplification is triggered. Each time some $ha$ node - sibling to $h\tilde{a}$ and maybe even $h\tilde{a}$ itself is chosen in \textit{SelectBest} to over-go resimplification. According to lemmas~\ref{lem:ValidStrategy} and \ref{lem:UCBboundsmonotonicity}, after some finite number of iterations for all of the sibling $ha$ nodes (including $h\tilde{a}$) it holds $\lucb(ha)=\text{UCB}(ha)=\uucb(ha)$ and some action can be picked. If different actions have identical values we choose one by the same rule UCB picks actions with identical values (e.g. lower index/random). Since Case~\ref{enum:case_2.2.2} is false, after some finite number of resimplification iterations, SITH-PFT will stop with bounds sufficient enough to make a decision; as Case~\ref{enum:case_2.2.1} is false it holds that $a^{\dagger} = \tilde{a}$. Thus we get a contradiction and the proof is complete. \qed

\subsection{Proof of Theorem \ref{thm:SolConsis}}\label{sec:SolConsis}
Since  same tree is built according to Theorem~\ref{thm:TreeConsis}, the only modification  is the final criteria at the end of the planning session at the root of the tree: $a^*=\maxim{\argmax}{a}\ Q(ha)$. Note we can set the exploration constant of UCB to $c=0$ and we get that UCB is just the $Q$ function. Thus if the bounds are not tight enough at the root to decide on an action, resimplification will be repeatedly called until SITH-PFT can make a decision. The action will be identical to the one chosen by UCB at PFT from similar arguments in the proof of Theorem~\ref{thm:TreeConsis}. Note that additional final criteria for action selection could be introduced, but it would not matter as tree consistency is kept according to Theorem~\ref{thm:TreeConsis} and the bounds converge to the  immediate rewards and $Q$ estimations. \qed

\subsection{Proof for Theorem \ref{theorem:LowerUpperReward}}\label{proof:LowerUpperReward}
Let us first prove that $u+\hat{\mathcal{H}} \geq 0$. the It holds 
\begin{align}
	&u+\hat{\mathcal{H}}  = \sum_{i \notin  A^s_{k+1}} \!\!\! w_{k+1}^i \!\cdot \! \log\left[ m \cdot \probd_Z(z_{k+1} | x_{k+1}^i)\right] + \label{eq:upperboundplusentropy}\\
	&\sum_{i \in A^s_{k+1}} \!\!\!\! w_{k+1}^i \!\cdot \! \log \!\! \left[\probd_Z(z_{k+1} | x_{k+1}^i) \!\! \sum_{j=1}^{n_x}\probd_T(x_{k+1}^i | x_k^j, a_k)w_k^j\right]\nonumber \!\! -\\
	&\sum_{i=1}^{n_x} \! w_{k+1}^i\!\cdot \!\log \!\! \left[\probd_Z(z_{k+1}| x_{k+1}^i) \!\! \sum_{j=1}^{n_x} \probd_T(x_{k+1}^i | x_k^j, a_k)w_k^j\right]\!\! =\nonumber 
\end{align}	
The Eq.~\eqref{eq:upperboundplusentropy} equals to 
\begin{align} 	
	&\sum_{i \notin  A^s_{k+1}}  \!\!\!  w_{k+1}^i \! \cdot \! \log\left[ m \cdot \probd_Z(z_{k+1} | x_{k+1}^i)\right]-\nonumber\\
	&\sum_{i\notin  A^s_{k+1}} \!\!\! w_{k+1}^i \! \cdot \! \log \! \left[\probd_Z(z_{k+1}| x_{k+1}^i) \sum_{j=1}^{n_x}\probd_T(x_{k+1}^i | x_k^j, a_k)w_k^j\right] \nonumber
\end{align}
Fix arbitrary index $i \notin  A^s_{k+1}$. The $\log$ is monotonically increasing function so it is left to prove that 
\begin{align}
	m  \probd_Z(z_{k+1} | x_{k+1}^i) \!\!\geq \! \probd_Z(z_{k+1}| x_{k+1}^i) \!\! \sum_{j=1}^{n_x}\!\probd_T(x_{k+1}^i| x_k^j, a_k)w_k^j \nonumber
\end{align}
If  $\probd_Z(z_{k+1} | x_{k+1}^i) =0$, we finished. Assume  $\probd_Z(z_{k+1} | x_{k+1}^i)  \neq 0  $.
It holds that 
\begin{align}
	&\probd_Z(z_{k+1}| x_{k+1}^i) \sum_{j=1}^{n_x} \max_{\substack{ x_{k+1} \\ x_{k}, a_k}}\probd_T(x_{k+1}^i | x_k^j, a_k)w_k^j \geq \\
	&\probd_Z(z_{k+1}| x_{k+1}^i) \sum_{j=1}^{n_x}\probd_T(x_{k+1}^i | x_k^j, a_k)w_k^j
\end{align}
We reached the desired result.
Now let us show the second part $\ell + \hat{\mathcal{H}} \leq 0$. Observe, that 
\begin{align}
	& 0 \geq  \ell + \hat{\mathcal{H}} = \label{eq:lowerboundplusentropy} \\
	&\sum_{i=1}^{n_x} w_{k+1}^i  \log \! \! \left[ \probd_Z(z_{k+1} | x_{k+1}^i)\!\!\sum_{j \in A^s_k} \probd_T(x_{k+1}^i | x_k^j, a_k)w_k^j\right]\!\! -  \nonumber\\
	&\sum_{i=1}^{n_x} w_{k+1}^i  \log \!\! \left[\probd_Z(z_{k+1}| x_{k+1}^i) \!\!\sum_{j=1}^{n_x}\probd_T(x_{k+1}^i| x_k^j, a_k)w_k^j\right] \nonumber
\end{align}
Select arbitrary index $i$.  We shall prove that 
\begin{align}
	&\log\left[ \probd_Z(z_{k+1} | x_{k+1}^i)\sum_{j \in A^s_k} \probd_T(x_{k+1}^i | x_k^j, a_k)w_k^j\right]-\nonumber\\
	&\log\left[\probd_Z(z_{k+1} | x_{k+1}^i) \sum_{j=1}^{n_x}\probd_T(x_{k+1}^i | x_k^j, a_k)w_k^j\right] \leq 0\nonumber
\end{align}
Again use that $\log$ is monotonically increasing and assume that $\probd_Z(z_{k+1} | x_{k+1}^i)  \neq 0  $. We have that 
\begin{align}
	&\sum_{j \in A^s_k} \probd_T(x_{k+1}^i | x_k^j, a_k)w_k^j-\sum_{j=1}^{n_x}\probd_T(x_{k+1}^i | x_k^j, a_k)w_k^j=\\
	& -\sum_{j \notin A^s_k} \probd_T(x_{k+1}^i | x_k^j, a_k)w_k^j\leq 0\nonumber
\end{align}
\qed 
\subsection{Proof for Theorem \ref{theorem:MonotonicityConv}}\label{proof:MonotonicityConv}
We first prove that 
\begin{align}
	\overline{\Delta}^s(b,a,b') \geq \overline{\Delta}^{s+1}(b,a,b') \geq 0
\end{align}
Recall that from the previous proof equation \eqref{eq:upperboundplusentropy}
\begin{align}
	&\overline{\Delta}^s(b,a,b') =  \nonumber \\
	&\sum_{i \notin  A^s_{k+1}} \!\!\! w_{k+1}^i\log \! \left[ m \cdot \probd_Z(z_{k+1} | x_{k+1}^i)\right]-\\
	&\sum_{i\notin  A^s_{k+1}} \!\!\! w_{k+1}^i\log \! \left[\probd_Z(z_{k+1}| x_{k+1}^i) \!\!\sum_{j=1}^{n_x}\probd_T(x_{k+1}^i| x_k^j, a_k)w_k^j\right] \nonumber
\end{align}
Suppose we promote the simplification level. Without loss of generality assume that $A^{s+1}_{k+1} = A^{s}_{k+1} \cup \{ q\}$. From the above we conclude that  $q \notin A^{s}_{k+1} $ 
\begin{align}
	 &\overline{\Delta}^{s+1}(b,a,b') = 	\overline{\Delta}^s(b,a,b') - \nonumber\\
	 & -w_{k+1}^q\Bigg(\log\left[ m \cdot \probd_Z(z_{k+1} | x_{k+1}^q)\right]-\\
	 &-\log\left[\probd_Z(z_{k+1}| x_{k+1}^q) \sum_{j=1}^{n_x}\probd_T (x_{k+1}^q | x_k^j, a_k)w_k^j\right]\Bigg) \nonumber
\end{align}
It is left to prove that
\begin{align}
	 &m \cdot \probd_Z(z_{k+1} | x_{k+1}^q) \geq \\
	 &\geq \probd_Z (z_{k+1}| x_{k+1}^q) \sum_{j=1}^{n_x}\probd_T (x_{k+1}^q| x_k^j, a_k )w_k^j \nonumber
\end{align} 
We already done that in previous theorem. 
Now we prove the second part 

\begin{align}
	\underline{\Delta}^{s}(b,a,b') \geq \underline{\Delta}^{s+1}(b,a,b') \geq 0 
\end{align}
The next equation is minus the equation \eqref{eq:lowerboundplusentropy}
\begin{align}
	&\underline{\Delta}^{s}(b,a,b') = \\
	&\sum_{i=1}^{n_x} w_{k+1}^i\log\!\left[\probd_Z(z_{k+1}| x_{k+1}^i)\!\! \sum_{j=1}^{n_x}\probd_T (x_{k+1}^i| x_k^j, a_k)w_k^j\right] \!\! -\nonumber\\
	&\sum_{i=1}^{n_x} w_{k+1}^i\log\!\left[ \probd_Z(z_{k+1} | x_{k+1}^i)\!\!\sum_{j \in A^s_k} \probd_T(x_{k+1}^i | x_k^j, a_k)w_k^j\right]\nonumber
\end{align}
Assume again without loosing generality that $A^{s+1}_k = A^s_{k} \cup \{ q\}$. In that case 
\begin{align}
	&\underline{\Delta}^{s}(b,a,b') -  \underline{\Delta}^{s+1}(b,a,b') =  \\
	&-\sum_{i=1}^{n_x} \! w_{k+1}^i  \log \!\! \left[ \probd_Z(z_{k+1} | x_{k+1}^i) \!\!\! \sum_{j \in A^s_k} \probd_T(x_{k+1}^i | x_k^j, a_k)w_k^j\right]\\
	&+\sum_{i=1}^{n_x} \! w_{k+1}^i  \log\!\!\left[ \probd_Z(z_{k+1} | x_{k+1}^i) \!\!\! \sum_{j \in A^{s+1}_k} \!\!\! \probd_T(x_{k+1}^i | x_k^j, a_k)w_k^j\right]
\end{align}
Select arbitrary index $i$. We got back to end to previous theorem. Note that by definition the bounds are convergent since we are using all the particles. To see it explicitly suppose that $\{i \notin A^{s}_{k+1}\} = \emptyset$ and  $\{i \notin A^{s}_{k}\} = \emptyset$. We have that 
\begin{align}
	\overline{\Delta}^s(b,a,b') = \underline{\Delta}^s(b,a,b') = 0. 
\end{align} 
This concludes the proof. 
\qed

\subsection{Bounds time complexity analysis}\label{sec:TimeComplexity}
We turn to analyze the time complexity of our method using the chosen bounds \eqref{eq:immediate_bounds_l} and \eqref{eq:immediate_bounds_u}.
We assume the significant bottleneck is querying the motion $\probd_T(x'|x,a)$ and observation $\probd_Z(z|x)$ models respectively.
Assume the belief is approximated by a set of $n_x$ weighted particles,
\begin{align}
	b = \{x^i, w^i\}_{i=1}^{n_x}. \label{eq:BeliefParticlesApp}
\end{align}
Consider the \cite{Boers10fusion} differential entropy approximation for belief at time $k+1$,
\begin{align}
	&\hat{\mathcal{H}}(b_{k}, a_k, z_{k+1}, b_{k+1}) \bydef \underbrace{\log \left[\sum_i \probd_Z(z_{k+1}| x_{k+1}^i) w_k^i\right]}_{(a)} + \label{eq:term_a} \\ 
	&\underbrace{\sum_i w_{k+1}^i\cdot\log\left[ \probd_Z(z_{k+1}| x_{k+1}^i)\sum_{j} \probd_T(x_{k+1}^i|x_k^j, a_k)w_k^j\right]}_{(b)} \label{eq:term_b}
\end{align}
Denote the time to query the observation and motion models a single time as $t_{obs}, t_{mot}$ respectively.
It is clear from \eqref{eq:BeliefParticlesApp}, \eqref{eq:term_a} (term a) and, \eqref{eq:term_b} (term b) that:
\begin{align}
	\forall b \text{ as in \eqref{eq:BeliefParticlesApp} }  \Theta(\hat{\mathcal{H}}(b)) = \Theta(n_x \cdot t_{obs} + n_x^2 \cdot t_{mot}).\label{eq:boers_complexity}
\end{align}
Since we share calculation between the bounds, the bounds' time complexity, for some level of simplification $s$, is:
\begin{align}
	\Theta(\ell^s + u^s) = \Theta(n_x\cdot t_{obs} + n_x^s\cdot n_x \cdot t_{mot}),\label{eq:bounds_complexity}
\end{align}
where $n_x^s$ is the size of the particles subset that is currently used for the bounds calculations, e.g.~$n_x^s= |A^s|$ ($A^s$ is as in \eqref{eq:immediate_bounds_l} and \eqref{eq:immediate_bounds_u}) and $\ell^s, u^s$ denotes the immediate upper and lower bound using simplification level $s$. Further, we remind the simplification levels are discrete, finite, and satisfy
\begin{align}
	s \in \{1, 2, \ldots, n_{\mathrm{max}}\},  \ \ \ell^{s=n_{\mathrm{max}}} = -\hat{\mathcal{H}}=u^{s=n_{\mathrm{max}}}. \label{eq:simpl_levels}
\end{align}

Now, assume we wish to tighten $\ell^s, u^s$ and move from simplification level $s$ to $s+1$. 
Since the bounds are updated incrementally (as introduced by \cite{Sztyglic22iros}), when moving from simplification level $s$ to $s+1$ the only additional data we are missing are the new values of the observation and motion models for the newly added particles. Thus, we get that the time complexity of moving from one simplification level to another is:
\begin{align}
	\Theta(\ell^s + u^s \rightarrow \ell^{s+1} + u^{s+1}) = \Theta((n_x^{s+1}-n_x^s)\cdot n_x \cdot t_{mot}),\label{eq:increase_bounds_complexity}
\end{align}
where $	\Theta(\ell^s + u^s \rightarrow \ell^{s+1} + u^{s+1})$ denotes the time complexity of updating the bounds from one simplification level to the following one. Note the first term from \eqref{eq:bounds_complexity}, $n_x\cdot t_{obs}$, is not present in \eqref{eq:increase_bounds_complexity}. This term has nothing to do with simplification level $s$ and it is calculated linearly over all particles $n_x$.  Thus, it is calculated once at the beginning (initial/lowest simplification level).

We can now deduce using \eqref{eq:bounds_complexity} and \eqref{eq:increase_bounds_complexity}
\begin{align}
	&\Theta(\ell^{s+1} + u^{s+1}) = \label{eq:bounds_update_complexity} \\
	&\Theta(\ell^s + u^s) +  \Theta(\ell^s + u^s \rightarrow \ell^{s+1} + u^{s+1}). \nonumber
\end{align}
Finally, using \eqref{eq:boers_complexity}, \eqref{eq:bounds_complexity}, \eqref{eq:simpl_levels}, \eqref{eq:increase_bounds_complexity}, and \eqref{eq:bounds_update_complexity}, we come to the conclusion that if at the end of a planning  session, a node's $b$ simplification level was $1\leq s \leq n_{\mathrm{max}}$ than the time complexity saved for that node is 
\begin{align}
	\Theta((n_x-n_x^s)\cdot n_x \cdot t_{mot}) \label{eq:time_saved}.
\end{align}
This makes perfect sense since if we had to resimplify all the way to the maximal level we get $s=n_{\mathrm{max}}\Rightarrow n_x^{s=n_{\mathrm{max}}}=n_x$ and by substituting $n_x^s=n_x$ in \eqref{eq:time_saved} we saved no time at all.

To conclude, the total speedup of the algorithm is dependent on how many belief nodes' bounds were not resimplified to the maximal level. The more nodes we had at the end of a planning session with lower simplification levels, the more speedup we get according to \eqref{eq:time_saved}.

\bibliographystyle{SageH}

\begin{thebibliography}{48}
\providecommand{\natexlab}[1]{#1}
\providecommand{\url}[1]{\texttt{#1}}
\providecommand{\urlprefix}{URL }
\expandafter\ifx\csname urlstyle\endcsname\relax
  \providecommand{\doi}[1]{DOI:\discretionary{}{}{}#1}\else
  \providecommand{\doi}{DOI:\discretionary{}{}{}\begingroup
  \urlstyle{rm}\Url}\fi

\bibitem[{Araya et~al.(2010)Araya, Buffet, Thomas and Charpillet}]{Araya10nips}
Araya M, Buffet O, Thomas V and Charpillet F (2010) A pomdp extension with
  belief-dependent rewards.
\newblock In: \emph{Advances in Neural Information Processing Systems (NIPS)}.
  pp. 64--72.

\bibitem[{Auer et~al.(2002)Auer, Cesa-Bianchi and Fischer}]{Auer02ml}
Auer P, Cesa-Bianchi N and Fischer P (2002) Finite-time analysis of the
  multiarmed bandit problem.
\newblock \emph{Machine learning} 47(2): 235--256.

\bibitem[{Auger et~al.(2013)Auger, Couetoux and Teytaud}]{Auger13Sp}
Auger D, Couetoux A and Teytaud O (2013) Continuous upper confidence trees with
  polynomial exploration--consistency.
\newblock In: \emph{Machine Learning and Knowledge Discovery in Databases:
  European Conference, ECML PKDD 2013, Prague, Czech Republic, September 23-27,
  2013, Proceedings, Part I 13}. Springer, pp. 194--209.

\bibitem[{Barenboim and Indelman(2022)}]{Barenboim22ijcai}
Barenboim M and Indelman V (2022) Adaptive information belief space planning.
\newblock In: \emph{the 31st International Joint Conference on Artificial
  Intelligence and the 25th European Conference on Artificial Intelligence
  (IJCAI-ECAI)}.

\bibitem[{Barenboim and Indelman(2023)}]{Barenboim23nips}
Barenboim M and Indelman V (2023) Online pomdp planning with anytime
  deterministic guarantees.
\newblock In: \emph{Advances in Neural Information Processing Systems (NIPS)}.

\bibitem[{{Boers} et~al.(2010){Boers}, {Driessen}, {Bagchi} and
  {Mandal}}]{Boers10fusion}
{Boers} Y, {Driessen} H, {Bagchi} A and {Mandal} P (2010) Particle filter based
  entropy.
\newblock In: \emph{2010 13th International Conference on Information Fusion}.
  pp. 1--8.
\newblock \doi{10.1109/ICIF.2010.5712013}.

\bibitem[{Burgard et~al.(1997)Burgard, Fox and Thrun}]{Burgard97ijcai}
Burgard W, Fox D and Thrun S (1997) Active mobile robot localization.
\newblock In: \emph{Intl. Joint Conf. on AI (IJCAI)}. Citeseer, pp. 1346--1352.

\bibitem[{Crisan and Doucet(2002)}]{Crisan02tsp}
Crisan D and Doucet A (2002) A survey of convergence results on particle
  filtering for practitioners.
\newblock \emph{{IEEE} Trans. Signal Processing} .

\bibitem[{Dressel and Kochenderfer(2017)}]{Dressel17icaps}
Dressel L and Kochenderfer MJ (2017) Efficient decision-theoretic target
  localization.
\newblock In: Barbulescu L, Frank J, Mausam and Smith SF (eds.)
  \emph{Proceedings of the Twenty-Seventh International Conference on Automated
  Planning and Scheduling, {ICAPS} 2017, Pittsburgh, Pennsylvania, USA, June
  18-23, 2017}. {AAAI} Press, pp. 70--78.

\bibitem[{Egorov et~al.(2017)Egorov, Sunberg, Balaban, Wheeler, Gupta and
  Kochenderfer}]{egorov2017pomdps}
Egorov M, Sunberg ZN, Balaban E, Wheeler TA, Gupta JK and Kochenderfer MJ
  (2017) Pomdps. jl: A framework for sequential decision making under
  uncertainty.
\newblock \emph{The Journal of Machine Learning Research} 18(1): 831--835.

\bibitem[{Elimelech and Indelman(2022)}]{Elimelech22ijrr}
Elimelech K and Indelman V (2022) Simplified decision making in the belief
  space using belief sparsification.
\newblock \emph{The International Journal of Robotics Research} 41(5):
  470--496.

\bibitem[{Farhi and Indelman(2021)}]{Farhi21arxiv}
Farhi E and Indelman V (2021) ix-bsp: Incremental belief space planning.
\newblock \emph{arXiv preprint arXiv:2102.09539} .

\bibitem[{Farhi and Indelman(2019)}]{Farhi19icra}
Farhi EI and Indelman V (2019) ix-bsp: Belief space planning through
  incremental expectation.
\newblock In: \emph{IEEE Intl. Conf. on Robotics and Automation (ICRA)}.

\bibitem[{Fehr et~al.(2018)Fehr, Buffet, Thomas and Dibangoye}]{Fehr18nips}
Fehr M, Buffet O, Thomas V and Dibangoye J (2018) rho-pomdps have
  lipschitz-continuous epsilon-optimal value functions.
\newblock In: Bengio S, Wallach H, Larochelle H, Grauman K, Cesa-Bianchi N and
  Garnett R (eds.) \emph{Advances in Neural Information Processing Systems 31}.
  Curran Associates, Inc., pp. 6933--6943.

\bibitem[{Fischer and Tas(2020)}]{Fischer20icml}
Fischer J and Tas OS (2020) Information particle filter tree: An online
  algorithm for pomdps with belief-based rewards on continuous domains.
\newblock In: \emph{Intl. Conf. on Machine Learning (ICML)}. Vienna, Austria.

\bibitem[{Garg et~al.(2019)Garg, Hsu and Lee}]{Garg19rss}
Garg NP, Hsu D and Lee WS (2019) Despot-$\alpha$: Online pomdp planning with
  large state and observation spaces.
\newblock In: \emph{Robotics: Science and Systems (RSS)}.

\bibitem[{Hoerger and Kurniawati(2021)}]{Hoerger21icra}
Hoerger M and Kurniawati H (2021) An on-line pomdp solver for continuous
  observation spaces.
\newblock In: \emph{IEEE Intl. Conf. on Robotics and Automation (ICRA)}. IEEE,
  pp. 7643--7649.

\bibitem[{Hoerger et~al.(2020)Hoerger, Kurniawati, Bandyopadhyay and
  Elfes}]{Hoerger20sp}
Hoerger M, Kurniawati H, Bandyopadhyay T and Elfes A (2020) Linearization in
  motion planning under uncertainty.
\newblock In: \emph{Algorithmic Foundations of Robotics XII: Proceedings of the
  Twelfth Workshop on the Algorithmic Foundations of Robotics}. Springer, pp.
  272--287.

\bibitem[{Hoerger et~al.(2019)Hoerger, Kurniawati and Elfes}]{Hoerger19isrr}
Hoerger M, Kurniawati H and Elfes A (2019) Multilevel monte-carlo for solving
  pomdps online.
\newblock In: \emph{Proc. International Symposium on Robotics Research (ISRR)}.

\bibitem[{Hollinger and Sukhatme(2014)}]{Hollinger14ijrr}
Hollinger GA and Sukhatme GS (2014) Sampling-based robotic information
  gathering algorithms.
\newblock \emph{Intl. J. of Robotics Research} : 1271--1287.

\bibitem[{Indelman et~al.(2015)Indelman, Carlone and Dellaert}]{Indelman15ijrr}
Indelman V, Carlone L and Dellaert F (2015) Planning in the continuous domain:
  a generalized belief space approach for autonomous navigation in unknown
  environments.
\newblock \emph{Intl. J. of Robotics Research} 34(7): 849--882.

\bibitem[{Kearns et~al.(2002)Kearns, Mansour and Ng}]{Kearns02jml}
Kearns M, Mansour Y and Ng AY (2002) A sparse sampling algorithm for
  near-optimal planning in large markov decision processes.
\newblock \emph{Machine learning} 49(2): 193--208.

\bibitem[{Kitanov and Indelman(2024)}]{Kitanov24ijrr}
Kitanov A and Indelman V (2024) Topological belief space planning for active
  slam with pairwise gaussian potentials and performance guarantees.
\newblock \emph{Intl. J. of Robotics Research} 43(1): 69--97.
\newblock \doi{10.1177/02783649231204898}.

\bibitem[{Kochenderfer et~al.(2022)Kochenderfer, Wheeler and
  Wray}]{Kochenderfer22book}
Kochenderfer M, Wheeler T and Wray K (2022) \emph{Algorithms for Decision
  Making}.
\newblock MIT Press.

\bibitem[{Kocsis and Szepesv{\'a}ri(2006)}]{Kocsis06ecml}
Kocsis L and Szepesv{\'a}ri C (2006) Bandit based monte-carlo planning.
\newblock In: \emph{European conference on machine learning}. Springer, pp.
  282--293.

\bibitem[{Kopitkov and Indelman(2017)}]{Kopitkov17ijrr}
Kopitkov D and Indelman V (2017) No belief propagation required: Belief space
  planning in high-dimensional state spaces via factor graphs, matrix
  determinant lemma and re-use of calculation.
\newblock \emph{Intl. J. of Robotics Research} 36(10): 1088--1130.

\bibitem[{Kopitkov and Indelman(2019)}]{Kopitkov19ijrr}
Kopitkov D and Indelman V (2019) General purpose incremental covariance update
  and efficient belief space planning via factor-graph propagation action tree.
\newblock \emph{Intl. J. of Robotics Research} 38(14): 1644--1673.

\bibitem[{Kurniawati et~al.(2008)Kurniawati, Hsu and Lee}]{Kurniawati08rss}
Kurniawati H, Hsu D and Lee WS (2008) {SARSOP}: Efficient point-based {POMDP}
  planning by approximating optimally reachable belief spaces.
\newblock In: \emph{Robotics: Science and Systems (RSS)}.

\bibitem[{Lev-Yehudi et~al.(2024)Lev-Yehudi, Barenboim and
  Indelman}]{LevYehudi24aaai}
Lev-Yehudi I, Barenboim M and Indelman V (2024) Simplifying complex observation
  models in continuous pomdp planning with probabilistic guarantees and
  practice.
\newblock In: \emph{AAAI Conf. on Artificial Intelligence}.

\bibitem[{Munos(2014)}]{Munos2014book}
Munos R (2014) \emph{From Bandits to Monte-Carlo Tree Search: The Optimistic
  Principle Applied to Optimization and Planning}.

\bibitem[{Papadimitriou and Tsitsiklis(1987)}]{Papadimitriou87math}
Papadimitriou C and Tsitsiklis J (1987) The complexity of {Markov} decision
  processes.
\newblock \emph{Mathematics of operations research} 12(3): 441--450.

\bibitem[{Pathak et~al.(2018)Pathak, Thomas and Indelman}]{Pathak18ijrr}
Pathak S, Thomas A and Indelman V (2018) A unified framework for data
  association aware robust belief space planning and perception.
\newblock \emph{Intl. J. of Robotics Research} 32(2-3): 287--315.

\bibitem[{Platt et~al.(2010)Platt, Tedrake, Kaelbling and
  Lozano-P\'erez}]{Platt10rss}
Platt R, Tedrake R, Kaelbling L and Lozano-P\'erez T (2010) Belief space
  planning assuming maximum likelihood observations.
\newblock In: \emph{Robotics: Science and Systems (RSS)}. Zaragoza, Spain, pp.
  587--593.

\bibitem[{Shienman and Indelman(2022)}]{Shienman22isrr}
Shienman M and Indelman V (2022) Nonmyopic distilled data association belief
  space planning under budget constraints.
\newblock In: \emph{Proc. of the Intl. Symp. of Robotics Research (ISRR)}.

\bibitem[{Silver and Veness(2010)}]{Silver10nips}
Silver D and Veness J (2010) Monte-carlo planning in large pomdps.
\newblock In: \emph{Advances in Neural Information Processing Systems (NIPS)}.
  pp. 2164--2172.

\bibitem[{Smith and Simmons(2004)}]{Smith04uai}
Smith T and Simmons R (2004) Heuristic search value iteration for pomdps.
\newblock In: \emph{Conf. on Uncertainty in Artificial Intelligence (UAI)}. pp.
  520--527.

\bibitem[{Spaan et~al.(2015)Spaan, Veiga and Lima}]{Spaan2015aamas}
Spaan MT, Veiga TS and Lima PU (2015) Decision-theoretic planning under
  uncertainty with information rewards for active cooperative perception.
\newblock \emph{Autonomous Agents and Multi-Agent Systems} 29(6): 1157--1185.

\bibitem[{Stachniss et~al.(2005)Stachniss, Grisetti and
  Burgard}]{Stachniss05rss}
Stachniss C, Grisetti G and Burgard W (2005) Information gain-based exploration
  using {Rao-Blackwellized} particle filters.
\newblock In: \emph{Robotics: Science and Systems (RSS)}. pp. 65--72.

\bibitem[{Sunberg and Kochenderfer(2018)}]{Sunberg18icaps}
Sunberg Z and Kochenderfer M (2018) Online algorithms for pomdps with
  continuous state, action, and observation spaces.
\newblock In: \emph{Proceedings of the International Conference on Automated
  Planning and Scheduling}, volume~28.

\bibitem[{Sutton and Barto(2018)}]{Sutton18book}
Sutton RS and Barto AG (2018) \emph{Reinforcement learning: An introduction}.
\newblock MIT press.

\bibitem[{Sztyglic and Indelman(2022)}]{Sztyglic22iros}
Sztyglic O and Indelman V (2022) Speeding up online pomdp planning via
  simplification.
\newblock In: \emph{IEEE/RSJ Intl. Conf. on Intelligent Robots and Systems
  (IROS)}.

\bibitem[{Thrun et~al.(2005)Thrun, Burgard and Fox}]{Thrun05book}
Thrun S, Burgard W and Fox D (2005) \emph{Probabilistic Robotics}.
\newblock The MIT press, Cambridge, MA.

\bibitem[{Van Den~Berg et~al.(2012)Van Den~Berg, Patil and
  Alterovitz}]{VanDenBerg12ijrr}
Van Den~Berg J, Patil S and Alterovitz R (2012) Motion planning under
  uncertainty using iterative local optimization in belief space.
\newblock \emph{Intl. J. of Robotics Research} 31(11): 1263--1278.

\bibitem[{Walsh et~al.(2010)Walsh, Goschin and Littman}]{Walsh10aaai}
Walsh T, Goschin S and Littman M (2010) Integrating sample-based planning and
  model-based reinforcement learning.
\newblock In: \emph{AAAI Conf. on Artificial Intelligence}, volume~24.

\bibitem[{Ye et~al.(2017)Ye, Somani, Hsu and Lee}]{Ye17jair}
Ye N, Somani A, Hsu D and Lee WS (2017) Despot: Online pomdp planning with
  regularization.
\newblock \emph{JAIR} 58: 231--266.

\bibitem[{Zhitnikov and Indelman(2022{\natexlab{a}})}]{Zhitnikov22arxiv}
Zhitnikov A and Indelman V (2022{\natexlab{a}}) Risk aware adaptive
  belief-dependent probabilistically constrained continuous pomdp planning.
\newblock \emph{arXiv preprint arXiv:2209.02679} .

\bibitem[{Zhitnikov and Indelman(2022{\natexlab{b}})}]{Zhitnikov22ai}
Zhitnikov A and Indelman V (2022{\natexlab{b}}) Simplified risk aware decision
  making with belief dependent rewards in partially observable domains.
\newblock \emph{Artificial Intelligence, Special Issue on ``Risk-Aware
  Autonomous Systems: Theory and Practice"} .

\bibitem[{Zhitnikov and Indelman(2024)}]{Zhitnikov24tro}
Zhitnikov A and Indelman V (2024) Simplified continuous high dimensional belief
  space planning with adaptive probabilistic belief-dependent constraints.
\newblock \emph{{IEEE} Trans. Robotics} .

\end{thebibliography}

\end{document}